\pdfoutput=1
\documentclass[10pt,journal]{IEEEtran}

\usepackage{caption}
\usepackage{amsmath}
\usepackage{graphicx}
\usepackage{subfig}
\usepackage{cite}
\usepackage{amssymb}
\usepackage{bm}
\usepackage{microtype}

\usepackage{url}

\usepackage{mathtools}  %
\usepackage{xcolor}
\usepackage{mdframed}

\newtheorem{theorem}{Theorem}[section]

\usepackage[crop=pdfcrop,cleanup={.tex, .dvi, .ps, .pdf, .log, .aux, .bbl, .out}]{pstool}

\usepackage{etoolbox}
\makeatletter
\pretocmd\start@gather{%
  \ifnum\mdf@envdepth>0
    \kern\glueexpr-\topskip-\abovedisplayskip\relax
  \fi
}{}{\fail}
\pretocmd\start@align{%
  \ifnum\mdf@envdepth>0
    \kern\glueexpr-\topskip-\abovedisplayskip\relax
  \fi
}{}{\fail}
\pretocmd\start@multline{%
  \ifnum\mdf@envdepth>0
    \kern\glueexpr-\topskip-\abovedisplayskip\relax
  \fi
}{}{\fail}
\makeatother

\title{Statistical Hypothesis Testing Based on Machine Learning: Large Deviations Analysis}

\author{Paolo~Braca,~\IEEEmembership{Senior~Member,~IEEE,}
        Leonardo~M.~Millefiori,~\IEEEmembership{Member,~IEEE,}
        Augusto~Aubry,~\IEEEmembership{Senior~Member,~IEEE,}
        Stefano~Marano,~\IEEEmembership{Senior~Member,~IEEE,}
        \newline
        Antonio~De~Maio,~\IEEEmembership{Fellow,~IEEE,}
        Peter~Willett,~\IEEEmembership{Fellow,~IEEE}%
}
\begin{document}

\maketitle
\begin{abstract}
   We study the performance---and specifically the rate at which the error probability converges to zero---of Machine Learning (ML) classification techniques.
   Leveraging the 
   theory of large deviations, we provide the mathematical conditions for a ML classifier %
   to exhibit error probabilities that vanish exponentially, say $\sim \exp\left(-n\,I + o(n) \right)$, where $n$ is the number of informative observations available for testing (or another relevant parameter, such as the size of the target in an image) and $I$ is the error rate. 
   Such conditions depend on the Fenchel-Legendre transform of the 
   cumulant-generating function of the 
   Data-Driven Decision Function (D3F, i.e., what is thresholded before the final binary decision is made) learned in the training phase. As such, the D3F and, consequently, the related error rate $I$, depend on the given training set, which is assumed of finite size. Interestingly, these conditions can be verified and tested numerically exploiting the available dataset, or a synthetic dataset, generated according to the available information on the underlying statistical model. In other words, the classification error probability convergence to zero and its rate can be computed on a portion of the dataset available for training. 
   Coherently with the large deviations theory, we can also establish the convergence, for $n$ large enough, of the normalized D3F statistic to a Gaussian distribution. This property is exploited to set a desired asymptotic false alarm probability, which empirically turns out to be accurate even for quite realistic values of $n$. 
   Furthermore, approximate error probability curves $\sim \zeta_n \exp\left(-n\,I \right)$ are provided, thanks to the refined asymptotic derivation (often referred to as exact asymptotics), where $\zeta_n$ represents the most representative sub-exponential terms of the error probabilities. Leveraging the refined asymptotic, we are able to compute an analytically accurate approximation of the classification performance for both the regimes of small and large values of $n$. 
   Theoretical findings are corroborated by extensive numerical simulations. Real-world data, acquired by an X-band maritime radar system for surveillance, are exploited to assess the validity of the proposed theory.
\end{abstract}

\begin{IEEEkeywords}
Machine Learning, Deep Learning, Large Deviations Principle, Exact Asymptotics, Statistical Hypothesis Testing, Fenchel-Legendre Transform, Extended Target Detection, Radar/Sonar Detection, X-band Maritime Radar. 
\end{IEEEkeywords}

\section{Introduction}
As we enter the Artificial Intelligence (AI) technological age, the ambition is to offer the augmentation and the potential replacement of tedious human tasks and activities within a wide range of industrial, intellectual and social applications, with an impact similar to that produced by %
the industrial revolution~\cite{dwivedi2021artificial}.  %
The impact of AI can be significant in many %
fields: finance, healthcare, manufacturing, retail, supply chain, logistics and utilities, all potentially disrupted by the onset of AI technologies~\cite{dwivedi2021artificial}. 

New breakthroughs in algorithmic Machine Learning (ML) and autonomous decision-making are offering countless opportunities for innovation. ML methods have been providing breakthroughs in important research problems, which in some cases have been open for more than 50 years. This is the case, for instance, of the neural network-based model, AlphaFold, that is able to predict with high accuracy the three-dimensional structure of a protein from its amino acid sequence~\cite{jumper2021highly}. Notably, AlphaFold is the first computational method that can predict protein structures %
even in cases in which no similar structure is known~\cite{jumper2021highly}. Another example is the success of AlphaZero, based on Reinforcement Learning (RL) principles \cite{bertsekas2021rollout,bertsekas2022lessons}. AlphaZero not only outperforms all chess programs but it discovers new ways to play chess \cite{silver2016mastering,silver2017mastering}.

In~\cite{lippert2017identification} ML methods are applied to predict human physical traits (e.g., face and voice) and other relevant information (e.g., height and weight) from genomic data. More recently, AI-based techniques provided critical support to combat the COVID-19 pandemic~\cite{vaishya2020artificial}; examples are the use of ML to analyze the epidemiological curve evolution~\cite{Soldi2021,braca2021decision,Shanglin_2021} and medical images~\cite{COVID-19_medical,Deep_learning_covid-19}. 

A fundamental problem addressed by ML, which spans several research and application fields, is
to discover intricate 
structures in large data sets, and in this context many applications make use of a class of techniques known as ``deep learning'' (DL)~\cite{lecun2015deep}. Conventional ML techniques were limited in their ability to process data in their raw form. For decades, constructing an ML system required careful engineering and considerable domain expertise to design a feature extractor that transformed the raw data into a suitable internal representation from which the learning subsystem could infer patterns in the input~\cite{lecun2015deep}. Conversely, DL methods are representation-learning\footnote{Representation learning is a set of methods that allows a machine to be fed with raw data and to automatically discover the representations needed for detection or classification.} methods with multiple levels of representation, obtained by composing simple non-linear modules, %
each of which transforms the representation at one level (starting with the raw input) into a representation at a higher, slightly more abstract level. By the composition of enough such transformations, very complex functions can be learned, and in quite a few contexts DL methods represent nowadays the state-of-the-art in terms of performance~\cite{lecun2015deep}. 

Examples of applications in which DL methods achieve state-of-the-art performance include the following. DL architectures based on Convolutional Neural Networks (CNNs) achieve unprecedented performance in skin cancer classification, with an accuracy comparable to that of dermatologists~\cite{esteva2017dermatologist}. DL architectures based on Recurrent Neural Networks (RNN) are able to decode the electrocorticogram with high accuracy and at natural-speech rates~\cite{makin2020machine}. These neural networks can be trained to encode sequences of neural activity into an abstract representation, and then decode this representation, word by word, into an English sentence with a 3$\%$ average word error rate~\cite{makin2020machine}. DL and deep RL are also key components of new-generation autonomous driving systems, see, e.g.,~\cite{mozaffari2020deep, grigorescu2020survey}, and nowadays are also widely exploited in surveillance systems, such as Synthetic Aperture Radar (SAR) imaging, see, e.g.,\cite{DeepSAR2016,CNN_SAR2016,SAR_ATR_2016,Martorella2022}. %
In space-based surveillance, DL offers the capability to accurately classify vessels from satellite sensor imaging (see, e.g.,~\cite{SoldiPartI,SoldiPartII}), and in the context of maritime situational awareness RNNs are able to accurately predict vessel positions several hours ahead~\cite{Capobianco2021,CapobiancoFusion21,murray2021ais,Capobianco2022}. In video analysis and image understanding, DL methods represent the state of the art for object detection~\cite{ObjectDetection_DL_2019} and multi-object tracking~\cite{traktor}. DL is also used in Multiple Input Multiple Output (MIMO) communications~\cite{Learning_to_Detect_TSP_2019}, active sensing for communications~\cite{sohrabi2021active},
radar and sonar processing~\cite{Wang_SPL_2019,SoldiICASSP2020,GaglioneSPL2020,DL_Radar_2021,McDonald2021,AES_Magazine2022}. 

Given the tremendous success of AI, the opportunities and challenges of merging AI and sensor data fusion are under investigation by several research groups with special focus on computational efficiency, improved decision making, security, multi-domain operations, and human-machine teaming~\cite{Blasch2021}. Ethical aspects concerning the AI are also of paramount importance;
in this regard, digital ethics for AI and information fusion in the context of the defense domain is discussed in~\cite{Koch2021}.

Compared to the great success of DL techniques, little is known about their mathematical properties. There is still a lack of methodological and systematic approaches to analyze DL techniques~\cite{yu2019understanding}, and no comprehensive understanding of the optimization process and internal organization of DL architectures is available; indeed, DL methods are usually regarded as ``black boxes,'' see, e.g.,~\cite{alain2017understanding}. 

In recent years, several attempts to fill this gap 
have been in progress, including those based on statistical and information theoretical interpretations~\cite{anden2014deep,tishby2015deep,mallat2016understanding,shwartz2017opening,yu2019understanding}. %
In~\cite{anden2014deep,mallat2016understanding}, interpretation of CNNs in terms of a cascade of filters implementing wavelet transforms and pointwise nonlinearities is provided. %
These works develop a mathematical framework to analyse CNN properties %
involving multiscale contractions with wavelets, linearization of hierarchical symmetries, and sparse separations. More recently, an asymptotic equivalence between \emph{infinitely wide} deep neural networks and Gaussian processes is established exploiting the Central Limit Theorem (CLT)~\cite{lee2018deep,novak2018bayesian}.
In~\cite{lee2018deep}, as well as in the %
seminal work~\cite{Neal1996}, the diverging parameter is the width of the hidden layers of the network. %

The lack of methodological and systematic approaches to investigate the optimization process and internal organization of DL techniques also exists for the derivation of their ultimate performance limit. The authors of the present paper are not aware of any results---or even ongoing efforts---to fill this gap. In particular, it is currently unknown if DL techniques can or cannot achieve classification error probabilities close to optimal, i.e., vanishing exponentially with the datasize~\cite{blahut1974hypothesis,Zeitouni92,bajovic2011distributed,Matta_IT2016,Matta_2016,Marano2019,Marano2020}. 

This challenge is here addressed using a formulation that departs from the setting adopted in %
the classical statistical learning theory~\cite{vapnik1999overview}. In particular, the parameters characterizing the ML procedure such as the NN weights, the network width, and the size of the training set are fixed; the diverging parameter~$n$  quantifies
the information available for the classification problem (i.e., the number of measurements in a sequence of data). 
For example, if one is trying to detect a target of extent $20$ pixels in a $400$ by $400$  image, $n=20$. And likewise if the image is $1000$ by $1000$ then $n$ remains $20$---that is, the background, at least to first order, does not matter. Continuing, if the target's extent is $30$ samples and it is observed in $10$ consecutive frames, then $n=300$. Note also that the notional detectability (say, the Signal-to-Noise Ratio, SNR) of the target in the image does not impact $n$, as it will be seen to be subsumed within the convergence rate $I$.\footnote{In setting different from that presented in this work, the parameter $n$ could also represent the step size in adaptive learning algorithms over decentralized networks~\cite{Matta_IT2016,Matta_2016}.}

Data-centric approaches exploit large amounts of data. Part of the data, the so-called training set, is used to train the network. The trained network is then fed the observation set to perform the desired classification. Additional datasets, here referred to as characterization sets\footnote{Validation set is a more common terminology in a machine learning context~\cite{xu2018splitting,Bishop_2013}.}, may also be available and are usually exploited to numerically infer the classification error probability of the trained network. Our vision is that these characterization sets can be exploited to derive \emph{analytical expressions} for the classification error probability, which allows one to infer the performance of the trained network in the limit of diverging $n$. Under mild assumptions that can be directly checked by inspection of the characterization sets, %
it is shown that the classification error probability converges to zero exponentially fast with~$n$, at a predictable rate. Thus, %
performance prediction beyond standard numerical estimation of the classification error probability is possible and ultimate classification limits can be established.
The main contributions of this article are as follows.
\begin{itemize}
    \item 
A mathematical framework based on the Large Deviations Principle (LDP)~\cite{Dembo-Zeitouni, hollander} is presented for performance prediction of ML techniques when $n$ diverges; the application domain of such a performance prediction includes, but it is not limited to, different neural network based architectures for classification purposes.

\item It is  shown that the asymptotic classification error probability depends  on a suitable transformation of the Log Moment-Generating Function (LMGF) of the decision statistic; in the case of DL, this decision statistic is determined by the output layer of the network. %

\item 
The mathematical conditions for a ML classifier to exhibit exponential classification error probability $\sim\exp\left(-n\,I + o(n) \right)$, where $I$ is the error rate, are provided; such conditions depend on the Fenchel-Legendre transform of the LMGF of the data-driven decision statistic
learned in the training phase. %

\item It is shown that the error rate $I$ approaches the optimal classification rate as the size of the training set grows; such an optimal rate describes the classification error probability of the log-likelihood ratio test and admits information-theoretical interpretation. 

\item Based on saddlepoint techniques, refined asymptotic expressions (often referred to as exact asymptotics) are derived in the form  $\sim \zeta_n \exp\left(-n\,I \right)$, where~$\zeta_n$ accounts for the most representative sub-exponential term of the classification error probability. 

\end{itemize}

\begin{figure}
    \centering%
    \includegraphics[width=.98\columnwidth, trim=120 250 120 250]{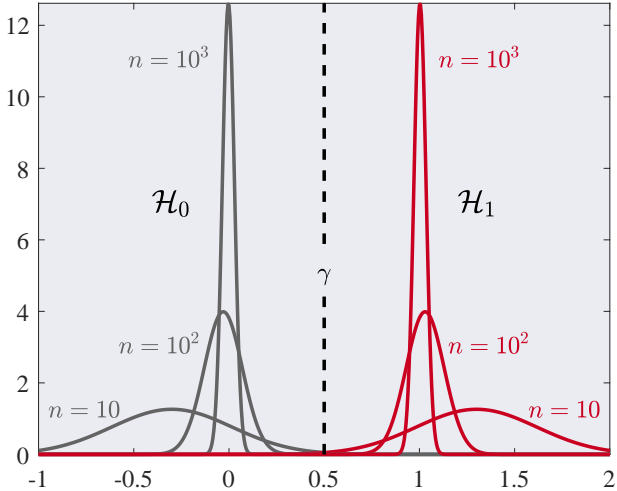}
    \caption{Pictorial representation of the distribution of $T^{(n)}$ for different values of $n$ under ${\cal H}_k$, $k=0,1$. As expressed in (\ref{eq:CLT_start}), the behaviour is the same of the CLT. The distribution of $T^{(n)}$ is approximately Gaussian with mean converging to the asymptotic mean $\mu_k$ and variance $\frac{\sigma_k^2}{n}$ that decreases linearly with $n$. In this example we have $\mu_0=0$, $\mu_1=1$, and $\sigma_k=1$ for $k=0,1$. The distribution of $T^{(n)}$ amasses around $\mu_k$ as $n$ increases. We have represented the threshold, $\gamma$, of the test~(\ref{eq:test0}) in the mid point $0.5$ between $\mu_0$ and $\mu_1$. It is easily seen that the error probabilities~(\ref{eq:error_prob}) converge to zero as $n$ increases. However, it is worthwhile stressing that the rates of convergence of the error probabilities are not ruled by the convergence in distribution reported in this figure; instead, they are ruled by the LDP~(\ref{eq:alpha_n})-(\ref{eq:beta_n}), involving the asymptotic behaviour of the tail probabilities.}
    \label{fig:gaussian_pdf}
\end{figure}

The paper is organized as follows. In Sec.~\ref{sec:preliminaries}, we provide a summary of the main theoretical results. In Sec.~\ref{sec:problem}, we formulate the problem in terms of statistical %
hypothesis testing, with simple hypotheses and assuming independent samples. In Sec.~\ref{sec:LDP}, we provide the convergence rate and the exact asymptotic formula to approximate the error probabilities. In Sec.~\ref{sec:LDP_dependent}, we extend the problem to composite hypotheses and dependent data. In Sec.~\ref{sec:estimation_rate}, we provide details about how to compute the rate function and the exact asymptotics. In Sec.~\ref{sec:experiments}, we discuss extensive numerical simulations and experimental results using real-world data acquired by an X-band marine radar. Finally, the conclusion is given in Sec.~\ref{sec:conclusion}, and mathematical details are provided in appendices.

\section{Preview of the main results}
\label{sec:preliminaries}

\begin{figure*}
    \centering%
    \subfloat[][]{
    \includegraphics[width=0.30\textwidth]{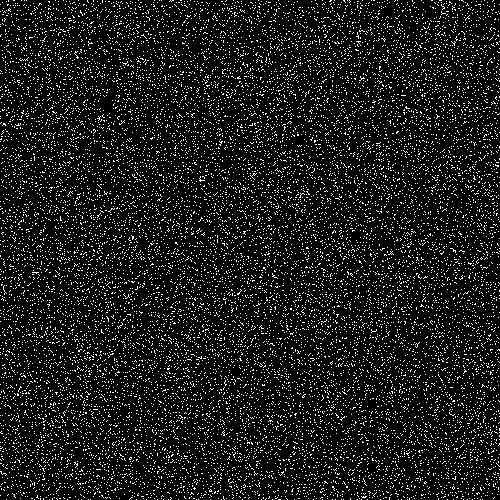}}%
    \hfil%
    \subfloat[][]{
    \includegraphics[width=0.30\textwidth]{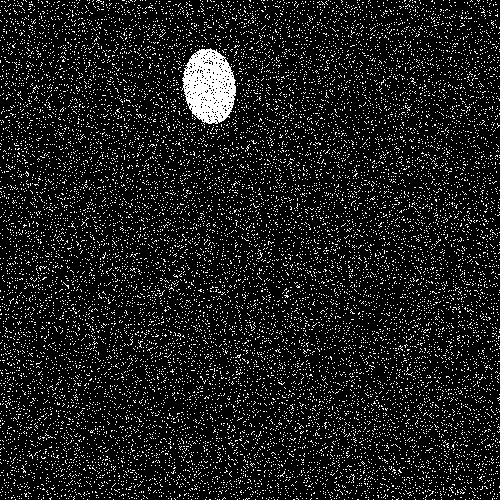}}
    \hfil%
    \subfloat[][]{
    \includegraphics[width=0.30\textwidth]{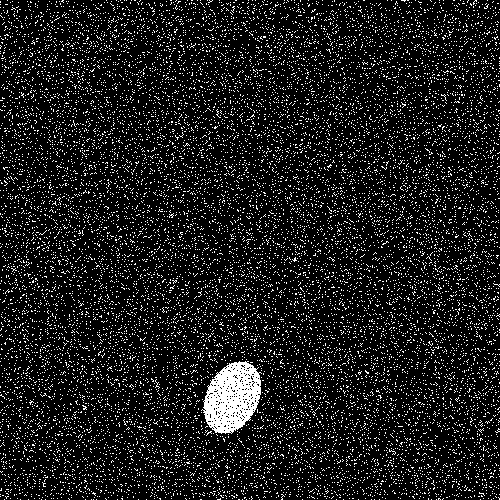}}
    \\%
    \subfloat[][]{
    \includegraphics[width=0.30\textwidth]{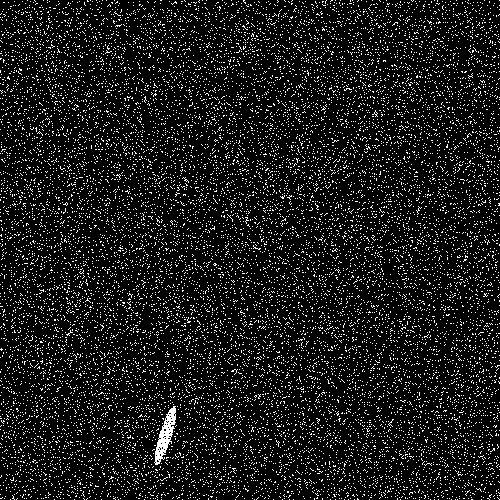}}
    \hfil%
    \subfloat[][]{
    \includegraphics[width=0.30\textwidth]{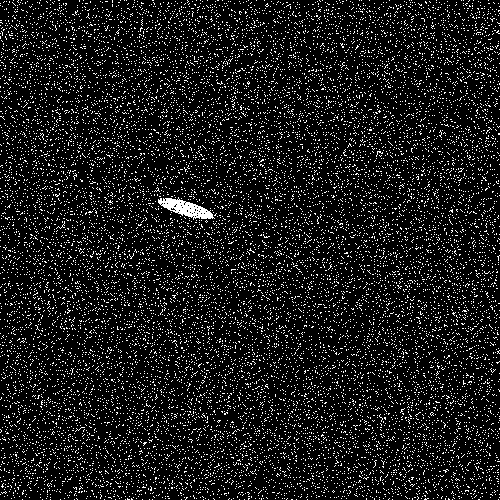}}
    \hfil%
    \subfloat[][]{
    \includegraphics[width=0.30\textwidth]{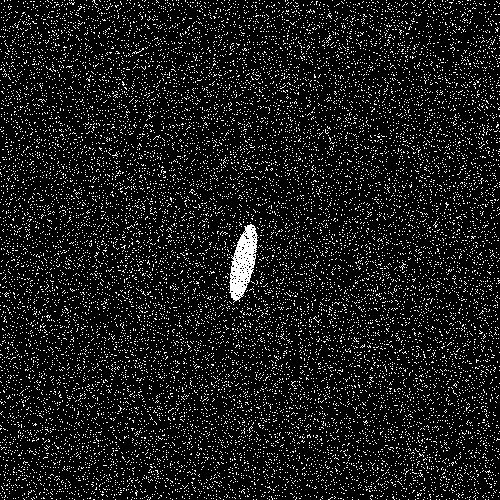}}
    \caption{Notional examples of binary images that represent the input data: (a) under ${\cal H}_0$ (absence of target); and (b)--(f) under ${\cal H}_1$, with the target present in different locations with different orientations and shapes.}
    \label{fig:ETT_Fig}
\end{figure*}

In this section, we describe briefly the main findings of this work. %
Consider a family of real-valued decision statistics $T^{(n)}$ that process a set of observations ${\cal X}^{(n)}$, referred to as the \emph{observation set}. The parameter $n$ has earlier been exemplified and is later defined with more precision, but can be thought of as quantifying the number of informative elements per test datum.
The goal is to decide between two hypotheses: ${\cal H}_0$ and ${\cal H}_1$. Since the distribution of the observations under ${\cal H}_0$ and ${\cal H}_1$ is often unknown, or too complex to derive, we focus on the case that the decision statistic is provided by a learning mechanism operating on a sufficiently large, finite, \emph{training set} ${\cal Y}$ of size $m_y$ available under each hypothesis, independent of ${\cal X}^{(n)}$. Then, the decision statistic $T^{(n)} = T^{(n)}_{\bm{\omega}}$ is referred to as the Data-Driven Decision Function (D3F)%
, where the parameters $\bm{\omega}$ are learned at the training phase. When appropriate, the ideal case that $T^{(n)}$ is the Log-Likelihood Ratio (LLR), see, e.g.,~\cite{Lehmann-testing}, is considered and this serves as a term of comparison with the D3F.

Consider the statistical test 
\begin{equation}
   \left\{ 
   \begin{array}{ll}
     T^{(n)} \geq \gamma_n   & \quad \textnormal{decide ${\cal H}_1$}  \\
      T^{(n)} < \gamma_n  & \quad \textnormal{decide ${\cal H}_0$},
   \end{array} \right.  
   \label{eq:test0}
\end{equation}
where $\gamma_n$ is the threshold, and the error probabilities are 
\begin{equation}
    \alpha_n = \mathbb{P} \left[T^{(n)} \geq \gamma_n \left| {\cal H}_0 \right. \right], \, \beta_n  = \mathbb{P} \left[T^{(n)}  < \gamma_n \left| {\cal H}_1 \right. \right].
    \label{eq:error_prob}
\end{equation}

We study the detection performance of $T^{(n)}$ when $n$ diverges, and propose suitable approximations for the finite sample-size regime of $n$. %
The detection performance is characterized on a distinct set of data, independent of ${\cal Y}$ and ${\cal X}^{(n)}$, that we indicate with ${\cal Z}$, of size $m_z$, and refer to as the \emph{characterization set}. The usage of the characterization set can be seen as similar to the usage of the \emph{validation set} (usually involved in the learning stage; see, e.g.~\cite{xu2018splitting,Bishop_2013}) to evaluate the detection performance. For our scope, an overlap between the characterization and training sets is acceptable as long as the estimators, %
introduced in Sec.~\ref{sec:estimation_rate}, %
properly converge to their expected values, which are the quantities of interest for us to characterize the test performance.\footnote{More details are given in Sec.~\ref{sec:estimation_rate}, and the mathematical conditions for the weak convergence are provided in Appendix~\ref{sec:appendixB}.} %
In general, both the sets ${\cal Y}$ and ${\cal Z}$ can depend on the parameter $n$, but this will be clarified case by case. 

Let us define the LMGF of $T^{(n)}$ 
\begin{equation}
\varphi_{n,k} (t) = \log {\mathbb E}\left[ \exp(t\,T^{(n)}) \left| {\cal H}_k \right.\right], 
    \label{eq:LMGF_start}
\end{equation}
where $ {\mathbb E}\left[ X | {\cal H}_k \right] $ is the expected value of $X$ under ${\cal H}_k$, $k=0,1$.
Under mild regularity conditions, see, e.g.,~\cite{touchette2009large,touchette2011basic}, the performance characterization of the asymptotic decision, based on $T^{(n)}$, depends on the limit of the scaled LMGF\footnote{The LMGF is also referred to as the cumulant generating function. Moreover, the sequence ${1,2,\dots, n}$ can also be replaced by a generic deterministic sequence $a_n$ (see details in~\cite{touchette2009large}). The sequence $a_n$ is referred to as the ``convergence speed,'' and when $a_n = n$, the speed is linear.}~\cite{hollander}
\begin{equation}
    \varphi_k (t)= \lim_{n \rightarrow \infty} \frac{1}{n} \varphi_{n,k} (n\,t). 
    \label{eq:LMGF_asy_start}
\end{equation}
Specifically, if the above limit exists under ${\cal H}_k$, usually two important properties hold. The former, referred to as \emph{small deviations}, implies that $T^{(n)}$ converges (in some sense) for $n\rightarrow \infty$ to an asymptotic value $\mu_k$ under ${\cal H}_k$. The latter, referred to as \emph{large deviations}, characterizes the convergence to zero of the probability that $T^{(n)}$ is ``far away'' from $\mu_k$. Specifically, the small deviations property is related to the convergence in distribution of the normalized statistic $\sqrt{n}(T^{(n)}-\mu_k)$ to a Gaussian random variable, which in our setting is ${\cal N} (0,\sigma_k^2)$, under ${\cal H}_k$. Then, we have:
\begin{equation}
    \sqrt{n}(T^{(n)}-\mu_k) \xrightarrow[]{d} {\cal N} (0,\sigma_k^2),
    \label{eq:CLT_start}
\end{equation}
where $\xrightarrow[]{d}$ indicates the convergence in distribution, whereas $\mu_k$ and $\sigma_k$ are respectively the asymptotic mean and the asymptotic standard deviation of the decision statistic. Such asymptotic values are related to~(\ref{eq:LMGF_asy_start}) as follows~\cite{touchette2009large}: %
\begin{equation}
    \varphi_k^\prime (0) = \mu_k, \quad\quad \varphi_k^{\prime\prime} (0) = \sigma_k^2.
    \label{eq:mu_k_sigma_k}
\end{equation}
The convergence in~\eqref{eq:CLT_start} is clearly related to the CLT, and this will be discussed in detail in the next sections, but a pictorial representation of this convergence is illustrated in Fig.~\ref{fig:gaussian_pdf}. %

Now, let us focus on the latter property, i.e., the LDP. In view of the asymptotic convergence~(\ref{eq:CLT_start}), %
$T^{(n)}$ gets closer to $\mu_k$ under ${\cal H}_k$ as $n$ increases, with a variance that vanishes, as illustrated in Fig.~\ref{fig:gaussian_pdf}. Intuitively, in this framework, for a large enough $n$, we should set the threshold $\gamma_n$ between $\mu_0$ and $\mu_1$ (otherwise we %
incur in the trivial situation that one of the error probabilities converges to unity). In other words, assuming that $\gamma_n \rightarrow \gamma$ and $\mu_0<\mu_1$, it is required that $\mu_0 \leq \gamma \leq \mu_1$.
Then, if $\varphi_k (t)$ is twice differentiable, the following large deviations results hold for $\gamma_n \rightarrow \gamma$ and $n$ sufficiently large:
\begin{mdframed}[backgroundcolor=black!20]
\begin{gather}
\alpha_n \approx \zeta_{n,0} \exp(-n\,I_0(\gamma)), \label{eq:alpha_n}\\ %
\beta_n  \approx \zeta_{n,1} \exp(-n\,I_1(\gamma)),  \label{eq:beta_n} %
\end{gather}
\end{mdframed}
where $I_k(\gamma)$ is referred to as the \emph{rate function} and is given by the Fenchel-Legendre transform of the limit of the scaled LMGF \footnote{In general, the rate function is given by the infimum value of the Fenchel-Legendre transform in the decision region $\Gamma_k$ of ${\cal H}_k$; more details on this are provided in Sec.~\ref{sec:LDP_dependent}.} in~(\ref{eq:LMGF_asy_start}). The terms $\zeta_{n,k}$ in~(\ref{eq:alpha_n})-(\ref{eq:beta_n}) model sub-exponential behaviours and can be %
computed with the so-called ``exact asymptotics,'' (see, e.g.,~\cite{Dembo-Zeitouni}) related to the saddlepoint approximation (see, e.g.,~\cite{Reid88}). The Fenchel-Legendre transform rules the convergence of the tail distribution of $T^{(n)}$, as formalized in~\eqref{eq:alpha_n}-\eqref{eq:beta_n}. 

Both the small deviations and the large deviations principles can be seen as two sides of the same medal; indeed, under mild regularity conditions (see, e.g.,~\cite{BRYC1993253} and the discussion in~\cite{touchette2009large}), the CLT can be obtained as a consequence of the
LDP. %
In other words, the considerations that lead to the LDP are strictly related to that of the CLT.\footnote{In the most general settings, the mathematical conditions required by the CLT and those required by LDP are different, being also mathematically different the two convergences. However, as already pointed out in many relevant situations LDP and CLT are both verified.} 
Finally, it is worth mentioning that the small deviations is useful to fix one of the error probabilities to a desired value, with the other one converging to zero at the maximum achievable rate given the convexity of the rate functions; more details on this aspect will be given in Sec.~\ref{sec:CLT_IID} and~\ref{sec:CLT_Dep}. To give an idea, let us select $\gamma_n \rightarrow \mu_0$, obtaining $I_0(\mu_0) = 0$ and $I_1(\mu_0)\geq I_1(\gamma)$ for all $\mu_0\leq\gamma\leq\mu_1$; the selected operational point is reported in the leftmost side of curves in Fig.~\ref{fig:laplace_iid_rate_distortion}, where $I_1(\gamma)$ is the $y$-axis, $I_0(\gamma)$ the $x$-axis, and $\gamma$ varies in the interval $[\mu_0,\mu_1]$.

Our goal is to derive the LDP for ML-based statistical hypothesis testing. We show that it is possible to compute numerically the error rate $I_k$, as well as the sub-exponential terms $\zeta_{n,k}$, from the characterization set ${\cal Z}$ without assuming %
knowledge of the original distributions. 

We consider the following important scenarios, which are relevant in a number of applications, see, e.g.,~\cite{Varshney_97,Blum_97,asymptotic-rc,bajovic2011distributed,TWR2015,Granstrom15,Matta_IT2016,McDonald2021, Prabhu_22}. The first is perhaps simplistic, but it is convenient to motivate our development.
\begin{enumerate}
    \item We begin by assuming, under ${\cal H}_k$, with $k=0,1$, Independent and Identically Distributed (IID) observations ${\cal X}^{(n)}= \left(x_i\right)_{i=1}^{n}$, 
    with simple hypotheses $x_i\sim f_k(x)$. We derive a decision statistic $T_{\bm{\omega}}^{(n)}$, which is the sample mean of the data in ${\cal X}^{(n)}$ processed accordingly to the elementwise D3F, $t_{\bm{\omega}}(\cdot)$, as in the conceptual diagram illustrated in Fig.~\ref{fig:d3f_arch_iid}; this scenario is described in Sec.~\ref{sec:problem} and~\ref{sec:LDP}.   
    \item We generalize the previous scenario to the case of composite hypotheses, with conditionally independent data given a parameter $\theta\in \Theta$. We have $x_i\sim f(x|\theta_0)$ with $\theta_0 \notin \Theta$ under ${\cal H}_0$ and $x_i\sim f(x|\theta)$ under ${\cal H}_1$, with $\theta$ distributed according to a prior discrete distribution $w_\theta$, that is unknown except through its influence on the training data. In this case, we train a different elementwise D3F for each possible value of $\theta$. Then, the complete decision statistic has a %
    structure that involves the %
    elementwise D3F applied to each sample in ${\cal X}^{(n)}$ for each possible value of $\theta$, as illustrated by the conceptual diagram in Fig.~\ref{fig:d3f_arch_dep}; this scenario is described in detail in Sec.~\ref{sec:LDP_conditionally_independent}.
    \item We further generalize to the case of dependent data under ${\cal H}_0$ and ${\cal H}_1$. This setting is particularly appealing also in the case of conditionally independent data and too large size of the space of the $\theta$ parameter, %
    making the scheme in 2) intractable. Specifically, we focus on a scenario in which the input data ${\cal X}^{(n)}$ are binary images that may or may not contain a target, 
    which occupies more than one pixel (``extended target''). The hypothesis ${\cal H}_0$ represents the absence of the target, and ${\cal H}_1$ represents its presence. The position, shape and orientation of the target are unknown in advance. %
    In this scenario, the parameter $n$ represents the number of pixels occupied by the target in the image. Some examples of the images used in this scenario are reported in Fig.~\ref{fig:ETT_Fig}, whereas the conceptual diagram of the D3F is illustrated in Fig.~\ref{fig:d3fcnn_arch}. Further details are available in Sec.~\ref{sec:LDP_images}. In this scenario, using real-world data acquired by a high-resolution marine radar, we prove the effectiveness of the proposed architecture as well as the capability to predict its decision performance via the developed theoretical tools.
\end{enumerate}
In the first two scenarios, we compare the performance of the D3F, implemented as a suitable combination of fully-connected Neural Networks (NNs), with the ideal test, based on the LLR. In the third scenario, the D3F is implemented as a Deep CNN (DCNN), and the D3F performance cannot be compared with the LLR ideal test because such test is not available.

It is worthwhile mentioning that the first two architectures implicitly exploit the dependency structure of the input data. Our design approach, inspired by the LLR structure, can be seen as a model-based ML approach, similar to, e.g., \cite{shlezinger2021model,shlezinger2022model}, where ML methods take advantage of the statistical modeling of the problem under consideration, enhancing their data-driven vocation.\footnote{The nomenclature ``model-based ML'' was originally introduced by Bishop in~\cite{Bishop_2013} to define an approach to ML where all the assumptions about the problem domain are made explicit in the form of a model. Indeed, it takes the wider meaning of a bespoke method, formulated for new applications, where the solution is expressed through a compact modelling language, and the corresponding custom ML code is then generated automatically.} 

An important remark is related to the computation of the error rate functions and the sub-exponential terms, which are seldom available in closed form, even when the analytical distribution of the data is available. In our framework, given that the distribution of the data may not be available, we propose to estimate the scaled LMGF~(\ref{eq:LMGF_asy_start}), the error rates, and the sub-exponential terms exploiting the characterization set. This is a distinct feature of the present paper. 

Moreover, we will show that the numerical stability of such estimation can be improved with Monte Carlo Methods (MCMs), as discussed in depth in~\cite{touchette2009large,touchette2011basic}. The use of MCMs requires principled generation of synthetic data from the so-called \emph{tilted} distribution via the Metropolis–Hastings algorithm~\cite{touchette2009large,touchette2011basic}. It is worthwhile noting that the generation of synthetic data is a well-explored topic in the ML literature, see, e.g., variational autoencoders (VAEs)~\cite{kingma2019introduction} and generative adversarial networks (GANs)~\cite{ledig2017photo,mao2017least}. In the framework of model-based ML, instead, the generation of synthetic data is straightforward, given that the statistical model is assumed to be (at least partially) known. Another possibility would be to train the D3F with real-world data, and then characterize the test with synthetic data assuming, for instance, different statistical models, or relevant parameters. Clearly, the opposite option is also valid, with the training performed by synthetic data and the testing done by real data. Indeed, this is the approach taken for the third scenario in the above list: the D3F is trained on synthetic data, an example of which is reported in Fig.~\ref{fig:ETT_Fig}, but the performance is then tested on real-world radar data.

\section{Small deviations of the D3F statistical hypothesis testing with IID Observations}
\label{sec:problem}

\subsection{Problem formulation and learning mechanism}

Let us consider the observations ${\cal X}^{(n)}=\left(x_i\right)_{i=1}^n$, where the $x_i$'s are IID according to $f_0(x)$ under hypothesis ${\cal H}_0$ and $f_1(x)$ under hypothesis ${\cal H}_1$. The goal is to decide between ${\cal H}_0$ and ${\cal H}_1$ based on the observed dataset ${\cal X}^{(n)}$.
It is well known that the optimal decision statistic $L^{(n)}$ is the LLR~\cite{Lehmann-testing}, which takes the form of a summation in the case of independent observations:
\begin{equation}
 L^{(n)} = \frac{1}{n} \sum_{i=1}^n l(x_i)  = \frac{1}{n} \sum_{i=1}^n \log \frac{f_1(x_i)}{f_0(x_i)}.
 \label{eq:Ln}
\end{equation}
As already discussed, in %
many practical applications, both distributions $f_k$, $k=0,1$, are unknown, and therefore the LLR test cannot be applied directly. In the present work, the LLR will be used as a term of comparison.

The distributions $f_k$, $k=0,1$, are unknown, %
and a dataset ${\cal Y}_k$ is available for each hypothesis ${\cal H}_k$, $k=0,1$. Specifically, ${\cal Y}_k = \left\{ y_{1,k}, y_{2,k}, \dots, y_{m_y,k} \right\}$, where $y_{j,k}$ is distributed according to $f_k(\cdot)$, 
and $m_y$ is the number of samples available for training under each hypothesis. Hence,
the union of ${\cal Y}_k$, $k=0,1$, is the labeled training set ${\cal Y}$, which is independent of the observed dataset ${\cal X}^{(n)}$~\cite{bishop-book}.  
 
 \begin{figure}
    \centering%
    \includegraphics[width=.9\columnwidth, trim=45 0 20 0]{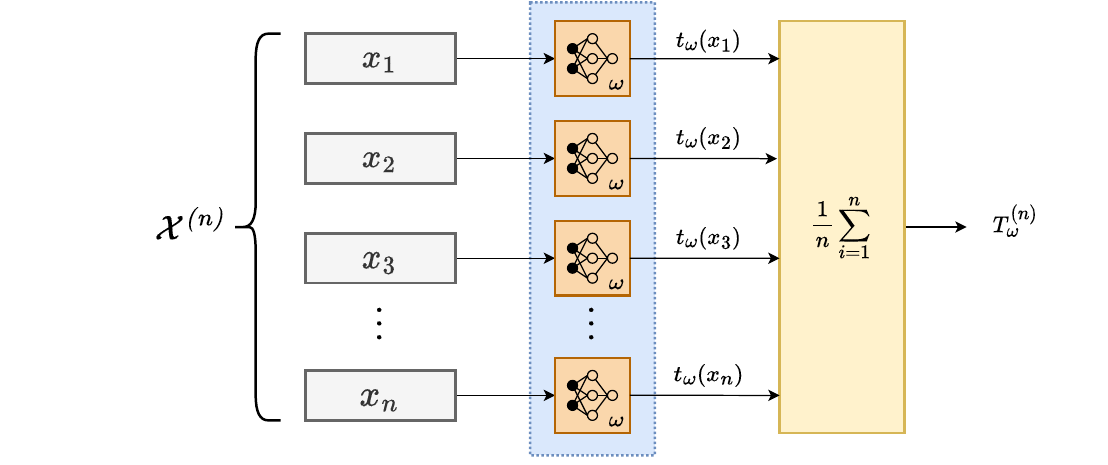}
    \caption{Conceptual diagram of the D3F $T^{(n)}_{\bm{\omega}}$~(\ref{eq:Tn}) with IID observations. The elementwise D3F $t_{\bm{\omega}} (x_j)$, $j=1,2,\dots,n$, is pictorially represented as the output of a NN whose parameters are $\bm{\omega}$ for each $j$.}
    \label{fig:d3f_arch_iid}
\end{figure}
 
A simple example of data-driven decision statistic is a NN with 2 layers, defined as~\cite{bishop-book}
\begin{equation}
    t_{\bm{\omega}} (x) = g \left( \sum_{j=0}^{M} \omega_{1j}^{(2)} h \left( \omega_{j1}^{(1)} x + \omega_{j0}^{(1)}\right) \right),
\end{equation}
where $M$ is the network width, while $g(\cdot)$ and $h(\cdot)$ are suitable nonlinearities~\cite{bishop-book}.  
The elementwise decision statistic $t_{\bm{\omega}}(x)$ is parameterized in $\bm{\omega}$, which can be learned (or estimated) by minimizing a suitable empirical loss function $\mathcal{E}$~\cite{bishop-book}
\begin{equation}
    \bm{\omega} = \arg \min_{\bm{\omega^\prime}} \sum_{j,k} {\cal E} \left(t_{\bm{\omega^\prime}} (y_{j,k})\right),   %
\end{equation}
where the summation follows from the independence assumption among the observations of the training set. 
In binary decision problems, a suitable loss function is the binary cross-entropy loss, defined as~\cite{bishop-book}
\begin{equation}
 {\cal E} \left(t_{\bm{\omega}} (y_{j,k})\right) = - k\, \log t_{\bm{\omega}}  (y_{j,k}) - (1-k)\log( 1-t_{\bm{\omega}} (x_j) ).
 \label{eq:cross-entropy}
\end{equation}
Alternative strategies to learn suitable elementwise D3F for binary detection are available in the recent literature; see, e.g.,~\cite{moustakides2019training,Veldhuis2021}. The choice of the loss function and decision statistic depend heavily on the specific problem. For generality, the following analysis is not restricted to particular choices. From now on, we will refer to $t_{\bm{\omega}}(x)$ as the elementwise D3F, which is a generic decision statistic function dependent on the parameter vector $\bm{\omega}$, learned from the dataset ${\cal Y}$.

Given the independence assumption and inspired by the optimal detector~\eqref{eq:Ln}, we assume the same structure as in~\eqref{eq:Ln}. The D3F is therefore constructed as follows (see also Fig.~\ref{fig:d3f_arch_iid}) 
\begin{mdframed}[backgroundcolor=black!20] 
\begin{equation}
    T^{(n)}_{\bm{\omega}} = \frac{1}{n} \sum_{i=1}^n t_{\bm{\omega}} (x_i).
    \label{eq:Tn}
\end{equation}
\end{mdframed}
 
In general, the D3F does not need to have an additive structure as in~\eqref{eq:Tn}. In principle, the structure of the D3F might even be learned during the training stage. For the case of \emph{dependent} observations discussed in Sec.~\ref{sec:LDP_images}, the dependency structure will be unknown, and consequently the D3F will not be forced to be a summation but will take the form of a DCNN.   

As mentioned in Sec.~\ref{sec:preliminaries}, we are going to exploit two asymptotic frameworks to study the classification performance of the D3F test; the first is based on the CLT~\cite{Lehmann-testing,Lehmann-large-sample} %
for setting an asymptotic level of false alarm or missed detection; the second is based on the LDP~\cite{Dembo-Zeitouni, hollander}.%

\subsection{Asymptotic properties based on the CLT}
\label{sec:CLT_IID}
Let us define $\mu_k$ and $\sigma_k$, respectively, as the %
mean and the standard deviation, under hypothesis ${\cal H}_k$, of the 
elementwise statistic, which is $l(x_i)$ in the case of the LLR and $t_{\bm{\omega}} (x_i)$ in the case of the D3F. 
In the case of the D3F, the aforementioned expectations are taken with respect to the observations ${\cal X}^{(n)}$, and the parameters $\bm{\omega}$ are fixed (they are learned during the training phase). Because of the additive nature of the decision statistics, we can invoke the CLT after normalizing the summation in~\eqref{eq:Ln} and~\eqref{eq:Tn} as follows~\cite{Lehmann-testing,Lehmann-large-sample}
\begin{align}
\widetilde{T}^{(n)}_k &= \sqrt{n}\, (T^{(n)} - \mu_{n,k}), \nonumber \\
\mu_{n,k} &= \mathbb{E}\left[ T^{(n)} \left| {\cal H}_k \right. \right] = \mu_k, \label{eq:CLT_0} \\
\sigma_{n,k} &= \mathbb{S} \left[ T^{(n)} \left| {\cal H}_k \right. \right] = \frac{\sigma_k}{\sqrt{n}}, \nonumber
\end{align}
where $k=0,1$ refers to the hypothesis ${\cal H}_k$, $T^{(n)}$ is provided by~\eqref{eq:Ln} or~\eqref{eq:Tn}, and $\mathbb{S} \left[ X \right]$ denotes the standard deviation of $X$. Assuming that both the mean and the variance are finite,  
then the CLT holds, %
and the normalized statistic $\widetilde{T}^{(n)}_k$ converges in distribution to a Gaussian random variable~\cite{Lehmann-testing,Lehmann-large-sample} 
\begin{equation}
    \widetilde{T}^{(n)}_k \xrightarrow[]{d} {\cal N} (0, \sigma_k^2), \quad \textnormal{under } {\cal H}_k,\,k=0,1.
    \label{eq:CLT}
\end{equation} 
In the case of the LLR, $T^{(n)} = L^{(n)}$, 
$\mu_1 = {\cal D} (f_1 || f_0) $ and $\mu_0 = - {\cal D} (f_0 || f_1) $, where ${\cal D} (p || q)$ is the Kullback-Leibler (KL) divergence between $p$ and $q$~\cite{Lehmann-testing}. Whereas, with the reference to the learned statistic, $T^{(n)} = T^{(n)}_{\bm{\omega}}$, and $\mu_k$ and $\sigma^2_k$ are the mean and the variance of $t_{\bm{\omega}}(x_i)$ under ${\cal H}_k$. It is worthwhile highlighting that we have assumed that the learning parameters $\bm{\omega}$ are fixed, or, in other words, that the convergence is statistically conditioned on the training set ($\bm{\omega}$ is fixed). 

\textit{Remark:} Under the summation structure in~(\ref{eq:Tn}) and given the independence assumption, the convergence shown in~(\ref{eq:CLT}) is well known in light of the CLT. However, as discussed in Sec.~\ref{sec:preliminaries}, under mild regularity conditions the CLT can derived as a consequence of the LDP~\cite{BRYC1993253}. Indeed, we will show, exploiting a numerical analysis, that such convergence seems to hold when the data are not independent and the D3F %
statistic %
does not have a summation structure but a more complicated function provided by a DCNN.

\subsection{Threshold setting}

Let us consider the test~(\ref{eq:test0}) and the error probabilities~(\ref{eq:error_prob}), where $ T^{(n)} = L^{(n)}$ for the LLR and $T^{(n)} = T^{(n)}_{\bm{\omega}}$ for the D3F. The threshold $\gamma_n$ %
is usually chosen to %
operate at a desired false alarm probability level. Without knowledge of the distribution under ${\cal H}_0$, the threshold can only be set using the available characterization set. %
Exploiting the CLT convergence~(\ref{eq:CLT}), we can set the asymptotic false alarm probability to a given value $0< \alpha<1$, when $n$ is large enough, as follows:
\begin{equation}
\gamma_n  = \mu_{n,0} + \sigma_{n,0} Q^{-1}(\alpha), \label{eq:alpha_CLT}    
\end{equation}
where $\mu_{n,0}$ and $\sigma_{n,0}$ are defined in~(\ref{eq:CLT_0}) and $Q^{-1}(\cdot)$ is the inverse of the $Q$-function, defined as $Q(x) = \frac{1}{\sqrt{2\pi}}\int_x^\infty\exp(-\frac{u^2}{2})du$. In fact, according to the CLT convergence in~(\ref{eq:CLT}) and the threshold in~\eqref{eq:alpha_CLT}, we have that
\begin{equation}
\lim_{n\rightarrow \infty} \alpha_n = \alpha.
\end{equation}
We assume to know both $\mu_{n,0} = \mu_0$ and $\sigma_{n,0} = \sigma_0/\sqrt{n}$, where $\mu_0$ and $\sigma_0$ can be estimated from the characterization set.\footnote{If the threshold is set with~\eqref{eq:alpha_CLT}, the miss detection probability of the LLR test vanishes exponentially fast with the best possible exponent, which is given by the KL divergence, see Stein's Lemma~\cite{Dembo-Zeitouni}. This behaviour is represented in Fig.~\ref{fig:laplace_iid_rate_distortion} that will be discussed later.}  

The expression of the threshold~(\ref{eq:alpha_CLT}) is therefore the summation of two terms: $\mu_{n,0}$ and $\sigma_{n,0} Q^{-1}(\alpha) = O(1/\sqrt{n})$. The second term is negligible in terms of error rate under ${\cal H}_1$, but it is instrumental to ensure that the asymptotic false alarm level is $\alpha$. 
An alternative choice for setting the threshold is $\gamma_n = \gamma$, with $\mu_0< \gamma <\mu_1$. For the LLR test, this choice would cause both the error probabilities to vanish exponentially. Based on the theory of large deviations~\cite{Dembo-Zeitouni,hollander}, our goal is to study the conditions for the D3F to exhibit exponentially %
vanishing error probabilities in $n$, and compute the convergence rate of the test. To this end, in the next section we will still assume independence of the observations, but we will then remove this assumption in Sec.~\ref{sec:LDP_dependent}.

\section{Large deviations of the D3F statistical hypothesis testing with IID observations}
\label{sec:LDP}

Large deviations analysis of statistical hypothesis testing is well established in the statistics literature (see, e.g.,~\cite{Dembo-Zeitouni, hollander}), and employed in %
several applications~\cite{varadhan1984large}; for instance, in sensor networks based on the running consensus paradigm~\cite{asymptotic-rc,braca2011consensus}. The assumption of IID observations is quite common in the sensor network literature and useful to establish theoretical results. The large deviations convergences are assessed in~\cite{bajovic2011distributed,jakovetic2012distributed,li2015distributed}, whereas in adaptive networks~\cite{sayed2014adaptive} large deviations are demonstrated for the steady-state distribution (slow adaptation regime)~\cite{Matta_IT2016,Matta_2016}. 
In the following, %
the fundamental theorems %
are reviewed and then applied to the test error probabilities~(\ref{eq:error_prob}).

\subsection{Large deviations principle}
Consider a sequence of IID random variables $z_1, z_2, \dots$, and their sample mean $S^{(n)} = \frac{1}{n} \sum_{i=1}^n z_i$. Cram\'er's theorem establishes the conditions for the sample mean to ``rarely'' deviate from its expected mean $\mu = \mathbb{E}[z_1]$. %
In the case that the probability of the event $\left[S^{(n)}\geq\gamma\right]$, with $\gamma>\mu>0$, vanishes exponentially with $n$, we say that $S^{(n)}$ obeys the LDP. The exponential rate of convergence is often referred to as the rate function, which depends on the threshold $\gamma$ and is provided by Cram\'er's theorem.

\vspace{5pt}
\begin{theorem}[Cram\'er's theorem~\cite{hollander}]
\label{Cramer}
consider a sequence of IID random variables $z_1, z_2, \dots$, with a LMGF
\begin{equation}
    \varphi(t) = \log \mathbb{E}\left[ \exp(t\,z_1) \right] < \infty \quad\quad \forall t \in \mathbb{R}. %
    \label{eq:LMGF_IID}
\end{equation}
Let $S^{(n)} = \frac{1}{n}\sum_{i=1}^n z_i$. Then, for all $\gamma>\mu$, 
\begin{equation}
    \lim_{n\rightarrow \infty} \frac{1}{n} \log \mathbb{P}\left[S^{(n)} \geq \gamma \right] = - I(\gamma),
    \label{eq:cramer}
    \end{equation}
    where $I(x)$ is the rate function, given by the Fenchel-Legendre transform of the LMGF, i.e.,
    \begin{equation}
        I(x) = \sup_{t \in \mathbb{R}} \left[x\,t - \varphi(t) \right].
        \label{eq:Fenchel-Legendre_IID}
    \end{equation}
\end{theorem}
\vspace{5pt}

We will not report here the proof of Cram\'er's theorem, for which we refer the reader to~\cite{hollander}, but it is worth mentioning that it is based on the \emph{squeeze theorem}, where the tightest Chernoff bound represents the upper bound. Such an upper bound is easily derived. Let us assume that $\mu=0$, $\gamma>0$, and consider the Chernoff bound on $S^{(n)}$, i.e.,
\begin{equation}
    \mathbb{P}\left[ S^{(n)} \geq \gamma\right] = \mathbb{P}\left[ \sum_{i=1}^n z_i \geq n\gamma\right] \leq \frac{\mathbb{E}\left[\exp\left(t z_1\right)\right]^n}{\exp\left(nt\gamma\right)},
\end{equation}
for $t\geq 0$. By applying the logarithm to both sides of the previous equation, we obtain
\begin{equation}
    \log \mathbb{P}\left[ \sum_{i=1}^n z_i \geq n\gamma\right] \leq n \log \mathbb{E}\left[\exp\left(t z_1\right)\right] - nt\gamma.
\end{equation}
Since the bound holds for every $t\geq 0$, we can rewrite the previous equation as
\begin{align}
    \frac{1}{n} \log \mathbb{P}\left[ S^{(n)}  \geq \gamma\right] &\leq - \sup_{t\geq 0} \left(\gamma t - \overbrace{\log \mathbb{E}\left[\exp\left(t z_1\right)\right]}^{\varphi(t)}\right), \nonumber  \\ %
    & = - \underbrace{\sup_{t \in \mathbb{R}} \left(\gamma t - \varphi(t)\right)}_{I(\gamma)}, \nonumber
\end{align}
where the last inequality stems from the fact that $\gamma t - \phi(t)$ is a concave function whose derivative at $t=0$ equals $\gamma - \mu > 0$. %

\textit{Remark:} In the general formulation of the LDP~\cite{hollander,Dembo-Zeitouni,Matta_IT2016} the convergence rate associated to an event $\left[ S^{(n)} \in A \right]$ is given by the infimum point of the Fenchel-Legendre transform within the region $A$, namely $\inf_{x\in A} I(x)$. In the case of Cram\'er's theorem, we have $A=[\gamma,\infty)$. %
Let us observe that the unconstrained minimum of $I(x)$ is attained at $x=\mu$ and is clearly null, meaning that $I(\mu) = 0$. As a consequence, being $I(x)$ strictly convex (see the discussion in~\cite{hollander}), if $\gamma \geq \mu$, then $\inf_{x \geq \gamma} I(x)=I(\gamma)$.

The application of Cram\'er's theorem will provide the error rate functions of the error probabilities in~(\ref{eq:error_prob}).

\subsection{LDP of the LLR test}

We report the error rate function of the LLR test in the following theorem~\cite{Dembo-Zeitouni,hollander} for continuous probability distributions. For discrete probability distributions, the interested reader 
is referred to Blahut~\cite{blahut1974hypothesis}. %

\vspace{5pt}
\begin{theorem}[Optimal error rate function]
\label{LLR_rate}
Let $\gamma \in (\mu_0, \mu_1)$, with $\mu_0 = - {\cal D} (f_0 || f_1)$ and $\mu_1 = {\cal D} (f_1 || f_0)$ being the KL divergences introduced in the previous section, then the error probabilities of the LLR test in~(\ref{eq:test0}) with $T^{(n)}$ given by~(\ref{eq:Ln}) and $\gamma_n = \gamma$, exhibit the following error rate functions
\begin{equation}
     \lim_{n\rightarrow \infty} \frac{1}{n} \log \alpha_n= - I_0(\gamma), \quad \lim_{n\rightarrow \infty} \frac{1}{n} \log \beta_n = - I_1(\gamma),
     \label{eq:LDP_LLR}
\end{equation}
where $I_k(\gamma)$ is the Fenchel-Legendre transform of the LMGF of the elementwise LLR $l(x)$ under ${\cal H}_k$.
\end{theorem}
\vspace{5pt}

It can be shown that the Fenchel-Legendre transform $I_1(\gamma)$ under ${\cal H}_1$ is given by a shift of $I_0(\gamma)$, namely $I_1(\gamma) = I_0(\gamma)-\gamma$ (see details in~\cite{Dembo-Zeitouni,hollander}).
Besides, according to Stein's Lemma (see details in~\cite{Dembo-Zeitouni}), forcing $\alpha_n<\epsilon$ the best exponent for $\beta_n$ is given by the KL divergence ${\cal D} (f_1 || f_0)$; this exponent can be achieved setting $\gamma_n$ as in~\eqref{eq:alpha_CLT}.

\subsection{LDP of the D3F test}

In this subsection, we provide the mathematical conditions that the D3F should satisfy in order to exhibit non-zero error rate functions. Specifically, let us observe that the D3F produces a transformation of the observed samples as follows 
\begin{equation}
    \tau_i =  t_{\bm{\omega}}(x_i),\quad \forall i=1,2,\dots,n.
\end{equation}
The LDP of the D3F test can be established applying Cram\'er's theorem to the sequence of IID random variables $\tau_i$. 

\vspace{5pt}
\begin{theorem}[Error rate function of the D3F test]
\label{D3F_rate}
Let $\gamma \in (\mu_0,\mu_1)$, with $\mu_k = \mathbb{E}[\tau_i | {\cal H}_k]$, with $k=0,1$, where the expectation is taken with respect to the measurements $x_i$. Then, the error probabilities of the D3F test~(\ref{eq:test0}) with $T^{(n)}$ given in~(\ref{eq:Tn}) and $\gamma_n = \gamma$ exhibit the following error rate functions\footnote{Note that there is a slight abuse of notation given that the D3F error rates~(\ref{eq:LDP_D3F}) are formally different from the LLR error rate~(\ref{eq:LDP_LLR}).}
\begin{equation}
     \lim_{n\rightarrow \infty} \frac{1}{n} \log\alpha_n= - I_0(\gamma), \quad \lim_{n\rightarrow \infty} \frac{1}{n} \log\beta_n = - I_1(\gamma),
     \label{eq:LDP_D3F}
\end{equation}
where $I_k(\gamma)$ is given by the Fenchel-Legendre transform of the LMGF of $\tau_i =  t_{\bm{\omega}}(x_i)$ under ${\cal H}_k$, with $k=0,1$. 
\end{theorem}
\vspace{5pt}

In other words, the performance of the %
test based on the D3F depends only on $I_k(\gamma)$, which is non-zero for $\mu_0 < \gamma < \mu_1$. Even if the LMGF is not available, we can estimate it from the characterization set and thus compute the Fenchel-Legendre transform (see %
Sec.~\ref{sec:estimation_rate}), which eventually leads to $I_k(\gamma)$. Moreover, resorting to the generalization of Cram\'er's theorem, namely the G\"{a}rtner-Ellis theorem~\cite{Dembo-Zeitouni,hollander}, which will be enunciated in the next section, it is not difficult to show that~\eqref{eq:cramer} holds true even if $\gamma$ is replaced by $\gamma_n$, provided that $\gamma_n \rightarrow \gamma$. Otherwise stated, if $\gamma_n \rightarrow \gamma$, then $\lim_{n\rightarrow \infty} \frac{1}{n} \log \mathbb{P}\left[S^{(n)} \geq \gamma_n \right] = - I(\gamma)$. As a consequence, considering $\gamma_n$ defined as in~(\ref{eq:alpha_CLT}), we have that $\gamma_n \rightarrow \mu_0$ and $\frac{1}{n} \log \alpha_n \rightarrow I_0(\mu_0) = 0$, which complies with the fact that $\alpha_n \rightarrow \alpha$ by construction. We expect then that $\beta_n$ vanishes exponentially, i.e., $I_1(\mu_0)>0$.

The rate functions describe the error probabilities’ scaling laws to zero; however, in practical applications it is important to have a good approximation of the entire error probability curves. This happens, for instance, when the asymptotic false alarm probability is controlled and fixed to a level $\alpha$; in this case, the error rate is exactly the same for all non-null values of $\alpha$, even if the curves themselves would be quite different. In the next subsection we will introduce a method to approximate the entire error probability curve thanks to the so-called ``exact asymptotics,'' that is closely related to the saddlepoint approximation.

\subsection{Exact asymptotics and saddlepoint approximation}
\label{sec:exact_asy}
Theorem~\ref{Bahadur_Rao}, provided in~\cite{Dembo-Zeitouni}, is important to approximate the error probability curves for finite sample-size regime of $n$.

\vspace{5pt}
\begin{theorem}[Bahadur and Rao]
\label{Bahadur_Rao}
Let $S^{(n)} = \frac{1}{n}\sum_{i=1}^{n} z_i$, where $z_i$ are IID real valued random variables with LMGF $\varphi(t) = \log \mathbb{E}[\exp(t\, z_1)]$. If the law of $z_1$ is non-lattice\footnote{The random variable $z_1$ has a lattice law if for some $z_0$, $d$, the random variable $d^{-1}(z_1 - z_0)$ is (a.s.) an integer number, and $d$ is the largest number with this property. The formulation of the theorem for lattice law is available in~\cite{Dembo-Zeitouni}, but not reported here for brevity.}, then
\begin{equation}
    \lim_{n\rightarrow\infty} J_n \mathbb{P}\left[S^{(n)} \geq \gamma \right] = 1,
\end{equation}
where $J_n=t_\gamma\sqrt{2\pi n \varphi^{\prime\prime}(t_\gamma)}\,\exp(n\,I(\gamma))$, $t_\gamma$ is the solution of the equation $\gamma = \varphi^\prime(t_\gamma)$, and $I(x)$ is the Fenchel-Legendre transform of the LMGF~(\ref{eq:Fenchel-Legendre_IID}).
\end{theorem}
\vspace{5pt}

We shall assume that $\gamma > \mu$, where $\mu$ is the expected mean of $z_i$; if this is not the case, as already discussed, the rate is zero. Then, we have the following approximation for the probability of the rare event $\left[S^{(n)} \geq \gamma\right]$:
\begin{mdframed}[backgroundcolor=black!20] 
\begin{align}
&\mathbb{P}\left[S^{(n)} \geq \gamma \right] \approx {\underbrace{\zeta_n}_{\substack{\text{sub-exponential} \\ \text{terms}}}} \,\, {\underbrace{\exp\left(-n\,I\left(\gamma\right)\right)}_{\mathclap{\text{exponential term}}}} \label{eq:exact_asy} \\
& \zeta_n = \left( t_\gamma\sqrt{2\pi n \varphi^{\prime\prime}(t_\gamma)} \right)^{-1}, \quad
t_\gamma:\, \varphi^\prime(t_\gamma)  = \gamma,  \nonumber 
\end{align}
\end{mdframed}
where $t_\gamma$ is the solution of the equation $\gamma = \varphi^\prime(t_\gamma)$, corresponding to the optimal solution associated with the Fenchel-Legendre transform of the LMGF~(\ref{eq:Fenchel-Legendre_IID}). The approximation~\eqref{eq:exact_asy} takes into account both the sub-exponential term and the exponential term provided by the LDP. The term $\zeta_n$ in~\eqref{eq:exact_asy} can be replaced by an asyptotically equivalent term $c_n$, with $c_n/\zeta_n \rightarrow 1$.
The term $c_n$ is defined as follows~\cite{Dembo-Zeitouni}
\begin{equation}
c_n = \int_0^\infty e^{-t}\left[Q(0)-Q\left(\frac{\zeta_n}{\sqrt{2\pi}}\,t\right)\right] dt,
\end{equation}
where $Q(\cdot)$ is the $Q$-function. 

An alternative derivation of~(\ref{eq:exact_asy}) can be obtained by the saddlepoint approximation (see Daniels' pioneering work in~\cite{daniels1954saddlepoint} and Reid's overview in~\cite{Reid88}), where the value $t_\gamma$ is often referred to as the saddlepoint~\cite{Reid88}. It is worthwhile noting that in both Theorem~\ref{Bahadur_Rao} and the saddlepoint approximation the higher order terms are neglected as not relevant asymptotically.

The saddlepoint approximation and the exact asymptotics~(\ref{eq:exact_asy}) do not need the IID assumption to provide a good approximation~\cite{wood1993saddlepoint}. Indeed, we exploit their general formulation to handle the case of dependent observations, detailed in the next section.\footnote{An alternative is the Lugannani-Rice formula (omitted for brevity); see, e.g.,~\cite{wood1993saddlepoint}. We have observed that, in the scenarios studied in this work, the Lugannani-Rice formula is numerically equivalent to~(\ref{eq:exact_asy}).}

Equation~(\ref{eq:exact_asy}) shows how $\alpha_n$ and $\beta_n$ in~(\ref{eq:alpha_n})-(\ref{eq:beta_n}) can be approximated in the finite sample-size regime of $n$. %
When the threshold is set as in~(\ref{eq:alpha_CLT}), the approximate error probability in~(\ref{eq:exact_asy}) can be further refined to take into account the decaying term $1/\sqrt{n}$ in the threshold~(\ref{eq:alpha_CLT}), contained in $\sigma_{n,0}$. A possible way to determine this refinement is to exploit the saddlepoint approximation at each value of $n$, which is valid in view of the ``non-asymptotic'' regime of the saddlepoint approximation~\cite{wood1993saddlepoint}. Basically, in~(\ref{eq:exact_asy}) we replace $I(\gamma)$ with $I_n(\gamma_n)$, which clearly converges to $I(\gamma)$ for $n$ large enough, and the related parameters, i.e., $t_{\gamma_n}$ and $\zeta_n$. This refinement is especially relevant in the case of dependent data, considered in the next section, and further discussed in Sec.~\ref{sec:estimation_rate}.

\section{Large deviations of the D3F statistical hypothesis testing in the general case via the G\"{a}rtner-Ellis theorem}
\label{sec:LDP_dependent}

Consider a generic sequence of random variables ${\cal X}^{(n)}$ and the sequence of decision statistics $T^{(n)} = T^{(n)}({\cal X}^{(n)})$. The generalization of Cram\'er's theorem is given by the G\"{a}rtner-Ellis theorem~\cite{Dembo-Zeitouni,hollander}, where $\varphi(t)$ is the limit of the scaled LMGF~\eqref{eq:LMGF_asy_start}. 
\vspace{5pt}
\begin{theorem}[G\"{a}rtner-Ellis theorem]
\label{GartnerEllis}
Consider a sequence random variables $z_1, z_2, \dots$, with an asymptotic scaled LMGF
\begin{equation}
    \varphi(t) = \lim_{n \rightarrow \infty} \frac{1}{n} \log \mathbb{E}\left[ \exp(n\,t\,z_n) \right] < \infty \quad\quad \forall t \in \mathbb{R}.
    \label{eq:LMGF_asy}
\end{equation}
If $\varphi(t)$ is differentiable in $\mathbb{R}$, then $z_n$ obeys the LDP  
\begin{equation}
    \lim_{n\rightarrow\infty} \log \mathbb{P}\left[z_n \in A \right] =  - \inf_{x\in A} I(x),
    \label{eq:LDP_asy}
\end{equation}
with rate function $I(x)$ given by the Fenchel-Legendre transform of $\varphi(t)$. %
\end{theorem}
\vspace{5pt}

An important property of Fenchel-Legendre transform holds when $\varphi(t)$ is differentiable, strictly convex, and diverging at $\pm \infty$ (see the discussion in~\cite{touchette2009large} and references therein). In this case, the Fenchel-Legendre transform reduces to the Legendre transform, see~\cite{touchette2009large}, and
there exists a unique root of $\varphi^\prime(t) = x$, which is denoted by $t_x$, leading to 
\begin{equation}
I(x) = t_x x - \varphi(t_x).     
\label{eq:dual_legendre}
\end{equation}
Therefore, in this case the slopes of $\varphi(t)$ are one-to-one related to the slopes of $I(x)$. This property is referred to as the duality property of the Legendre transform (see more details in~\cite{touchette2009large}) and is automatically verified in the context of IID observations in Sec.~\ref{sec:LDP}. %

In the case that $\varphi(t)$ is twice differentiable and is strictly convex ($\varphi^{\prime\prime}(t)>0$), then $I(x)$ must also be strictly convex, given that deriving~(\ref{eq:dual_legendre}) we have $I^{\prime\prime} (x) = \frac{1}{\varphi^{\prime\prime}(t_x)}$, and the curvature of $I(x)$ is the inverse curvature of $\varphi(t)$~\cite{touchette2009large}. Given that $I(x)$ is strictly convex, then the infimum in~(\ref{eq:LDP_asy}) can be easily handled as in Cram\'er's theorem. If $A=[\gamma, \infty)$, then the large deviations in~(\ref{eq:LDP_asy}) are given by $I(\gamma)$ for $\gamma$ larger than the asymptotic mean of the sequence $z_n$.

In the following we will exploit the G\"{a}rtner-Ellis theorem under both hypotheses for a D3F, which can take a generic structure, not necessarily a summation. In the first case of study, we consider data that are conditionally independent.

\subsection{Conditionally independent observations: Composite hypothesis testing}
\label{sec:LDP_conditionally_independent}

Let us assume ${\cal X}^{(n)} = \left(x_i\right)_{i=1}^n$ is a sequence of observations, and consider the case of a composite hypothesis under ${\cal H}_1$, where the observations are conditionally independent given a parameter $\theta \in \Theta$. The hypothesis under ${\cal H}_0$ is simple (clearly the generalization to a composite hypothesis is similar). Then, we have $x_i \sim f (\cdot | \theta)$ under ${\cal H}_1$ and $x_i \sim f (\cdot | \theta_0)$ under ${\cal H}_0$ with $\theta_0 \notin \Theta$.

The general framework of composite hypothesis testing is studied in~\cite{Dembo-Zeitouni}, where an universal hypothesis testing procedure is proposed, and studied in terms of large deviations. The popular Generalized Likelihood Ratio Test (GLRT) is investigated in terms of asymptotic optimality in \cite{Zeitouni92}, where conditions for asymptotic optimality of the GLRT in the Neyman-Pearson sense are studied and discussed. In our framework, where the hypotheses structure is unknown, we cannot rely on the GLRT approach or the universal detector, but for the practical needs of the machine learning procedure it is convenient to assume a Bayesian framework based on a given prior on the parameter $\theta$. Specifically, we adopt an embedded learning architecture to mimic the LLR in the aforementioned Bayesian framework. 
In this setting the LLR is given by the following:
\begin{align}
    L^{(n)} &= \frac{1}{n} \log \frac{f( {\cal X}^{(n)} | {\cal H}_1)}{f( {\cal X}^{(n)} | {\cal H}_0)} \nonumber \\ 
    &= \frac{1}{n} \log \frac{\sum_{\theta \in \Theta} w_\theta \prod_{i=1}^n f (x_i | \theta)}{\prod_{i=1}^n f (x_i | \theta_0)},
    \label{eq:conditional_Ln}
\end{align}
where we have assumed a finite support of $\theta$, namely $|\Theta|<\infty$, and a non trivial prior $w_\theta>0$. Clearly, the case of an infinite support of $\theta$ can also be handled making further regularity assumptions on the prior. In the case that the parameter is continuous, the summation is replaced by an integral. 

For $n$ sufficiently large, under ${\cal H}_1$ and given the true value of $\theta$, indicated with $\theta_*$, the mixture distribution in the numerator of~(\ref{eq:conditional_Ln}) is ``close,'' in terms of KL divergence, to the distribution of the observations (see also the discussion in~\cite{barron1987bayes,clarke1990information}). 

At this point, to handle the LDP, we need to compute the limit of the scaled LMGF~(\ref{eq:LMGF_asy}) of the LLR under both the hypotheses. Specifically, we assume that under ${\cal H}_1$ all the data are drawn from $f (\cdot | \theta_*)$, where $\theta_*$ indicates the true value of $\theta$, while under ${\cal H}_0$ all the data are drawn from $f (\cdot | \theta_0)$. Under some regularity assumptions, the limit of the scaled LMGF of the LLR~\eqref{eq:conditional_Ln} is then given by:
\begin{mdframed}[backgroundcolor=black!20] 
\begin{align}
     \varphi_k(t) & = \lim_{n \rightarrow \infty} \frac{1}{n} \log \mathbb{E}\left[ \exp(n\,t\,L^{(n)}) \left| {\cal H}_k \right.\right]
    \label{eq:LLR_scaled_LMGF} \\ & = \left\{\begin{array}{cc}
      \log \mathbb{E}\left[ \exp\left( t\,\log \frac{f (x_1 | \theta_*)}{f (x_1 | \theta_0)}\right)\right]    &  \textnormal{under } {\cal H}_1, \vspace{5pt}\\
      \log \mathbb{E}\left[ \exp\left( t\,\log \frac{f (x_1 | \theta_m)}{f (x_1 | \theta_0)}\right)\right]    &  \textnormal{under } {\cal H}_0, \nonumber
     \end{array}\right. 
\end{align}
\end{mdframed}
where the expectation in $\varphi_k(t)$ is taken assuming that the data are generated according to $f(\cdot|\theta)$ with $\theta = \theta_*\in\Theta$ under ${\cal H}_1$ and $\theta = \theta_0$ under ${\cal H}_0$; $\theta_m$ is the closest value of $\theta$ to $\theta_0$ in terms of the KL divergence. The derivation of the aforementioned formula is not, as far as we can tell, available in the literature, so we report it in Appendix~\ref{sec:appendix_lmgf}. 

In~\eqref{eq:LLR_scaled_LMGF} the prior distribution $w_\theta$ is absent, which is in agreement with classic Bayesian formulations (see, e.g.,~\cite{Lehmann}), where the prior does not affect the asymptotic behaviour.       

\begin{figure*}
    \centering%
    \includegraphics[width=.75\textwidth, trim=45 0 20 0]{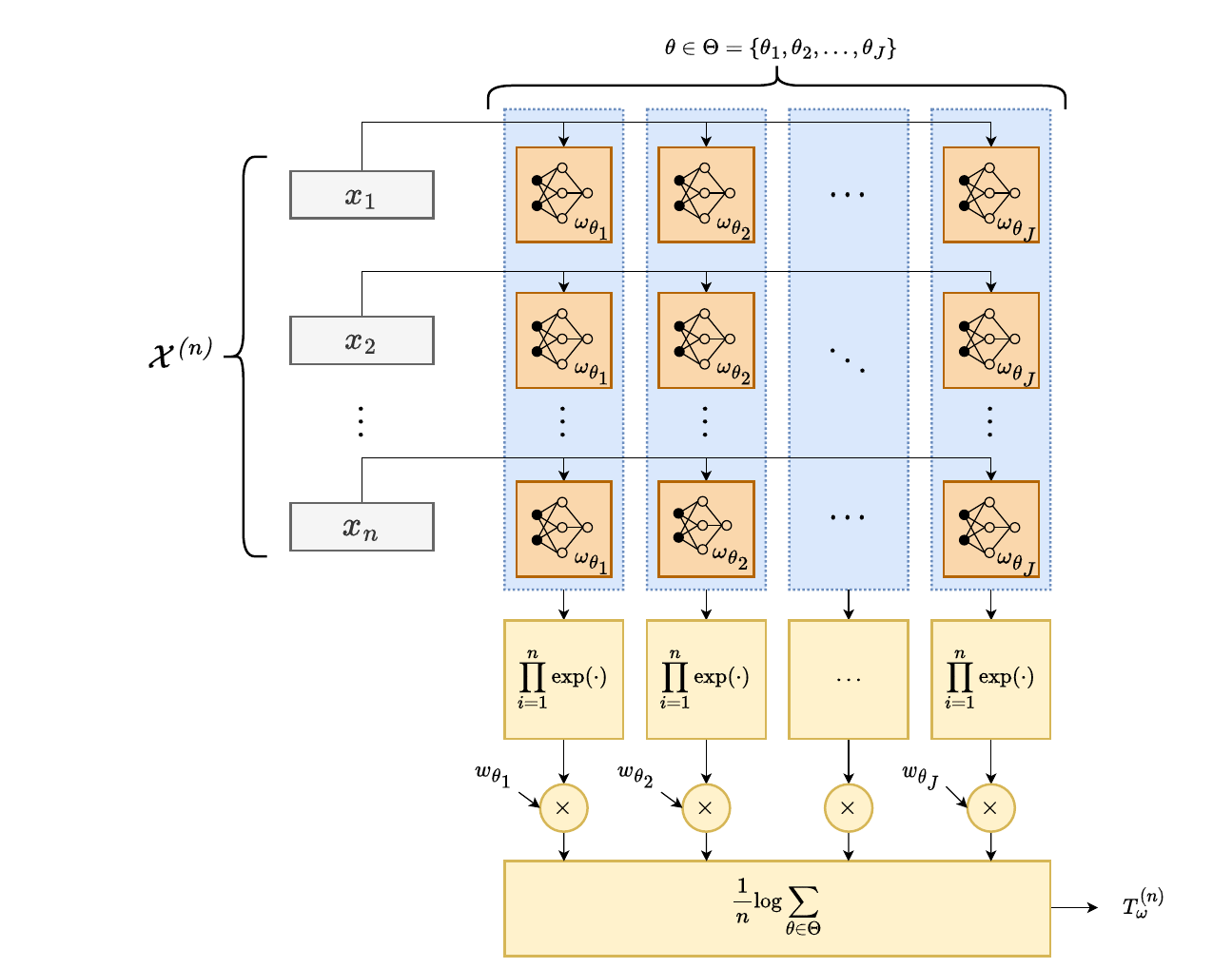}
    \caption{Conceptual diagram of the D3F $T^{(n)}_{\bm{\omega}}$ with conditionally independent observations, see~(\ref{eq:conditional_Tn}). The elementwise D3F $t_{{\bm \omega}_{\theta}}(x,\theta)$, conditioned to a specific value of $\theta$, is pictorially represented as a NN.}
    \label{fig:d3f_arch_dep}
\end{figure*}

Let us introduce the D3F for conditionally independent observations. Assume that the elementwise D3F $t_{{\bm \omega}_{\theta}}(x,\theta)$ is available (from the training stage), 
which approximates the elementwise LLR $\log \frac{f (x | \theta)}{f (x | \theta_0)}$ for a given value of the parameter $\theta$ (e.g., see Fig.~\ref{fig:decision_rules} for an illustration of the elementwise D3F in the case of a shift-in-mean problem with Laplace and Gaussian observations). As a consequence, the elementwise D3F $t_{{\bm \omega}_{\theta}}(x,\theta)$ can be derived as for the simple hypotheses, introduced in Sec.~\ref{sec:problem}, assuming that a specific value of $\theta$ is in force under ${\cal H}_1$. This means that the training stage will be repeated a number of times equal to %
the cardinality of $\Theta$, and there are as many training sets, one for each specific value of the parameter $\theta\in\Theta$.

Assuming a structure like that in~(\ref{eq:conditional_Ln}), the D3F counterpart is the following\footnote{From an implementation perspective, both the decision statistics~(\ref{eq:conditional_Ln}) and~(\ref{eq:conditional_Tn}) need to be computed with the well-known log-sum-exp trick~\cite[Sec.~3.5.3]{murphy2012machine} to avoid numerical underflow.}
\begin{mdframed}[backgroundcolor=black!20]
\begin{equation}
    T^{(n)}_{\bm \omega} = \frac{1}{n} \log \sum_{\theta \in \Theta} w_\theta \prod_{i=1}^n \exp\left(t_{{\bm \omega}_{\theta}}(x_i,\theta)\right),
    \label{eq:conditional_Tn}
\end{equation}
\end{mdframed}
where ${\bm \omega} = \left({\bm \omega}_{\theta}\right)_{\theta \in \Theta}$; the D3F structure is illustrated in Fig.~\ref{fig:d3f_arch_dep}. In~\eqref{eq:conditional_Tn}, the prior $w_\theta$ has to be estimated from the training set, or assumed uniform; in both the cases, the prior $w_\theta$ is not relevant from an asymptotic point of view. 

\begin{figure*}
    \centering%
    \includegraphics[width=0.9\textwidth, trim=70 280 100 140 ]{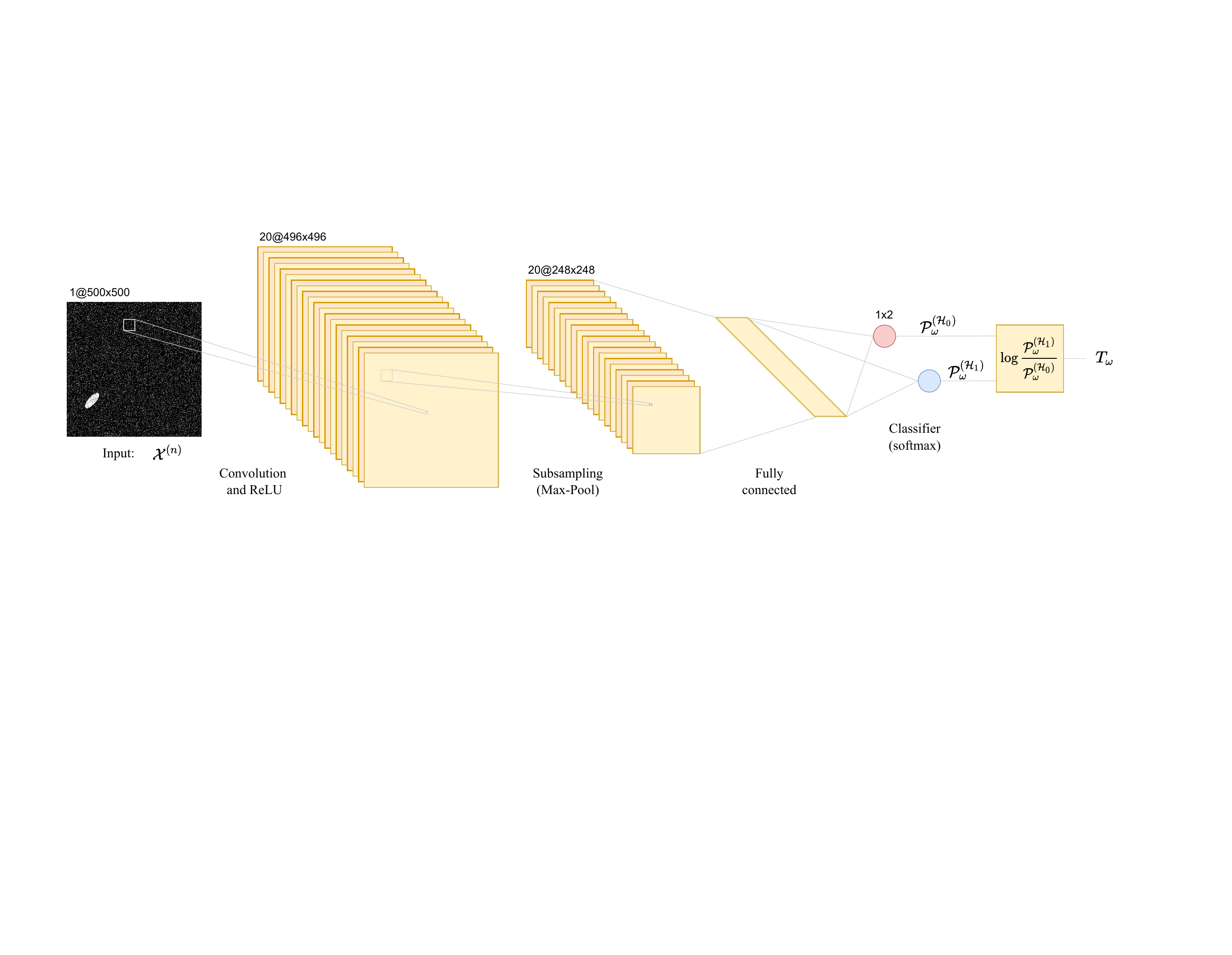}
    \caption{Illustration of the employed CNN model. The architecture comprises three layers: one convolutional layer, one pooling layer, and one fully-connected classification layer. The convolutional layer uses a $5\times 5$ convolutional kernel and has $20$ neurons; the pooling layer uses a $2\times 2$ pooling kernel. The last layer is a fully-connected classification layer with softmax activation function, whose output is the final $\mathcal{H}_0$ vs $\mathcal{H}_1$ classification.}
    \label{fig:d3fcnn_arch}
\end{figure*}

Given that for the D3F is constructed to have the same structure as the LLR, we can interrogate whether the scaled LMGF convergence holds similarly to~(\ref{eq:LLR_scaled_LMGF}). The key aspect to consider is that $R_n$, defined in~(\ref{eq:L_n_derivation}) in Appendix~\ref{sec:appendix_lmgf} and related to the LLR, vanishes for $n$ large enough. Under ${\cal H}_1$ the terms at the exponent in $R_n$ converge to the KL divergence between $\theta_*$ and $\theta$ when $\theta_*$ is the true parameter; then, $R_n$ vanishes exponentially fast with $n$. With a similar argument we can compute the term $R_n$ for the D3F, that is the following
\begin{equation}
    \sum_{\theta\in\Theta,\theta\neq\theta_*} \frac{w_\theta}{w_{\theta_*}} \exp \left[-n \left(\frac{1}{n} \sum_{i=1}^n t_{\omega_{\theta_*}}(x_i,\theta_*) - t_{\omega_{\theta}}(x_i,\theta) \right) \right].
    \label{eq:Rn_D3F}
\end{equation}
Given that the observations $x_i$ are IID we have that both the sample means at the exponent in~(\ref{eq:Rn_D3F}) converge a.s. to their expected values:
\begin{equation}
    \frac{1}{n} \sum_{i=1}^n t_{\omega_{\theta}}(x_i,\theta) \xrightarrow[]{a.s.} \mathbb{E}\left[t_{\omega_{\theta}}(x,\theta) | \theta_* \right] = \mu(\theta|\theta_*).
\end{equation}
If the expected value of the D3F under $\theta_*$ is larger than the expected values in $\theta \neq \theta_*$, namely $\mu(\theta_*|\theta_*) > \mu(\theta|\theta_*)$, then $R_n$ vanishes to zero and the convergence is similar to that in~(\ref{eq:LLR_scaled_LMGF}). %
It is worthwhile to note that such a condition is not necessary to have a large deviations result for the D3F, but it is useful to study the scaled LMGF convergence, and consequently the test performance.

\emph{Remark:} Another valid learning strategy could be to train the D3F to distinguish not only any given parameter $\theta$ from $\theta_0$, but also to distinguish $\theta \in \Theta$ from any other value $\widetilde{\theta} \in \Theta$, with $\theta \neq \widetilde{\theta}$. A possible training strategy would be resorting to the multi-class cross-entropy cost function; see, e.g.,~\cite{bishop-book}. 

\subsection{Dependent observations: Target detection in binary images}
\label{sec:LDP_images}

In this subsection we describe a statistical decision problem, where the observations ${\cal X}^{(n)} = \left(x_j \right)_{j=1}^{N_c}$ are dependent, and such dependence cannot be easily modeled. We consider the problem of deciding if a target is present or absent in the image ${\cal X}^{(n)}$, where $n\leq N_c$ represents the size of the target when present, and $N_c$ is the size of the image. It is worthwhile to note that, different from the previous scenarios, the size of the data ${\cal X}^{(n)}$ is $N_c$ and not $n$.\footnote{The parameter $n$ is not anymore the size of the data, but it has the same role as before; it is just quantified in a different way. The key idea is that $n$ still rules the detection performance, but the rate function changes to take into account the redundancy introduced by the dependence.} 
The data $x_i$, $i=1,\dots,N_c$, are represented by the pixels of the image %
(or, in the radar/sonar context, resolution cells), where the index $i$ represents the $i$th pixel. We indicate with ${\cal H}_1$ the target-present hypothesis and with ${\cal H}_0$ the target-absent one. The target can be \emph{extended}, in the sense that it can occupy multiple cells; see, e.g.,~\cite{koch08,Granstrom15} in the context of Extended Target Tracking (ETT), and~\cite{Melo_2020,williams2020toward,Williams_2021} in the context of underwater object classification. 

To fix ideas, we provide here a statistical model for the problem of deciding the presence/absence of an extended target, and we stress that this is an example and not intended to delimit a general use case. Let us assume that, if the target is present in the $i$th pixel of the image, $\theta_i=1$, otherwise $\theta_i=0$.
We assume that each pixel provides a binary observation, $x_i \in \left\{0,1 \right\}$ according to a Bernoulli distribution with success probability $p_1$ if the target occupies the $i$th pixel, i.e., $\theta_i=1$, or with success probability $p_0$ if $\theta_i=0$. We have basically assumed that the image is the output of a preliminary detection stage performed at the pixel level, which is typical in radar/sonar processing. The probabilities $p_1$ and $p_0$ are respectively the pixel-wise detection probability and false alarm probability. The classic notion of SNR would describe the relation between $p_1$ and $p_0$.  
Clearly, another option would be to formalize the pixel observation with a continuous distribution, assuming a suitable clutter and target modelling. Let us indicate $\Theta^{(n)} = \left(\theta_i\right)_{i=1}^{N_c}$, where $n$ is given by the size of the target, namely $n = \sum_{i=1}^{N_c} \theta_i$. We will be obviously under the ${\cal H}_1$ hypothesis, since $n\geq 1$. Let us also assume for simplicity that, under ${\cal H}_1$ the data are conditionally independent given the target position and shape, i.e., $\Theta^{(n)}$. Otherwise stated, the distribution of the data is given by 
\begin{IEEEeqnarray}{rCl}
f_1\left({\cal X}^{(n)} \left| \Theta^{(n)} \right.\right) &=& \prod_{i, \theta_i=1} p_1^{x_i} (1-p_1)^{1-{x_i}} \nonumber \\
&&\times\> \prod_{i, 
\theta_i=0} p_0^{x_i} (1-p_0)^{1-{x_i}}.
\label{eq:dep_H1}
\end{IEEEeqnarray}
Analogously, the distribution of the data under ${\cal H}_0$ is given by
\begin{IEEEeqnarray}{rCl}
f_0\left({\cal X}^{(n)}\right) = \prod_{i} p_0^{x_i} (1-p_0)^{1-{x_i}}.
\label{eq:dep_H0}
\end{IEEEeqnarray}
Note that the $\mathcal{H}_0$ hypothesis is simple, i.e., independent on $n$ because of the absence of the target. 

If we assume perfect knowledge about the target position and shape, it is easy to verify that the LLR rule is equivalent to a test between two sequences of Bernoulli random variables of length $n$. In this scenario, the aim is to generalize the detection to the case of \textit{unknown} target position and shape. In principle, one could always implement the LLR as in~\eqref{eq:conditional_Ln}, obtaining   
\begin{equation}
    L^{(n)} = \frac{1}{n} \log \frac{\sum_{\Theta^{(n)} \in {\cal T}^{(n)}} w_{\Theta^{(n)}} f_1\left({\cal X}^{(n)} \left| \Theta^{(n)} \right.\right)}{f_0\left({\cal X}^{(n)}\right)},
    \label{eq:conditional_Ln_2}
\end{equation}
where ${\cal T}^{(n)}$ and $w_{\Theta^{(n)}}$ are respectively the space and the (discrete) distribution of all target positions and shapes under consideration. Modeling such space and distribution is clearly infeasible in a real-world application, especially if we consider that, in principle, the target can have any possible shape. A convenient choice for the tractability of the problem that is usually made in the ETT literature %
is to model the target as an ellipsoid, with the target shape posterior in some cases given by an Inverse Wishart distribution (e.g., see~\cite{koch08, Granstrom15,Vivone16} and references therein). 
In an ETT problem, %
the relevant parameters of the target, such as its position, velocity, and shape %
are estimated sequentially. Although the ellipsoid assumption is quite convenient, it is also somehow limiting. %

Different from the ETT literature, we focus primarily on the target detection task, where the D3F decision statistic is based on a DCNN, whose architecture is represented in Fig.~\ref{fig:d3fcnn_arch}. The DCNN is a proper choice to capitalize the dependency structure embedded in the input data, which can be learned during the training phase.

We generate the synthetic training sets ${\cal Y}_k$ 
under ${\cal H}_k$, $k=0,1$, each composed by $m_y$ images. Under ${\cal H}_1$, The images in ${\cal Y}_k$ are generated according to~(\ref{eq:dep_H1})-(\ref{eq:dep_H0}), with varying target positions, shapes and orientations; thus, the size of the target $n_j$ varies for $j=1,\dots,m_y$. Under ${\cal H}_0$, only false alarms are randomly generated. %
The training set does not include any explicit information about the actual size, shape and position of the target; therefore the DCNN-based D3F does not process directly $n$ or $\Theta^{(n)}$. 

The D3F decision statistic is defined as
\begin{mdframed}[backgroundcolor=black!20] 
\begin{equation}
    T_{\bm{\omega}} \left( {\cal X}^{(n)} \right) = 
    \log \frac{{\cal P}_{\bm{\omega}}^{({\cal H}_1)}({\cal X}^{(n)})}{{\cal P}_{\bm{\omega}}^{({\cal H}_0)}({\cal X}^{(n)})},
    \label{eq:D3F_dep}
\end{equation}
\end{mdframed}
and is the log-ratio of the softmax classifier probabilities computed by the DCNN, i.e., ${\cal P}_{\bm{\omega}}^{({\cal H}_k)}({\cal X}^{(n)})$ of hypothesis ${\cal H}_k$, $k=0,1$, with ${\cal P}_{\bm{\omega}}^{({\cal H}_0)}({\cal X}^{(n)}) + {\cal P}_{\bm{\omega}}^{({\cal H}_1)}({\cal X}^{(n)}) = 1$. The $\bm{\omega}$ vector represents the network parameters learned in the training phase. The D3F $T_{\bm{\omega}}$ is the natural generalization of~(\ref{eq:Tn}) for IID observations, and plays the role of the LLR~\eqref{eq:conditional_Ln_2}.

\subsubsection{Asymptotic framework}
The model~(\ref{eq:dep_H1})-(\ref{eq:dep_H0}) assumes that the size of the input data is larger than the target, i.e., $n\leq N_c$. Apparently, the limit for $n$ that diverges demands that an infinite input sequence is available. %
In this respect, it is implicitly assumed that the DCNN can process increasingly large images ($N_c \rightarrow \infty$). From a practical perspective, this amounts to assuming that the size number of pixels in the image $N_c$ is always sufficiently larger than the size of the target $n$.

The distribution of the input data under ${\cal H}_0$, and under ${\cal H}_1$ conditioned on the target $\Theta^{(n)}$, remains a sequence of independent Bernoulli random variables. As in Sec.~\ref{sec:LDP_conditionally_independent}, both the detection performance and the asymptotic scaled LMGF shall be conditioned to a realization of the parameter under ${\cal H}_1$. In the current scenario, we shall fix a shape for the target, for instance a circle, and a growing rule when $n$ increases. For instance, the growing rule could be that the radius $r_{\Theta}$ of the circle diverges, with $n$ that therefore also diverges (quadratically) with the radius. 

Then, we consider the sequence of random variables $T^{(n)}_{\bm{\omega}}$ given by the normalized D3F decision statistic 
\begin{equation}
T^{(n)}_{\bm{\omega}} = \frac{1}{n} T_{\bm{\omega}} \left( {\cal X}^{(n)} \right),
\label{eq:D3F_dep_Tn}
\end{equation}
where $n = \sum_{i \geq 1} \theta_i$ increases, and the randomness of the observations ${\cal X}^{(n)}$ is given by the Bernoulli realizations in~(\ref{eq:dep_H1})-(\ref{eq:dep_H0}). Having fixed the shape of the target, we can generate the characterization set for each value of $n$ that is of interest, and then we can numerically compute the asymptotic scaled LMGF and the relevant parameters of the exact asymptotics to construct the approximate error probability curves~(\ref{eq:error_prob}). More details are given in Sec.~\ref{sec:estimation_rate}.

\begin{figure*}
    \captionsetup[subfigure]{labelformat=empty}%
    \centering%
    \subfloat[][]{%
      \includegraphics[width=\columnwidth]{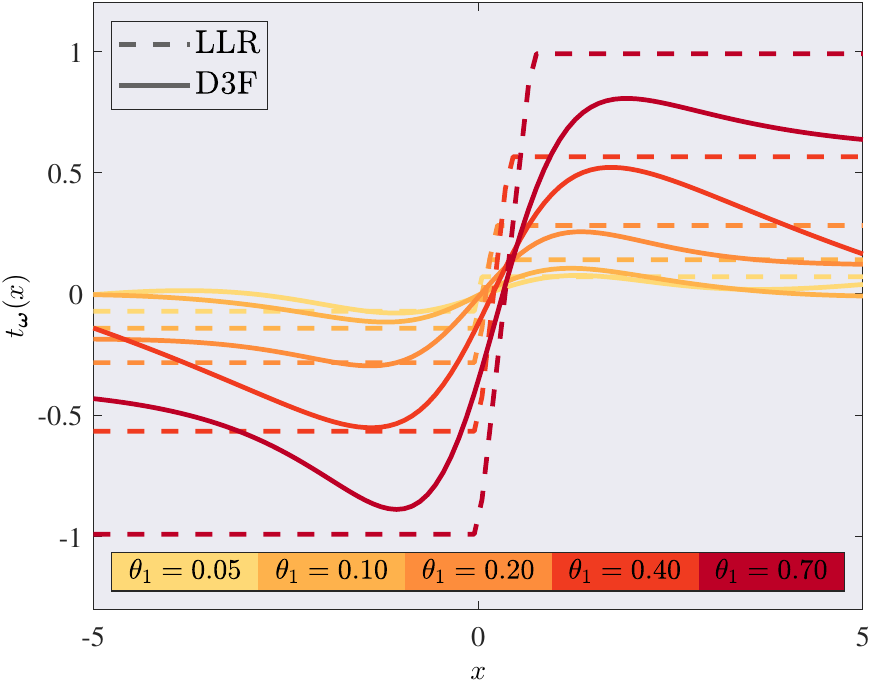}
      \label{fig:decision_rule_gaussian}
        }%
    \hfil %
    \subfloat[][]{%
      \includegraphics[width=\columnwidth]{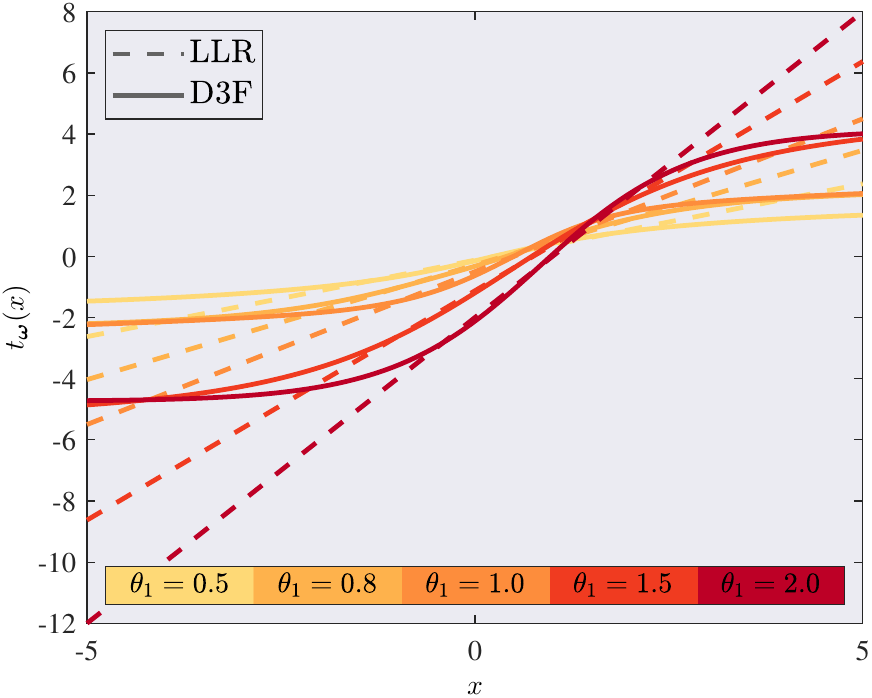}
        \label{fig:decision_rule_laplace}
        }%
    \vspace{-1em}
    \caption{Elementwise D3F, defined as 2~layers NN trained with $m=10^4$ independent samples, distributed according to a Laplace (left) and Gaussian (right) with $\theta_0=0$ under ${\cal H}_0$ and different values of $\theta_1$ under ${\cal H}_1$. The LLR function is also reported for the different combinations of $\theta_0$ and $\theta_1$.}
    \label{fig:decision_rules}
\end{figure*}

\subsection{Asymptotic normality and threshold setting}
\label{sec:CLT_Dep}

Assuming that the limit~(\ref{eq:LMGF_asy}) exists, the decision statistic is then asymptotically normal around the asymptotic expected value, given by $\mu_k=\varphi_k^\prime(0)$; see~\eqref{eq:mu_k_sigma_k} and the discussion in Sec.~\ref{sec:preliminaries}. Specifically, 
similarly to~(\ref{eq:CLT_0}) we can define the normalized statistic under ${\cal H}_k$:
\begin{align}
\widetilde{T}^{(n)}_k &= \sqrt{n}\, (T^{(n)} - \mu_{n,k}), \label{eq:CLT_general} \\
\label{eq:CLT_dep}
\mu_{n,k} &= \mathbb{E}\left[ T^{(n)} \left| {\cal H}_k \right. \right], \nonumber \\
\sigma_{n,k} &= \mathbb{S} \left[ T^{(n)} \left| {\cal H}_k \right. \right],  \nonumber 
\end{align}
where we indicate with $T^{(n)}$ the LLR~\eqref{eq:conditional_Ln}, or the D3F statistics~\eqref{eq:conditional_Tn} and~\eqref{eq:D3F_dep_Tn}. For $n$ sufficiently large, $\mu_{n,k} \rightarrow \mu_k$ and $\sqrt{n}\, \sigma_{n,k} \rightarrow \sigma_k = \sqrt{\varphi^{\prime\prime}_k(0)}$, and by virtue of the results in~\cite{BRYC1993253}, we have that $\widetilde{T}^{(n)}_k$ converges in distribution to ${\cal N} (0, \sigma_k^2)$ under ${\cal H}_k$.\footnote{Note that in the problem of extended target detection in previous section, the statistic under ${\cal H}_0$ does not depend on $\Theta^{(n)}$, and thus $n$.} 
As in~(\ref{eq:alpha_CLT}), we can then set the asymptotic false alarm probability $\alpha =\lim_{n\rightarrow \infty} \alpha_n$ as follows
\begin{equation}
\gamma_n = \mu_{n,0} + \sigma_{n,0} Q^{-1}(\alpha), \label{eq:alpha_CLT_general}
\end{equation}
where $\mu_{n,0}$ and $\sigma_{n,0}$ are given in~(\ref{eq:CLT_general}). 

\subsubsection{Conditionally independent observations}
To grasp further inside about the asymptotic normality of the normalized statistic in~(\ref{eq:CLT_general}), let us consider the scenario with conditionally independent observations, discussed in Sec.~\ref{sec:LDP_conditionally_independent}. Let us begin with $T^{(n)}$ given by the LLR in~(\ref{eq:conditional_Ln}). The convergence can be established considering that the term $R_n$ in~(\ref{eq:L_n_derivation}) vanishes in probability. Specifically, under ${\cal H}_1$ and with a derivation similar to that reported in Appendix~\ref{sec:appendix_lmgf} [see~(\ref{eq:L_n_derivation})], we can show that the LLR is given by the %
clairvoyant LLR (i.e., that has knowledge of $\theta_*$, which is the true value of $\theta$) plus a remainder term, referred to as $\widetilde{R}_n$, i.e.,
\begin{equation}
    L^{(n)} = L^{(n)}_{\theta_*} + \widetilde{R}_n, 
\end{equation} 
with $\widetilde{R}_n = \frac{1}{n}\log(w_{\theta_*} (1+R_n))$ that converges to zero in probability (given that $R_n$ vanishes in probability; see details in Appendix~\ref{sec:appendix_lmgf}). By virtue of Slutsky's theorem~\cite{Lehmann-large-sample}, we have that $L^{(n)}$ converges in distribution to the limit distribution of the ideal LLR $L^{(n)}_{\theta_*} = \frac{1}{n} \log \prod_{i=1}^n \frac{f(x_i|\theta_*)}{f(x_i|\theta_0)}$, which is the LLR of IID observations distributed under ${\cal H}_1$ with parameter $\theta_*$. Thanks to the CLT, the normalized LLR of IID observations converges in distribution to a Gaussian distribution, as already expressed in~(\ref{eq:CLT}). Analogously, a similar convergence holds under ${\cal H}_0$; indeed, considering~(\ref{eq:L_n_derivation_H0}), we have that $L^{(n)}$ converges in distribution to the limit distribution of $L^{(n)}_{\theta_m} = \frac{1}{n} \log \prod_{i=1}^n \frac{f(x_i|\theta_m)}{f(x_i|\theta_0)} $, which is the LLR of IID observations distributed under ${\cal H}_0$ with the parameter $\theta_m$ which is the minimum KL divergence among $\theta \in \Theta$ between $f(x|\theta)$ and $f(x|\theta_0)$.   
Summarizing, we have that $\sqrt{n}(L^{(n)}-\mu_{n,k})$ converges in distribution to a zero-mean Gaussian distribution ${\cal N} (0,\sigma_k^2)$ under ${\cal H}_k$, with the expected values given by the following KL divergences 
\begin{equation}
\mu_{n,k} \rightarrow \mu_k = \left\{\begin{array}{cc}
      {\cal D}(f(x|\theta_*)||f(x|\theta_0))    &  k=1 \vspace{5pt}\\
      -{\cal D}(f(x|\theta_0)||f(x|\theta_m))    &  k=0, \nonumber
     \end{array}\right.      
\end{equation}
and $\sigma_k = \mathbb{S} \left[\log \frac{f(x|\theta) }{f(x|\theta_0) } | {\cal H}_k \right]$, with $\theta = \theta_*$ under ${\cal H}_1$ and $\theta = \theta_m$ under ${\cal H}_0$. %
Note that the convergence of $\mu_{n,k}$ is coherent with the convergence of the scaled LMGF~(\ref{eq:LLR_scaled_LMGF}); indeed, it is easy to verify that $\mu_k = \varphi^\prime_k(0)$ from the properties of the asymptotic scaled LMGF~\cite{touchette2009large}. %

For the convergence of the D3F~(\ref{eq:conditional_Tn}), we can proceed analogously, %
assuming that the D3F $R_n$ term vanishes in probability to zero. However, this latter condition is not strictly necessary to establish the aforementioned convergence, as reported in Sec.~\ref{sec:experiments}. In other words, the convergence of $R_n$ to zero is just a sufficient condition for the Gaussian convergence of the D3F.

\subsubsection{Target detection in binary images}
In the problem described in Sec.~\ref{sec:LDP_images}, the LLR is practically infeasible. However, it is worth mentioning that it would converge asymptotically in distribution to a Gaussian distribution %
because of the aforementioned %
conditional independence of observations.

Let us consider the sequence of decision statistic $T_{\bm{\omega}}^{(n)}$ defined in~(\ref{eq:D3F_dep_Tn}).
The DCNN training set used in this work is composed by different target shapes, positions, orientations, and size $n$. In other words, the DCNN (and the resulting D3F) is not dependent on the parameter $n$; this dependence is only induced through the input observations. Given that under ${\cal H}_0$ the observations contain only false alarms, then the normalized statistic~(\ref{eq:CLT_general}) is independent on $\Theta^{(n)}$. In other words, the normalized statistic~(\ref{eq:D3F_dep_Tn}), given by the D3F~\eqref{eq:D3F_dep} divided by $n$, is a.s. converging to zero.

However, we have observed empirically, as illustrated in Fig.~\ref{fig:histograms_extended_target} (where we plot the histograms of the normalized statistic~\eqref{eq:D3F_dep_Tn}), that the D3F~\eqref{eq:D3F_dep} is approximately Gaussian-distributed under ${\cal H}_0$; in other words:
\begin{equation}
T_{\bm{\omega}} \left( {\cal X}^{(n)} \right) \sim {\cal N} (\mu_0, \sigma_0^2),
\label{eq:gaussian_dep}
\end{equation}
where, as for the other scenarios, both parameters $\mu_0$ and $\sigma_0$ can be estimated from the characterization set. Thanks to~(\ref{eq:gaussian_dep}), the threshold can be again set as in~(\ref{eq:alpha_CLT_general}): 
\begin{equation}
\gamma = \mu_0 + \sigma_0 Q^{-1}(\alpha),   
\label{eq:gamma_image}
\end{equation}
where $\alpha$ is the desired false alarm rate. The normalized statistic~(\ref{eq:D3F_dep_Tn}) clearly vanishes under ${\cal H}_0$ for $n$ that diverges, and therefore the LDP does not hold under ${\cal H}_0$.
The LDP can be instead verified under ${\cal H}_1$. Indeed, we also observe empirically that the normalized decision statistic~(\ref{eq:D3F_dep_Tn}) converges to a Gaussian distribution, as illustrated in Fig.~\ref{fig:histograms_extended_target}.

\textit{Remark:} It is possible to justify that the D3F statistic~\eqref{eq:gaussian_dep} under ${\cal H}_0$ is approximately Gaussian assuming that the width of any layers in the network diverges; a similar argument can be found in~\cite{Neal1996,lee2018deep}. Specifically, at the first DCNN layer, the samples at the input of the different non-linearities (i.e., at the output of the convolutional layers) can be modeled, under some mild assumptions, as Gaussian random variables. This is thanks to the CLT for independent (but non-identically distributed)  samples~\cite{Lehmann-large-sample} and exploiting the IID assumption of ${\cal X}^{(n)}$ under ${\cal H}_0$. Then, we conjecture that this mechanism propagates among the layers. Specifically, we suppose that the CLT can be still exploited to model the samples at the input of the $(i+1)$\textsuperscript{st} non-linearity, if the $i$th layer's width is sufficiently large and %
assuming that the correlations among the $i$th layer's output samples are weak enough to invoke the CLT. 

It is worthwhile observing that the assumptions made in~\cite{Neal1996,lee2018deep} require the network weights being IID; conversely, here the weights are fixed and the network inputs are random. A further analysis of this convergence is left to future works, given that here we are instead focusing on the convergence in the parameter $n$.

\section{Rate function estimation and approximate error probability curves}
\label{sec:estimation_rate}

In the previous section we have established the main large deviations properties of the D3F test~\eqref{eq:test0}. In this section we make explicit how it is possible to compute rate functions, the approximate error probability curves, and the threshold parameters from the \emph{characterization set} ${\cal Z}$. %
In some circumstances an overlap between the characterization set and the training set is possible. For this reason, in Appendix~\ref{sec:appendixB} we show that the estimators based on the sample mean, which will be detailed in the next subsections, converges in probability to the proper quantities of interest. In other words, the independence between characterization set and training set is not strictly necessary.   

Before entering into details, let us consider that the characterization set depends on the specific parameter $\theta\in\Theta$ under consideration when the observations are conditionally independent (see the definition in Sec.~\ref{sec:LDP_conditionally_independent}), as well as on the extended target $\Theta^{(n)}$ under consideration, defined in Sec.~\ref{sec:LDP_images}.

\subsection{Rate function estimation}
We propose two procedures to estimate the rate function. The former is based on the \emph{direct estimation} of the LMFG, or its scaled version, and relies on solving numerically the Fenchel-Legendre transform thanks to a numerical representation of the (scaled) LMGF. The latter, which is more accurate, instead requires to generate data from the exponential tilting distribution via MCMs.
\subsubsection{Direct estimation approach}
A statistically \emph{consistent} estimator of the LMGF, i.e., $\log\mathbb{E}\left[ \exp(t\,\tau(x)) |  {\cal H}_k\right]$, %
for the first scenario (Sec.~\ref{sec:problem}) consists in substituting the expected value with the sample mean:
\begin{equation}
    \hat{\varphi}_k(t) = \log \frac{1}{m_z}\sum_{j = 1 }^{m_z} \exp(t\,\tau_j),
    \label{eq:LMGF_est}
\end{equation}
where $\tau_j = t_{\bm{\omega}}(z_j)$ is the elementwise D3F; the samples $z_j$ are taken from the characterization set ${\cal Z}$ and are assumed IID according to ${\cal H}_k$, $k=0,1$. Thanks to the law of large numbers, $\hat{\varphi}_k(t) \xrightarrow[]{a.s.} \varphi_k(t)$ for $m_z\rightarrow\infty$.\footnote{If the characterization set is overlapped with the training set, then the strong convergence should be replaced with the weak convergence, see details in Appendix~\ref{sec:appendixB}.} Then, it is possible to compute numerically the Fenchel-Legendre transform of the estimated LMGF $\hat{\varphi}_k(t)$, which leads to the rate function. The same approach can be adopted when we deal with the scaled LMGF in~(\ref{eq:LMGF_asy}), for different values of $n$: 
\begin{equation}
    \hat{\varphi}_{n,k}(t) = \frac{1}{n} \log \frac{1}{m_z}\sum_{j = 1 }^{m_z} \exp(n\,t\,T^{(n)}_j),
    \label{eq:scaled_LMGF_est}
\end{equation}
where we have again substituted the expected value with the sample mean and implicitly assumed the dependency on the parameter $\theta$ or $\Theta^{(n)}$. The asymptotic behaviour of interest should clearly be approximated with a sufficiently large value of $n$, provided that $m_z$ is also large enough to accommodate an accurate estimation of the scaled LMGF. For each value of $n$, the approximate scaled LMGF can be exploited to compute the saddlepoint, as in~(\ref{eq:exact_asy}), but with a rate function $I_n$ given by the Fenchel-Legendre transform of the estimated version of the scaled LMGF for a given $n$. This approach is valid in view of the non-asymptotic regime; see, e.g., the discussion in~\cite{wood1993saddlepoint}. We apply this strategy to compute the approximate error probability curves for the problem of deciding the presence/absence of an extended target in an image, discussed in Sec.~\ref{sec:LDP_images}.     

\subsubsection{Rate function estimation via the exponential change of measure, and sampling from the tilted distribution}
The direct estimation of the scaled LMGF can be inaccurate when $m_z$ is not large enough to accommodate possibly high values of $t$ and/or $n$. A valid alternative is use an importance sampling procedure to generate samples from the \emph{exponential tilted} distribution; such procedure is also referred to as exponential change of measure~\cite{touchette2011basic,asmussen2007stochastic,bucklew2004introduction}. The approach is general enough to be applied to the case of a dependent sequence of data $x_j$. The main drawback is that we need to generate samples according to the distribution of the input data, as required by the Metropolis-Hastings (MH) algorithm to sample data from the tilted distribution. As already discussed, the possibility of sampling data from the distribution that models ${\cal X}^{(n)}$ is perfectly acceptable in several scenarios, for instance in the context of GANs and VAE, as discussed in Sec.~\ref{sec:preliminaries}. 

The main idea is to estimate directly the derivative of the (asymptotic) scaled LMGF, based on which the rate function estimation can be computed. To this end, let us observe that 
\begin{equation}
   \lim_{n\rightarrow\infty} \mathbb{E}_{f^{(t)}_{n,k}}\left[T^{(n)}({\cal X}^{(n)})\right] = \varphi^{\prime}_k(t),
   \label{eq:phi_prime_tilted}
\end{equation}
provided that ${\cal X}^{(n)}$ is sampled from the tilted distribution, indicated with $f^{(t)}_k ({\cal X}^{(n)})$ under ${\cal H}_k$, and defined as follows
\begin{equation}
    f^{(t)}_{n,k} ({\cal X}^{(n)}) = \frac{\exp(n\,t\,T^{(n)}({\cal X}^{(n)}))}{W_n^{(t)}}\,f_k({\cal X}^{(n)}),
    \label{eq:tilted}
\end{equation}
where $f_k({\cal X}^{(n)})$ is the distribution of ${\cal X}^{(n)}$ under ${\cal H}_k$, and $W_n^{(t)}$ is the distribution normalizing factor. The derivation is straightforward and is reported, e.g., in~\cite{touchette2011basic}.
We assume data can be sampled from the tilted distribution with the MH algorithm,\footnote{Details about the well-known MH algorithm implementation are omitted for brevity, however relevant information can be found e.g. in~\cite{chib1995understanding} and about its application in the LDP context in~\cite{touchette2011basic}.} or with an alternative strategy.

As a consequence, a viable estimator of $\varphi^{\prime}_k(t)$ can be obtained by the sample mean of $T^{(n)}({\cal X}_j^{(n)})$, where data ${\cal X}_j^{(n)}$, $j=1,2\dots,m_z^\prime,$ are drawn from the tilted distribution~(\ref{eq:tilted}): %
\begin{equation}
    \hat{\varphi}^{\prime}_{n,k}(t) = \frac{1}{m_z^{\prime}} \sum_{j=1}^{m_z^{\prime}} T^{(n)}({\cal X}_j^{(n)}), \quad {\cal X}_j^{(n)}\sim f^{(t)}_{n,k} ({\cal X}^{(n)}). 
    \label{eq:LMGF_prime}
\end{equation} 
The rate function is given by the Fenchel-Legendre transform, that can be efficiently solved numerically exploiting the estimated derivative of the scaled LMGF~(\ref{eq:LMGF_prime}) thanks to the duality property discussed in Sec.~\ref{sec:LDP}. Specifically, the approximate point that solves the Fenchel-Legendre transform is given by the equation $\gamma = \hat{\varphi}^{\prime}_{k,n}(t_\gamma)$. Then, based on~\eqref{eq:dual_legendre}, the rate function can be computed by integrating numerically the derivative of the scaled LMGF with the boundary condition that the scaled LMGF is null in $t=0$.

Finally, it is worthwhile stressing that the number of data $m_z^\prime$ required by the MH-based method to achieve a reliable estimate of the rate function (which implicitly requires a large value of $n$) is significantly smaller if compared to the direct estimation method~(\ref{eq:LMGF_est})-(\ref{eq:scaled_LMGF_est}); see details, e.g., in~\cite{touchette2011basic}. 

\emph{Remark:} The MH algorithm requires some knowledge of the distribution of the original data to compute the MH acceptance ratio. In principle, the acceptance ratio can be derived generating data via VAE and GAN techniques. As already mentioned, a partial knowledge of the statistical model of the input data is one of the distinguishing features of model-based ML approaches, see, e.g.,~\cite{shlezinger2021model,shlezinger2022model}.

\subsection{Approximate error probability curves}
\label{sec:approximate_error_prob}
The approximate error probability curve for a given value of $n$ is provided by the exact asymptotics, or the saddlepoint approximation, introduced in Sec.~\ref{sec:exact_asy}. Specifically, the approximate expression of the error probability is provided in~(\ref{eq:exact_asy}), which is valid in more general frameworks than the case of IID observations~\cite{Dembo-Zeitouni}. To compute~(\ref{eq:exact_asy}) we need basically the following ingredients: 
\begin{enumerate}
    \item The saddlepoint $t_{\gamma,k}$ corresponding to the solution $\varphi^{\prime}_{n,k}(t_{\gamma,k}) = \gamma$ under ${\cal H}_k$, $k=0,1$;
    \item The value of the rate function $I_k(\gamma)$, under ${\cal H}_k$, $k=0,1$, where $\gamma$ is the asymptotic threshold, i.e. $\gamma_n \rightarrow \gamma$;
    \item The second derivative of the (scaled) LMGF in the saddlepoint, $\varphi^{\prime\prime}_{n,k}(t_{\gamma,k})$ under ${\cal H}_k$, $k=0,1$.
\end{enumerate}
The first two points are automatically satisfied when we estimate the rate function, and the third point is easy to address by deriving the (scaled) LMGF (details are omitted for brevity). 

Special attention should be given when the threshold is moving with $n$, given that especially for the smaller values of $n$ the approximation can be inaccurate, as it will be discussed in Sec.~\ref{sec:experiments}. In these cases, we can resort to the ``non-asymptotic'' saddlepoint approximation~\cite{wood1993saddlepoint}, which basically consists in computing the previous three ingredients for each value of $\gamma_n$, instead of $\gamma$. 
This leads to compute the (scaled) LMGF, as in~(\ref{eq:scaled_LMGF_est}), the saddlepoint $t_{\gamma_n,k}$, and an approximate rate $I_n(\gamma_n)$, all varying with $n$.

\begin{figure}
\includegraphics[width=.98\columnwidth]{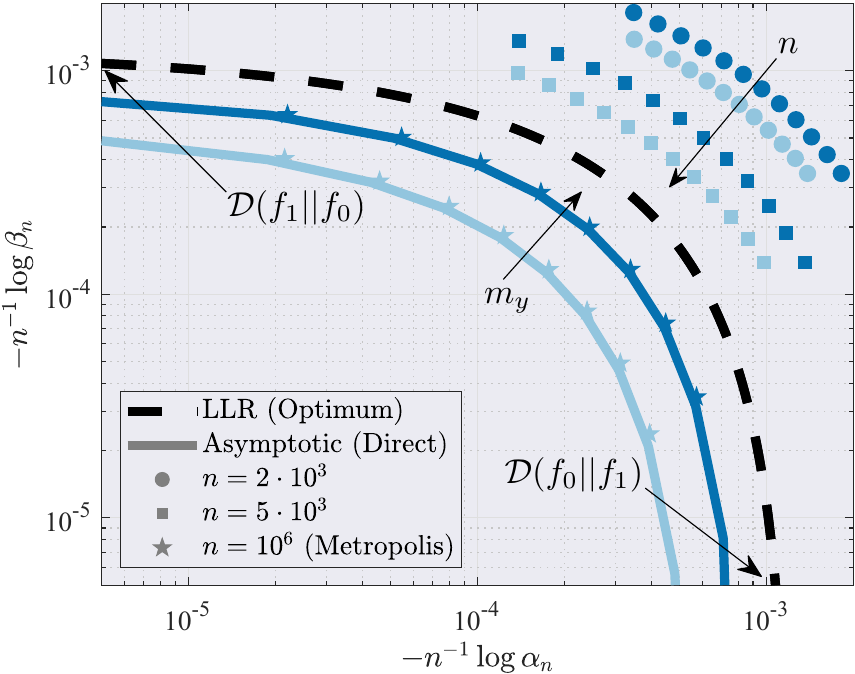}
\caption{Rate function $I_1(\gamma)$ ($y$-axis) versus $I_0(\gamma)$ ($x$-axis) of the LLR (dashed black) and the D3F (solid lines), together with their respective finite sample-size version $-n^{-1} \log \alpha_n$ and $-n^{-1} \log \beta_n$ for different values of $n$ computed via Monte Carlo simulations (markers). The maximum achievable rate is the KL divergence, attained by the LLR, when the rate under the other hypothesis is null.  
Data are IID and generated according to a Laplace distribution with $\theta_0=0$, $\theta_1=0.05$, and $\sigma=1$. Number of Monte Carlo runs is $10^5$ for the empirical curves when $n=2\cdot10^{3}$ and $n=5\cdot10^{3}$, while the curve for $n=10^6$ is obtained sampling from the tilted distribution via the MH algorithm.}
\label{fig:laplace_iid_rate_distortion}
\end{figure}

With this approach, i.e., leveraging the properties of the scaled LMGF for a given value of $n$, it is not guaranteed that the approximate rate is strictly positive as the intervals $[\gamma_n,\infty)$ for $\alpha_n$ and $(-\infty,\gamma_n)$ for $\beta_n$ may include the related expected value of the decision statistic, i.e., $\mathbb{E}\left[T^{(n)}\left| {\cal H}_k \right. \right]$. In this case, we can exploit the Gaussian convergence instead of the exact asymptotics; see~(\ref{eq:CLT}) and~(\ref{eq:CLT_general}). Clearly, the Gaussian approximation is not expected to perform well when $n$ diverges, as already pointed out in the previous sections, and as we will verify in Sec.~\ref{sec:experiments}. However, under the assumption that $\mu_0<\gamma<\mu_1$ it is expected that, for large enough values of $n$, the aforementioned critical situations do not occur.

\begin{figure*}
\captionsetup[subfigure]{labelformat=empty}%
\centering%
\subfloat[][]{%
\includegraphics[width=\columnwidth]{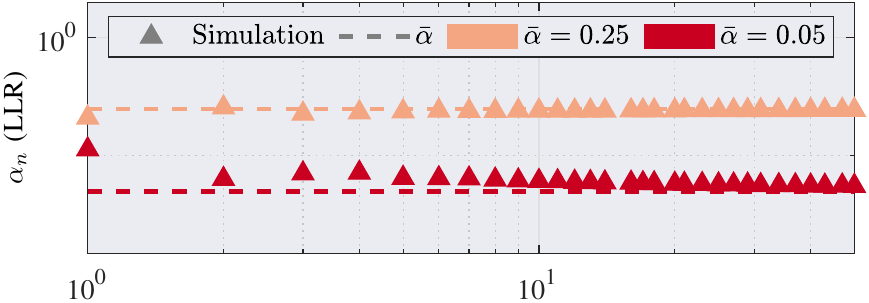}}%
\hfil%
\subfloat[][]{%
\includegraphics[width=\columnwidth]{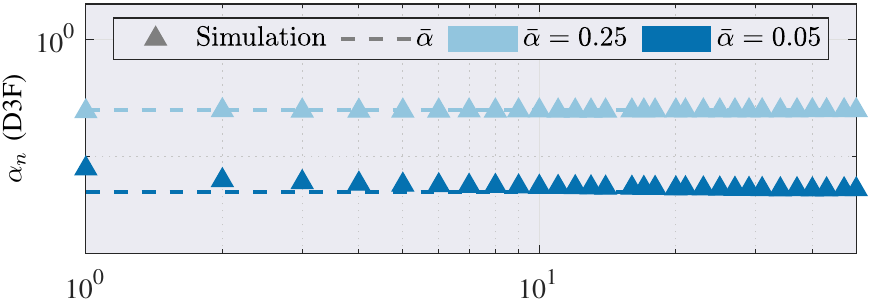}}%
\\[-2em]
\subfloat[][]{%
\includegraphics[width=.98\columnwidth]{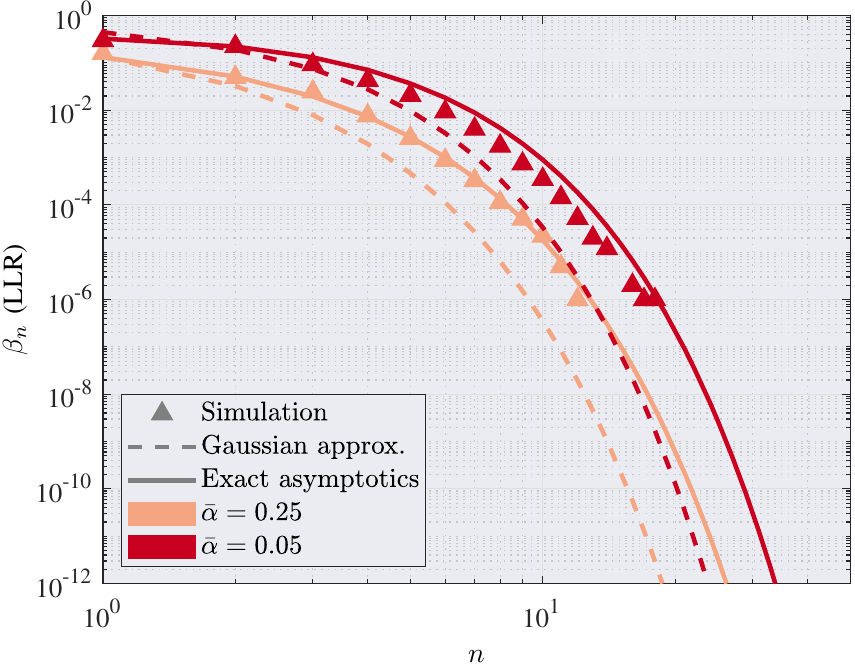}}%
\hfil%
\subfloat[][]{%
\includegraphics[width=.98\columnwidth]{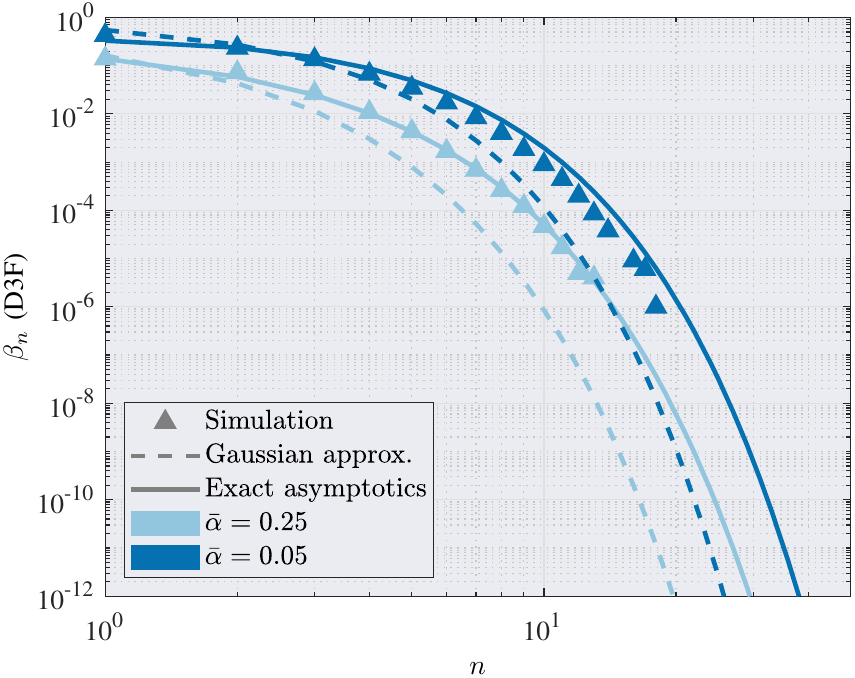}}%
\caption{Error probabilities for Laplace IID observations. Plots in the left column refer to the LLR (shades of red), and those in the right column to the D3F (shades of blue). The plots in the top row illustrate the behaviour of $\alpha_n$, which tends to the desired values $\bar{\alpha}$ equal to $0.25$ (light red/blue) and $0.05$ (dark red/blue). The plots in the bottom row show the behaviour of $\beta_n$ (markers), computed via Monte Carlo simulations, and its two approximations: one based on the exact asymptotics (solid line), and the other one based on the Gaussian convergence (dashed line). The parameters are $\theta_0=0$ and $\theta_1=2$, and the training set size is $m_y=10^3$; the number of Monte Carlo runs is $10^6$.}
\label{fig:laplace_iid_clt_beta}
\end{figure*}

\subsection{Estimation of $\mu_{n,k}$ and $\sigma_{n,k}$ to set the test threshold}

The threshold $\gamma_n$ of the test in~(\ref{eq:test0}) can be selected following two strategies:
\begin{enumerate}
    \item By fixing one of the error probabilities to a desired level, for instance the false alarm, see~(\ref{eq:alpha_CLT}) and~(\ref{eq:alpha_CLT_general}), where $\gamma_n$ converges to $\mu_0$; or
    \item In such a way %
    that both error probabilities vanish, with $\gamma_n = \gamma \in (\mu_0,\mu_1)$.
\end{enumerate}
In the first strategy it is necessary to estimate $\mu_{n,k}$ and $\sigma_{n,k}$. It is straightforward to compute them, as they are the expected value and the standard deviation of the decision statistic under ${\cal H}_k$, respectively. Both $\mu_{n,k}$ and $\sigma_{n,k}$ %
can be estimated via the sample mean exploiting the characterization set. In the second strategy, we only need the asymptotic means $\mu_k$, which can be estimated again with the sample mean, or as the derivative of the scaled LMGF, as reported in~(\ref{eq:mu_k_sigma_k}). Alternatively, from the rate functions we can compute the asymptotic means which are the points where the rates are nulls, i.e., $I_k(\mu_k) = 0$.

\begin{figure*} %
    \captionsetup[subfigure]{labelformat=empty}%
  \centering%
  \begin{minipage}{\columnwidth}
    \subfloat[][]
      {\includegraphics[width=.98\columnwidth]{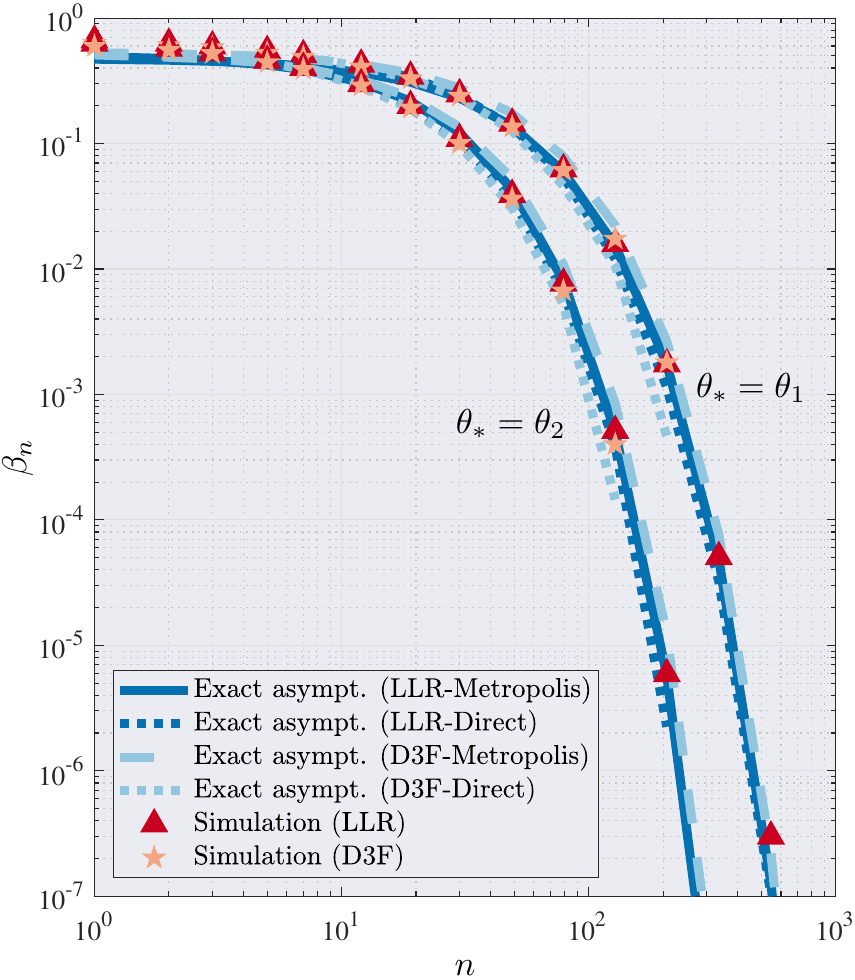}
      }%
  \end{minipage}%
  \hfill %
  \begin{minipage}{\columnwidth}
    \subfloat[][]
       {\includegraphics[width=.98\columnwidth]{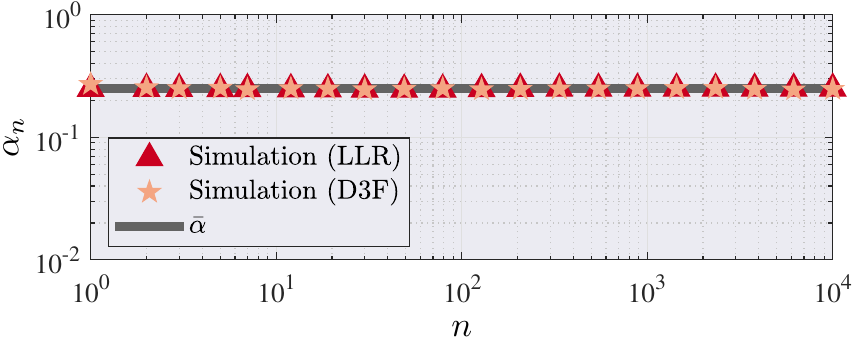}}\\[-1em]
    \subfloat[][]
       {\includegraphics[width=.98\columnwidth]{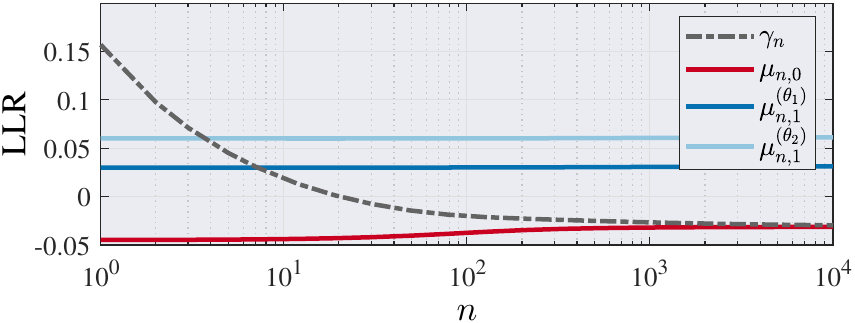}}\\[-1em]
   \subfloat[][]
       {\includegraphics[width=.98\columnwidth]{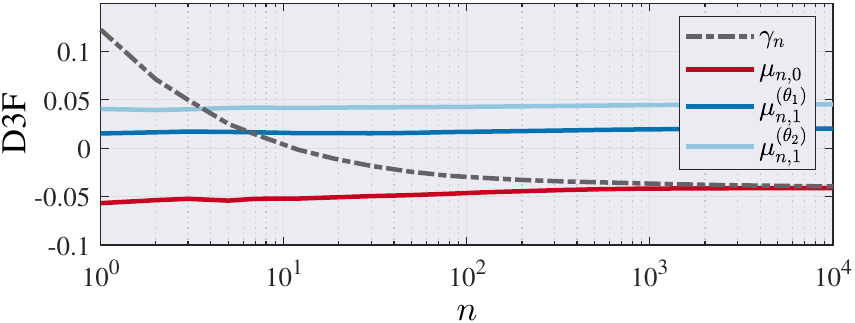}}%
  \end{minipage}%
  \caption{Composite hypothesis test with Gaussian observations. The D3F is trained with $m=10^3$ samples. Leftmost panel: error probability $\beta_n$ of the LLR and D3F; exact asymptotic approximation via exponential tilting sampling for the LLR (solid line) and the D3F (dotted line), Monte Carlo simulations of LLR ($\blacktriangle$, $10^7$ runs), and D3F ($\star$, $10^4$ runs) for each values of the true parameter $\theta_*$. Rightmost upper panel: false alarm probability $\alpha_n$, desired level $\bar{\alpha}=0.25$ (solid line), LLR ($\blacktriangle$) and D3F ($\star$). Rightmost mid and bottom panels: threshold $\gamma_n$ of LLR and D3F compared to the mean $\mu_{n,0}$ under ${\cal H}_0$ and the mean $\mu_{n,1}(\theta)$ under ${\cal H}_1$ where $\theta \in \Theta$.  \label{fig:gaussian_dep_clt}
  }%
\end{figure*}

\section{Experimental results and\\ numerical simulations}
\label{sec:experiments}

In this section we provide a numerical analysis and the sanity check of the theoretical results previously stated. Before entering into details, we remind the distinction between the number of samples $m_y$, namely the size of the training set, and the number of samples $m_z$ of the characterization set, used to compute the means $\mu_{n,k}$ and standard deviations $\sigma_{n,k}$, as well as the rate functions and the approximate error probability curves. In the numerical simulations that will follow, the size of the characterization set is always equal to the number of Monte Carlo runs used to compute the empirical error probabilities.
For the MH algorithm, we used the parameter $n=n_{MH}$, with $n_{MH}=10^4$, to sample from the tilted distribution~\eqref{eq:tilted}; this is done in order to compute the numerical version of~\eqref{eq:phi_prime_tilted}.
Then, in the exact asymptotic formula~\eqref{eq:exact_asy}, the variable $n \geq 1$ is let varying to describe the error probability curve.

\begin{figure*}
    \captionsetup[subfigure]{labelformat=empty}%
    \centering%
    \subfloat[][]{
    \includegraphics[width=0.48\textwidth, trim=120 330 110 330]{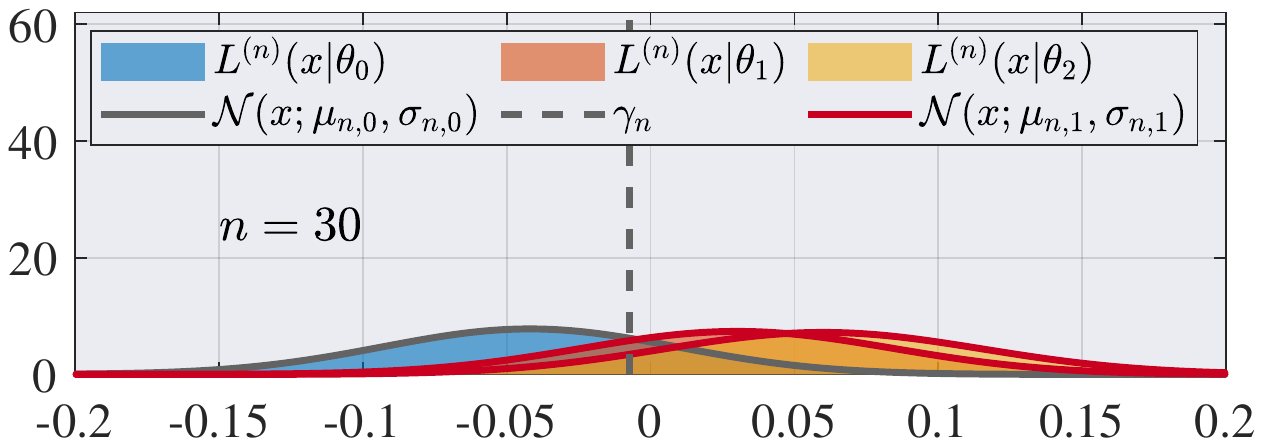}}%
    \hfil%
    \subfloat[][]{
    \includegraphics[width=0.48\textwidth, trim=120 330 110 330]{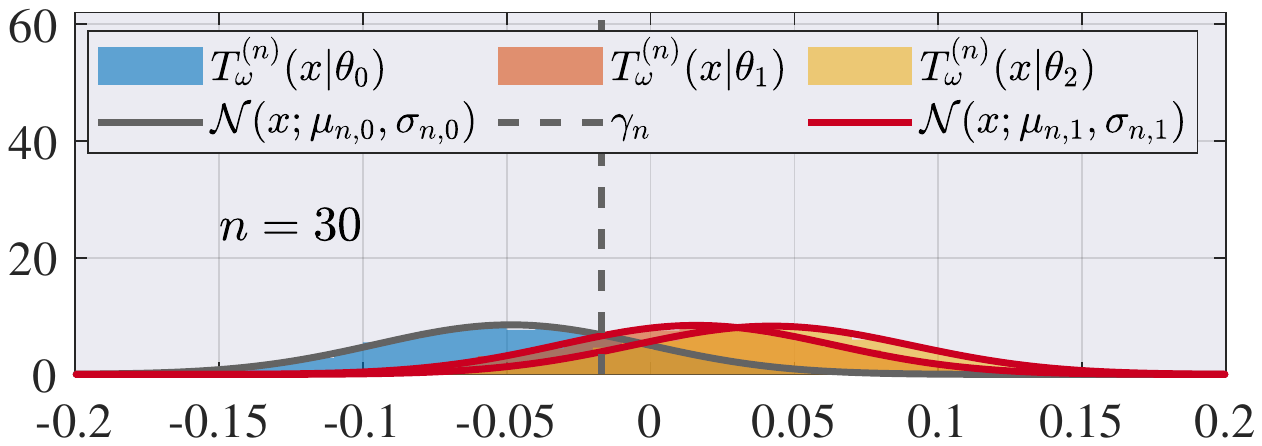}}
    \\[-2em]%
    \subfloat[][]{
    \includegraphics[width=0.48\textwidth, trim=120 330 110 330]{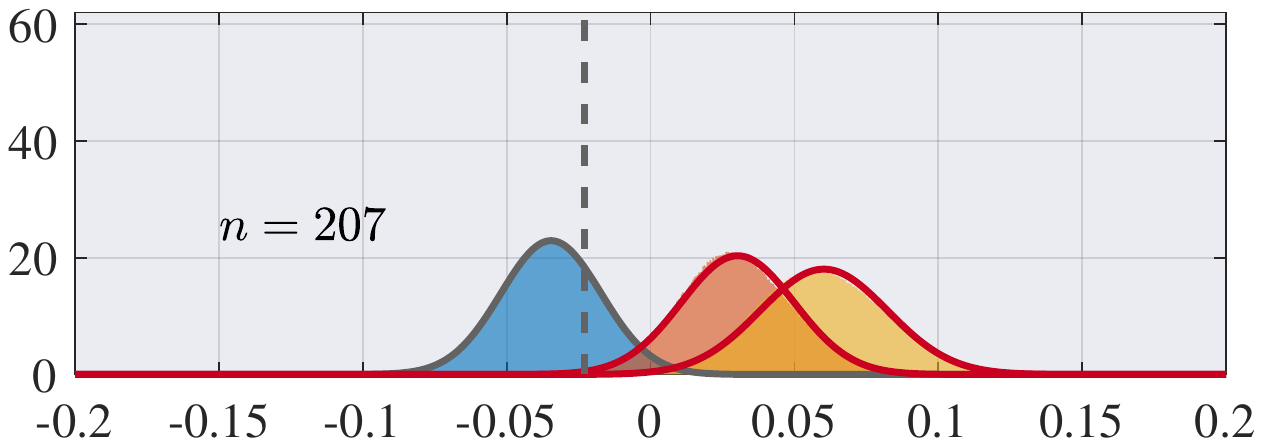}}
    \hfil%
    \subfloat[][]{
    \includegraphics[width=0.48\textwidth, trim=120 330 110 330]{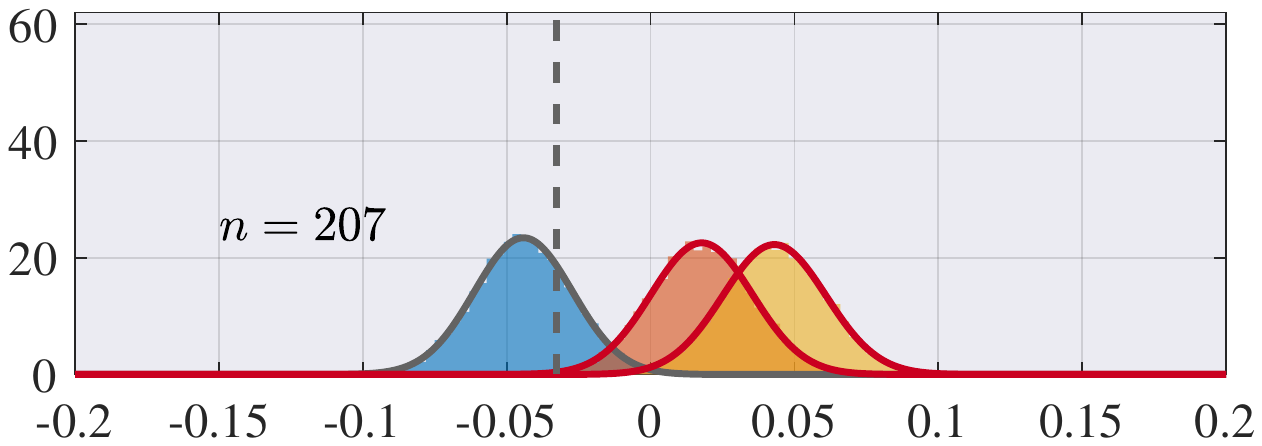}}
    \\[-2em]%
    \subfloat[][]{
    \includegraphics[width=0.48\textwidth, trim=120 330 110 330]{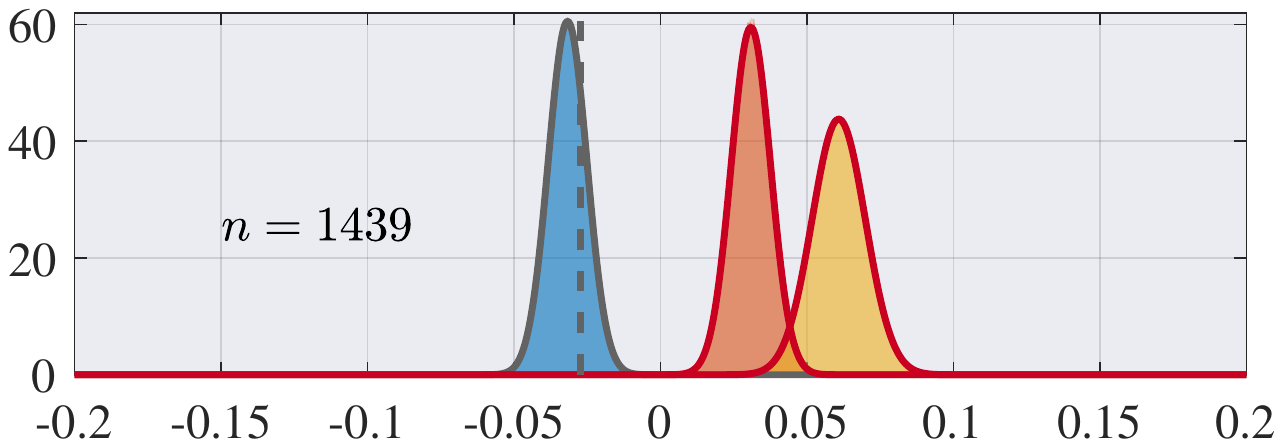}}
    \hfil%
    \subfloat[][]{
    \includegraphics[width=0.48\textwidth, trim=120 330 110 330]{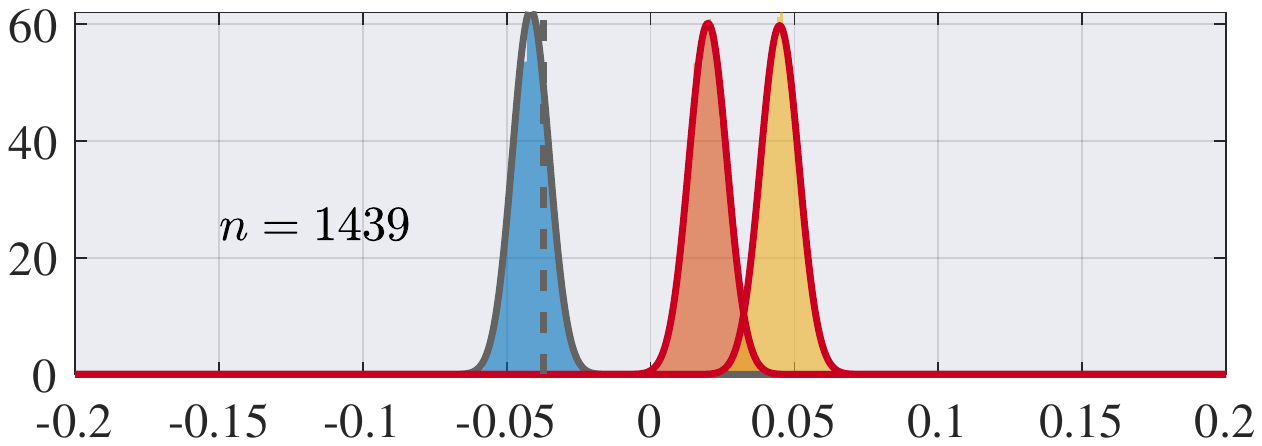}}
    \caption{Distribution of the decision statistics (left column: LLR, right column: D3F) in the case of composite Gaussian hypotheses, where $\theta_0=0$ under $\mathcal{H}_0$ and $\theta \in \{\theta_1, \theta_2\}$ under $\mathcal{H}_1$, with $\theta_1=0.25$ and $\theta_2=0.35$. For the D3F case, the neural network was trained with $m=10^3$ samples. The distributions of the LLR decision statistics have been estimated with $10^6$ realizations, while those for the D3F statistics with $10^4$ realizations. The threshold $\gamma_n$ (dashed gray vertical line) moves with $n$ so that $\alpha$ is fixed to $0.25$.}
    \label{fig:histograms_gaussian_dep}
\end{figure*}

\subsection{IID observations}
We start by analysing the case of IID observations presented in Sec.~\ref{sec:problem} and~\ref{sec:LDP}. Specifically, we consider the case of a shift-in-mean detection problem with noise distributed according to a Laplace distribution. Let us denote by $\mathcal{L}(a, b)$ a (shifted) Laplace distribution with shift parameter $a$ and scale parameter $b$, i.e., having the probability density function
\begin{equation}
f_\mathcal{L} (x) = \frac{1}{2b}e^{-\frac{|x-a|}{b}}.
\end{equation}
Specifically, we test $x_i\sim\mathcal{L}(\theta_0, \sigma)$ under $\mathcal{H}_0$ and $x_i\sim\mathcal{L}(\theta_1, \sigma)$ under $\mathcal{H}_1$. The elementwise D3F is trained with $m_y/2$ independent observations under both ${\cal H}_0$ and ${\cal H}_1$. As loss function, the binary cross-entropy loss~\eqref{eq:cross-entropy} is adopted. The D3F is defined as a simple fully connected NN with two hidden layers of $5$~neurons each. The first hidden layer has a $\tanh$ activation function, and the second is followed by a $\operatorname{softmax}$ activation function; the elementwise D3F $t_{\bm{\omega}}(x)$ is given by the log-ratio of the two $\operatorname{softmax}$ outputs. Figure~\ref{fig:decision_rules} illustrates the shape of the D3F $t_{\omega}(x)$ trained with data drawn from a Laplace (left panel) and Gaussian distribution (right panel)
for different values of $\theta_1$ with $\theta_0 =0$ and $\sigma=1$, as well as the shape of the LLR computed for the same parameters. The scenario with Gaussian data is described in the next subsection. It is apparent that the D3F and the LLR functions are somehow similar, but also quite different, especially for larger values of $x$. However, we will see later that this shape is not informative in terms of detection performance.  
\begin{figure*}
\captionsetup[subfigure]{labelformat=empty}%
    \centering%
    \subfloat[][]{
    \includegraphics[width=0.48\textwidth]{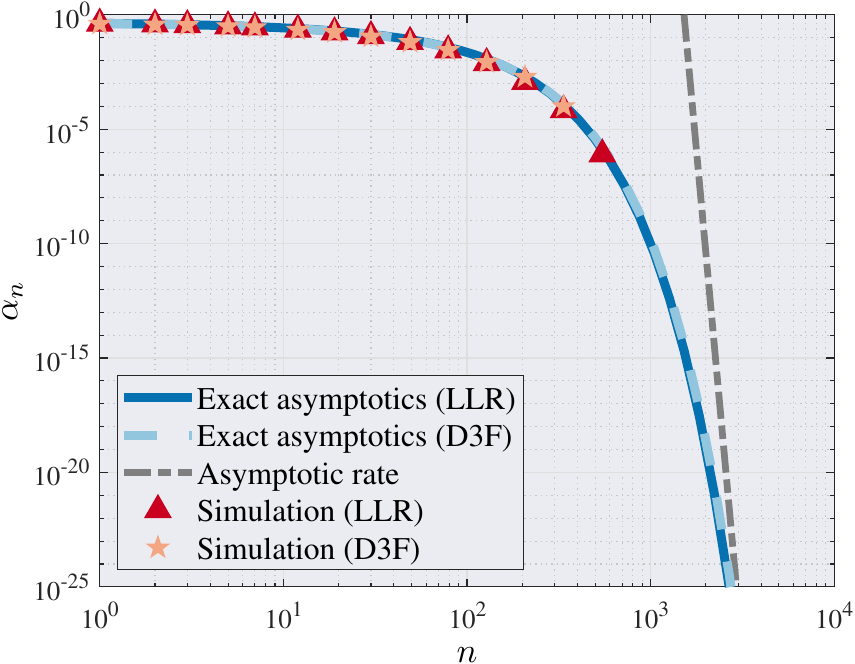}}%
    \hfil%
    \subfloat[][]{
    \includegraphics[width=0.48\textwidth]{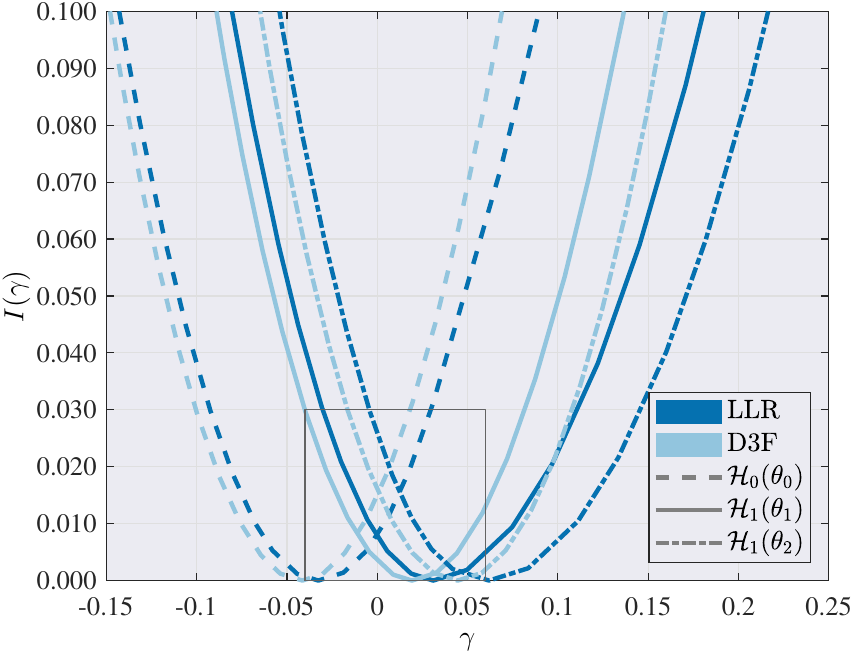}}
    \\[-1em]%
    \subfloat[][]{
    \includegraphics[width=\columnwidth]{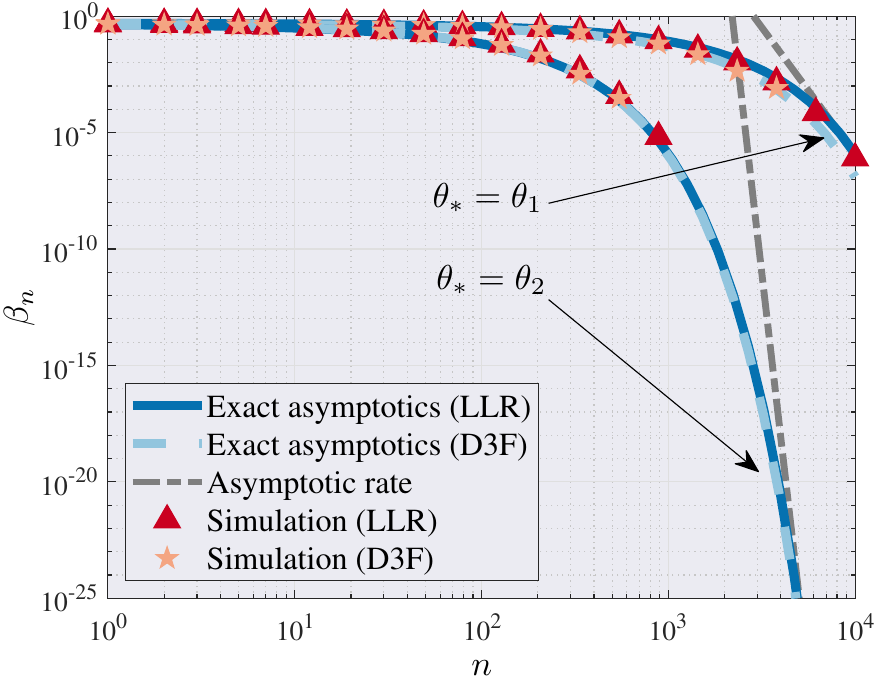}}
    \hfil%
    \subfloat[][]{
    \includegraphics[width=0.48\textwidth]{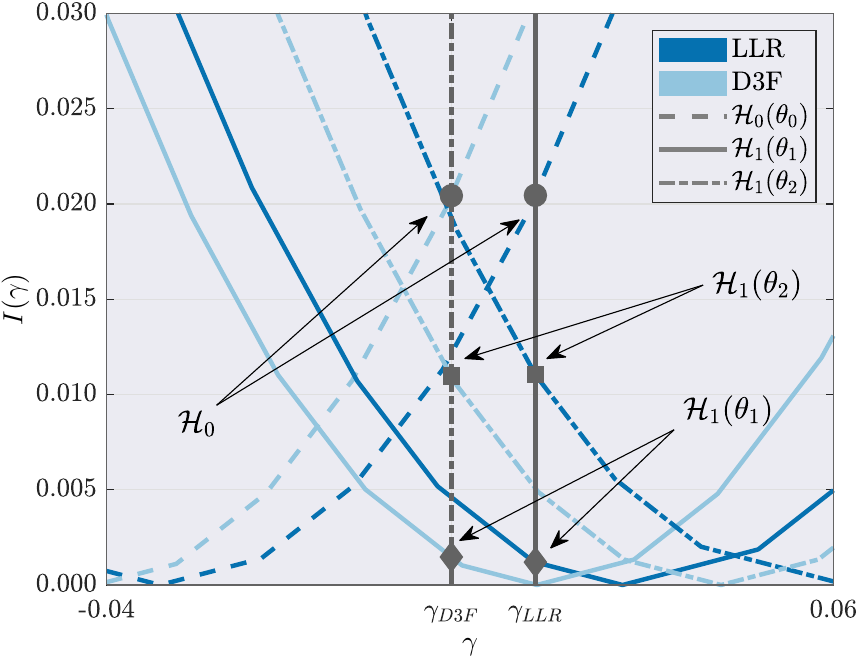}}
    \caption{Composite hypothesis test with Gaussian observations. The leftmost plots show: $\alpha_n$ and $\beta_n$ of the LLR and the D3F; the exact asymptotics computed via exponential tilting for the LLR (solid) and the D3F (dashed); the Monte Carlo simulations for the LLR ($\blacktriangle$, $10^7$ runs) and the D3F ($\star$, $10^4$ runs); the asymptotic rates (dot-dashed) given by~(\ref{eq:LLR_scaled_LMGF}). The rightmost plots show the rate functions under both hypotheses, computed via exponential tilting. The bottom-right plot is a close-up of the rectangular region represented by the box in the top-right plot and shows the test operational points given by the thresholds $\gamma_{D3F}$ and $\gamma_{LLR}$ for the D3F and LLR, respectively. The D3F is trained with $m=10^3$ samples.}
    \label{fig:gaussian_dep_fixed_threshold}
\end{figure*}

In Fig.~\ref{fig:laplace_iid_rate_distortion} we analyse the scenario with $\theta_0=0$, $\theta_1=0.05$ and $\sigma=1$. We show both the rate functions: $I_0(\gamma)$ under ${\cal H}_0$ and $I_1(\gamma)$ under ${
\cal H}_1$ for the LLR (dashed line) and D3F (solid line) and two different values of $m_y$ (represented with dark and light blue). The rates are computed by varying the threshold $\gamma$ and are evaluated with the direct method described in Sec.~\ref{sec:estimation_rate}, except for %
for $n=10^6$, in which case are obtained by sampling from the tilted distribution (see Sec.~\ref{sec:estimation_rate}). The reason is that for $n=10^6$ the error probabilities would have been prohibitively small for a direct Monte Carlo simulation.\footnote{To give a rough idea, with $n=10^6$ and a rate $I=10^{-4}$, the error probability would be $\sim \exp(-10^2) \approx 3.7 \times 10^{-44}$.} %
The convergence is also assessed numerically for different values of $n$ (i.e., $2~\cdot~10^3$ and $5~\cdot~10^3$) by computing the error probabilities $\alpha_n$ and $\beta_n$ (markers) empirically via Monte Carlo simulations. We can see that have $-{n^{-1}}\log \alpha_n \rightarrow I_0(\gamma)$ and $-{n^{-1}}\log \beta_n \rightarrow I_1(\gamma)$, as established by~(\ref{eq:LDP_LLR})-(\ref{eq:LDP_D3F}). %
In Fig.~\ref{fig:laplace_iid_rate_distortion}, as expected, being the optimal detector, the LLR achieves the best rates. %
However, we observe that increasing $m_y$, from $10^3$ to $10^4$, the D3F curve gets closer to the curve of the LLR. The extreme values of the LLR rate, where $I_0 = 0$ or $I_1=0$, provide respectively the best achievable rate, which is equal to the KL divergence $\mathcal{D}(f_0||f_1)$ when $I_1=0$ and $\mathcal{D}(f_1||f_0)$ when $I_0=0$. We analyze the test performance of both the decision statistics in one of these extreme points in Fig.~\ref{fig:laplace_iid_clt_beta}, where $I_0 (0) = 0$, setting the threshold as in~(\ref{eq:alpha_CLT}) and thus obtaining $\alpha_n \rightarrow \bar{\alpha}$. The desired values of false alarm are set to $0.25$ and $0.05$, and, in this case study, we set the parameter under ${\cal H}_1$ to $\theta_1 = 2$,  while $\theta_0 = 0$. 

In Fig.~\ref{fig:laplace_iid_clt_beta} the plots on the left refer to the LLR, while the those on the right refer to the D3F. The markers represent the empirical probabilities computed via Monte Carlo simulations. The upper panels report the behaviour of $\alpha_n$, which tends to the desired values (dashed lines) even for small values of $n$. The lower panels show instead the behaviour of $\beta_n$ for the two desired asymptotic false alarm levels and their approximations based on the exact asymptotics~(\ref{eq:exact_asy}) (solid line), properly refined as described in Sec.~\ref{sec:approximate_error_prob}, and based on the Gaussian convergence~(\ref{eq:CLT}) (dashed line). We shall note that for values of $n$ lower than $10$, the Gaussian approximation is accurate, especially when the false alarm is smaller, while the exact asymptotics provide a good approximation. However, it becomes evident as $n$ increases that the Gaussian approximations converge to zero faster than the exact asymptotic curves, which instead match much and better the empirical probability values. This behaviour of the Gaussian approximation is perfectly aligned with the theory discussed in the previous sections, as the Gaussian approximation is more accurate when the event is not too far from the mean where the random variable sequence is converging; %
in this region, the theory of large deviations shall be used to achieve a better accuracy. It is interesting to highlight that %
in Fig.~\ref{fig:laplace_iid_clt_beta}, %
even if the convergence rate is the same for the two different asymptotic false alarm probabilities, the sub-exponential terms intervene %
to %
separate the curves. The curves exhibit the same rate, meaning that for large $n$ the curves converge to zero in parallel with the same slope.  

\subsection{Conditional independent observations: Composite hypothesis testing}

Let us now move on the composite hypothesis testing problem described in Sec.~\ref{sec:LDP_conditionally_independent}. This time we consider Gaussian observations $x_i \sim {\cal N}(\theta,\sigma)$; the Laplace example is similar and not reported for brevity. We assume $\sigma =1$ under both the hypotheses. Under ${\cal H}_0$, we set the parameter $\theta_0=0$, whereas under ${ \cal H}_1$ it can take two different possible values in $\Theta = \left\{\theta_1,\theta_2\right\}$; specifically, we set $\theta_1 = 0.25$ and $\theta_2 = 0.35$. The two parameter values are a priori equally probable. The D3F is trained exactly as for the IID case, but the training procedure is repeated for each value of $\theta$ under ${ \cal H}_1$, as described in Sec.~\ref{sec:LDP_conditionally_independent}. The number of training samples is $m_y=10^3$ under each hypothesis and for each value in $\Theta$. The shape of the elementwise D3F is illustrated in the right-hand side plot of Fig.~\ref{fig:decision_rules}. 
In Fig.~\ref{fig:gaussian_dep_clt}, we report the convergence of $\beta_n$ to zero (left-side panel) when the true parameter $\theta_*$ under ${ \cal H}_1$ is equal to $\theta_1$ and $\theta_2$ with an asymptotic false alarm fixed to $0.25$, and resorting to a threshold selection $\gamma_n$ according to the derivation in Sec.~\ref{sec:CLT_Dep}. We can see that the false alarm $\alpha_n$ (right side, top panel) converges quickly to the desired value even for small values of $n$ for both the LLR and the D3F.
We compare the exact asymptotic approximations (both the direct computation and the exponential tilting sampling methods described in Sec.~\ref{sec:approximate_error_prob}) with the Monte Carlo simulation of the error probability $\beta_n$ of the D3F ($10^4$ runs) and the LLR ($10^7$ runs).\footnote{The number of Monte Carlo runs of the LLR is larger than the D3F thanks to a convenient manipulation of the conditional LLR in~\eqref{eq:conditional_Ln}.} It is important to highlight that in this specific setting, different from the Laplace example, both the LLR and the D3F achieve the same performance in terms of error probabilities. This will be confirmed in the rate function analysis in Fig.~\ref{fig:gaussian_dep_fixed_threshold}. 

We also report the convergence of $\gamma_n$ for both LLR and D3F (middle and bottom, respectively, right-side panels), and, as expected, $\gamma_n \rightarrow \gamma = \mu_{n,0}$. It is worthwhile to note that the agreement between the exact asymptotic approximations and the Monte Carlo simulations is less accurate when $n$ is smaller than 10, this is because the threshold $\gamma_n$ is larger than $\mu_{n,1}(\theta)$ %
and consequently there is not a meaningful saddlepoint in this condition as explained in Sec.~\ref{sec:approximate_error_prob}. 

A pictorial analysis of the ``small deviations'' regime %
is reported in Fig.~\ref{fig:histograms_gaussian_dep}, %
which illustrates the convergence to the Gaussian distribution discussed in Sec.~\ref{sec:CLT_Dep}. Left-column panels refer to the LLR, and right-column panels to the D3F. Each row refers to a different value of $n$, specifically $30, 207, 1439$. The decision statistic distribution is simulated under ${\cal H}_0$ and ${\cal H}_1$ for each value of $\theta \in \Theta$, and we report in Fig.~\ref{fig:histograms_gaussian_dep} the related histograms (colored area). Moreover, we also plot the approximated Gaussian distributions ${\cal N}(x; \mu_{n,k},\sigma_{n,k})$ (solid line) where $\mu_{n,k} \sim \mu_k$ and $\sigma_{n,k} \sim \frac{\sigma_k}{\sqrt{n}}$. There is almost a perfect match between the histograms and the Gaussian distributions for all the values around the means; however, as we already discussed, the agreement is expected to be increasingly less accurate on the tails\footnote{The distributions' tails cannot be easily visualized in the histograms, but they are analyzed in terms of error probabilities in Figs.~\ref{fig:gaussian_dep_clt} and~\ref{fig:gaussian_dep_fixed_threshold}.}. We observe that the larger $n$, the more the decision statistic distributions concentrate around the asymptotic mean $\mu_k$, with a variance that is decreasing, which  confirms the convergence in~(\ref{eq:CLT_general}). We also show how the threshold moves with $n$ so that the asymptotic false alarm remains fixed at the desired level of $0.25$, as in Fig.~\ref{fig:gaussian_dep_clt}.

In Fig.~\ref{fig:gaussian_dep_fixed_threshold}, we set a fixed threshold $\gamma\in\left(\mu_0,\min_{\theta \in \Theta}\mu_{1}(\theta)\right)$ in order to allow the convergence to zero of both the error probabilities (see left-hand side panels of Fig.~\ref{fig:gaussian_dep_fixed_threshold}). By doing so, trivial operating points are avoided.
For this reason, we report in the right hand side panels of Fig.~\ref{fig:gaussian_dep_fixed_threshold} the Fenchel-Legendre transform (computed by means of the exponential tilting method) of both the LLR (dark blue) and the D3F (light blue), under ${\cal H}_0$ and under ${\cal H}_1$ for both values of $\theta$. The LLR curves have a parabolic shape, in perfect agreement with~(\ref{eq:LLR_scaled_LMGF}); indeed, the asymptotic scaled LMGF for conditionally independent Gaussian observations is given by the LMGF of the elementwise LLR, which is still Gaussian. Then, it is easy to compute the Fenchel-Legendre transform, which turns out to be a parabola (see, e.g.,~\cite{Matta_IT2016}). Interestingly, the D3F rate curves seem to be approximately an horizontal shift of the LLR ones. However, in order to have a meaningful comparison, we also need to select two thresholds, indicated with ${\gamma}_{LLR}$ and ${\gamma}_{D3F}$ for the LLR and the D3F, respectively (see bottom-right panel in Fig.~\ref{fig:gaussian_dep_fixed_threshold}, which provides a zoom of the upper-right panel), so that the rates under ${\cal H}_0$ for the LLR and D3F are equal. Once the thresholds have been chosen, the rates under ${\cal H}_1$ are automatically determined for both the LLR and the D3F in each of the points $\theta_1$ and $\theta_2$.
In the left-hand side panels of Fig.~\ref{fig:gaussian_dep_fixed_threshold} we illustrate $\alpha_n$ (top) and $\beta_n$ (bottom). Specifically, we report the empirical error probabilities computed by means of Monte Carlo simulations, along with the asymptotic approximations, which again are in a very good agreement. Similar to the previous example (Fig.~\ref{fig:gaussian_dep_clt}), the LLR and D3F achieve the same performance; this is also due to the fact that they have very similar rate functions, as reported in the Fenchel-Legendre transform curves in the right-side panels of Fig.~\ref{fig:gaussian_dep_fixed_threshold}. We also plot the theoretical asymptotic rate provided by~(\ref{eq:LLR_scaled_LMGF}), which intercepts the curves when $n$ is sufficiently large. It is interesting to observe that $\alpha_n$ converges faster to zero than $\beta_n$, being the rate around $0.02$, while under $\beta_n$ has two rates, the first is around $0.011$ when $\theta_*=\theta_2$ and the second is below $0.005$ when $\theta_*=\theta_1$. Our intuition suggests that the performance will always be worse for $\theta_*=\theta_1$, given that this value is closer to $\theta_0$ than $\theta_2$. Indeed, in the left-side part of the parabola the rate function of $\theta_2$ is always larger than the rate function of $\theta_1$ for both the LLR and the D3F, and this behaviour is confirmed in Fig.~\ref{fig:gaussian_dep_clt}, where the error converges to zero faster when $\theta_*=\theta_2$; see the rate functions of ${\cal H}_1$ in the point in which the rate of ${\cal H}_0$ is null in $\mu_0$ (around $-0.04$ for the LLR and $-0.03$ for the D3F). We have the opposite behaviour on the right-side of the parabola, where the rate of $\theta_1$ is larger than the rate of $\theta_2$; however, this region is not relevant for our scopes, as there are always no large deviations for $\beta_n$ under $\theta_1$, and no large deviations for both values of $\theta$ when the threshold is larger than $\mu_1(\theta_2)$.

\begin{figure}
\captionsetup[subfigure]{labelformat=empty}%
\centering%
\subfloat[][]{%
\includegraphics[width=.99\columnwidth]{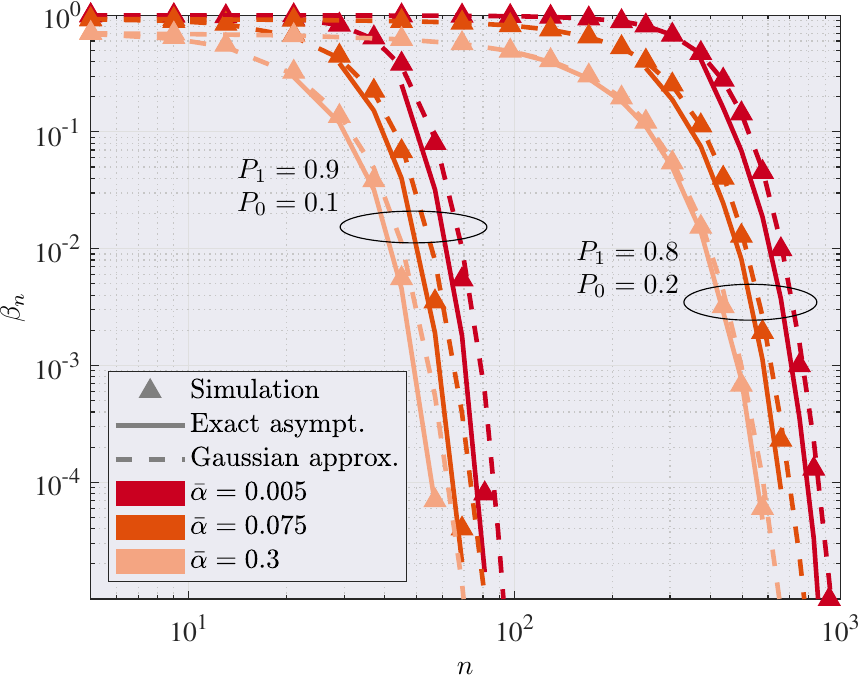}}%
\\[-1.2em]
\subfloat[][]{%
\includegraphics[width=.99\columnwidth]{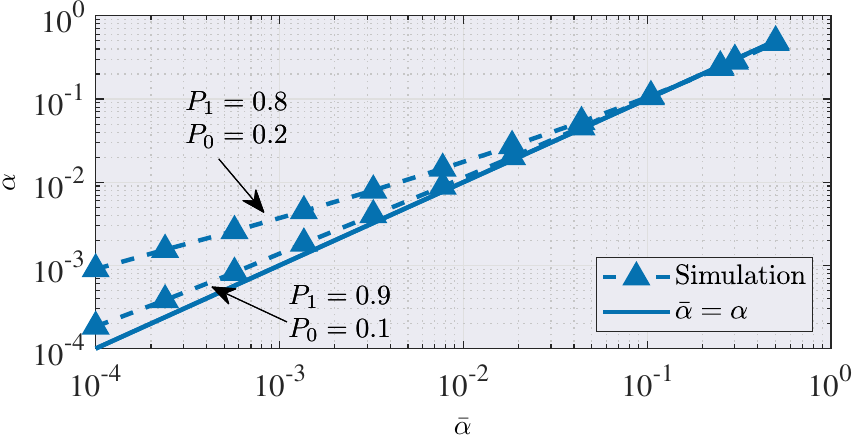}}%
\\[-1.2em]
\subfloat[][]{%
\includegraphics[width=.99\columnwidth]{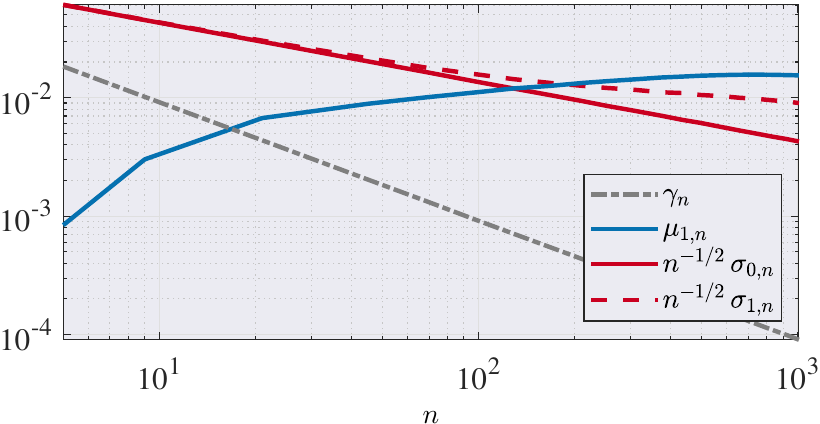}}
\vspace{-1.5em} \caption{Detection of presence/absence of an extended target in a binary image. Top panel: miss detection probability $\beta_n$, with $n$ being the number of resolution cells occupied by the target. The error probability is computed for three different false alarm levels via Monte Carlo simulations ($\blacktriangle$) with $10^5$ runs, and compared with the exact asymptotic and Gaussian approximations. Two scenarios are considered: in the former, the detection probability and false alarm probability of each cell resolution (pixel) are $P_1 = 0.9$ and $P_0 = 0.1$, respectively; in the latter, they are $P_1 = 0.8$ and $P_0 = 0.2$, respectively. Middle pane: desired false alarm level $\bar{\alpha}$ vs false alarm level observed in the Monte Carlo simulation ($\blacktriangle$). Bottom panel: convergence of $\gamma_n$, $\mu_{1,n}$ and $\sqrt{n}\sigma_{k,n}$ for $k=0,1$, where $\gamma_n = \frac{\gamma}{n}$, with $\gamma$ given in \eqref{eq:gamma_image}, while $\mu_{1,n}$ and $\sqrt{n}\sigma_{k,n}$ are the parameters of the normalized decision statistic~\eqref{eq:D3F_dep_Tn}.}
\label{fig:extended_target_clt_beta}
\end{figure}

\begin{figure*}
    \captionsetup[subfigure]{labelformat=empty}%
    \centering%
    \subfloat[][]{
    \includegraphics[width=0.98\columnwidth, trim=120 345 110 330]{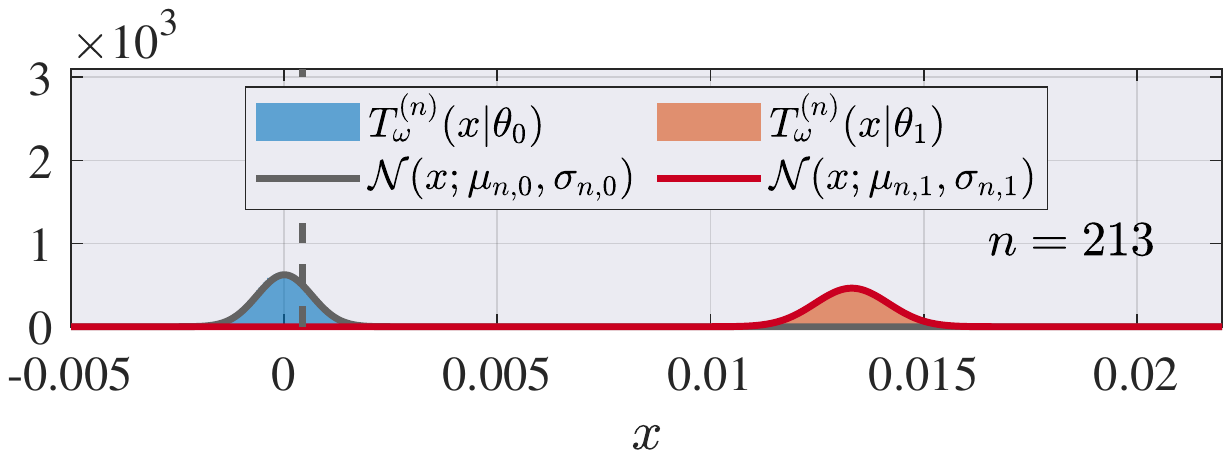}}%
    \hfil%
    \subfloat[][]{
    \includegraphics[width=0.98\columnwidth, trim=120 345 110 330]{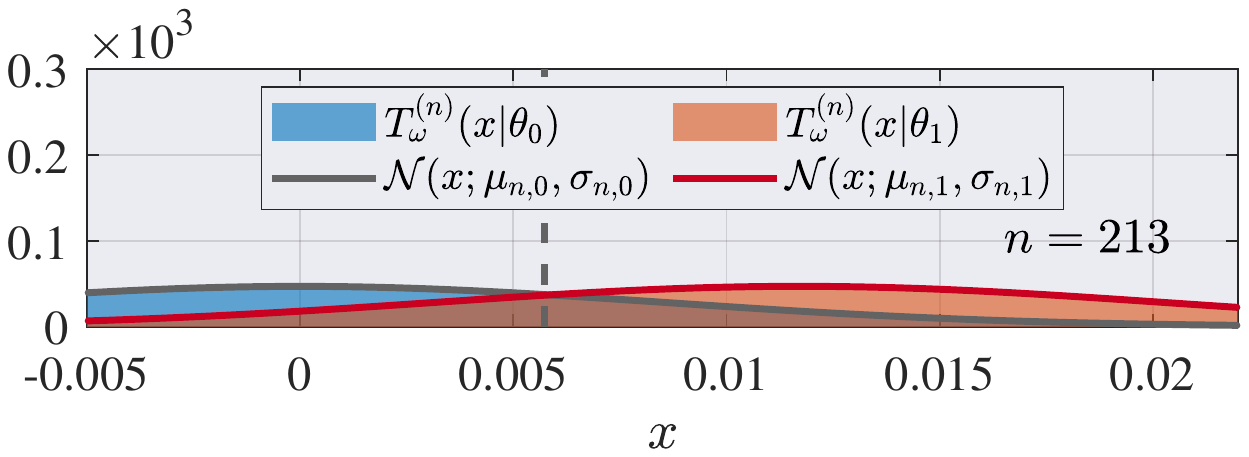}}
    \\[-1.4em]
    \subfloat[][]{
    \includegraphics[width=0.98\columnwidth, trim=120 345 110 330]{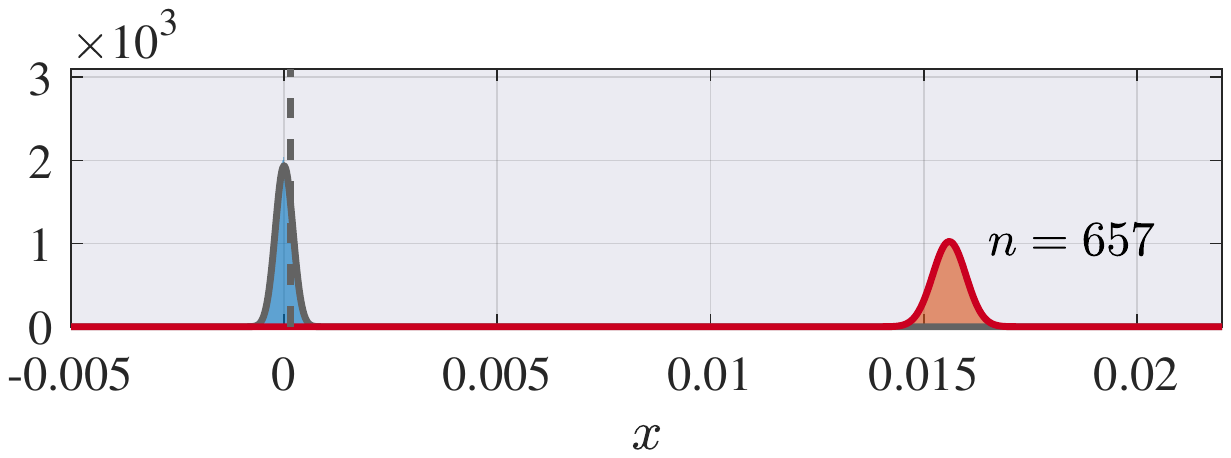}}%
    \hfil%
    \subfloat[][]{
    \includegraphics[width=0.98\columnwidth, trim=120 345 110 330]{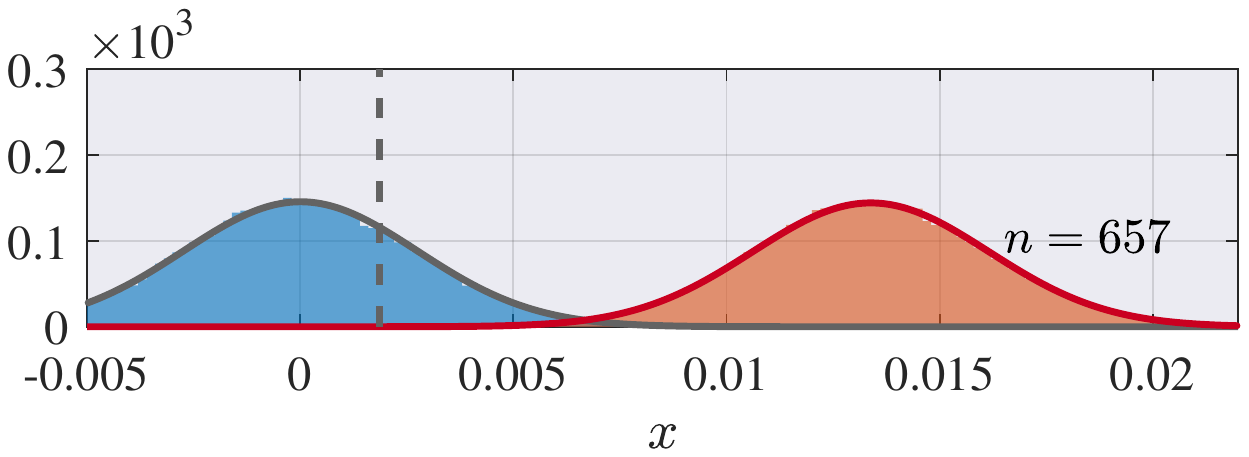}}
    \\[-1.4em]
    \subfloat[][]{
    \includegraphics[width=0.98\columnwidth, trim=120 345 110 330]{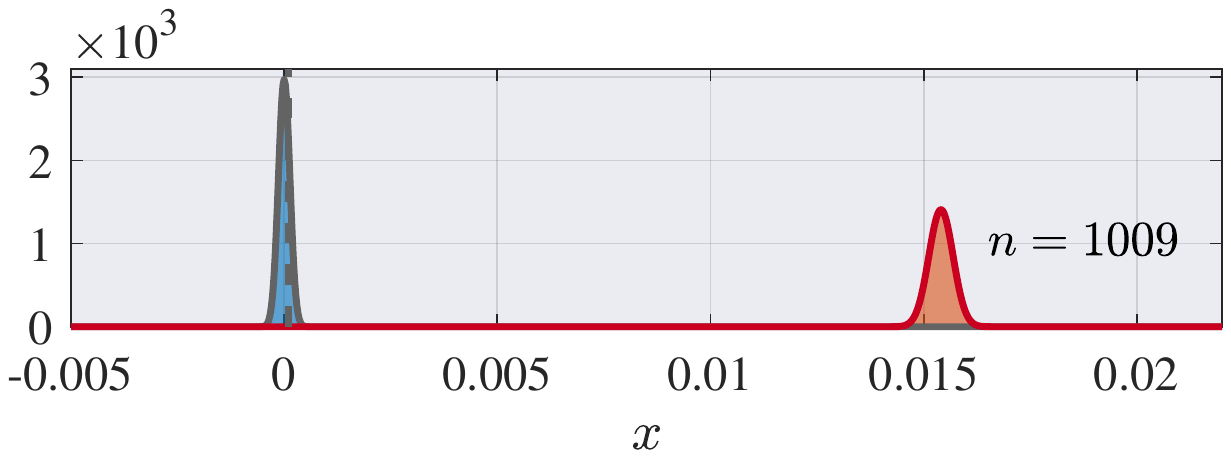}}%
    \hfil%
    \subfloat[][]{
    \includegraphics[width=0.98\columnwidth, trim=120 345 110 330]{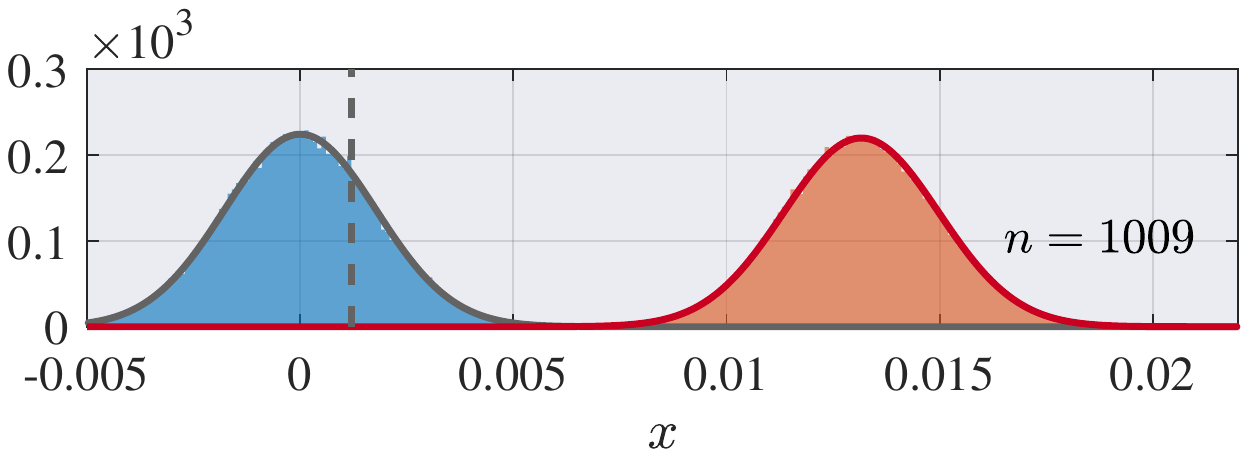}}
    \caption{Distribution of the normalized D3F statistics~\eqref{eq:D3F_dep_Tn} for the problem of detecting an extended target in an image under ${\cal H}_0$ (blue) and ${\cal H}_1$ (red) (left column, matched scenario: $P_{1}=0.9$, $P_{0}=0.1$. Right column, mismatched scenario: $P_{1}=0.8$, $P_{0}=0.2$). The D3F statistic is compared with the Gaussian distributions centered in $\mu_{n,k}$ with variance $\sigma_{n,k}^2$ for different values of $n$, where we have translated the decision statistics to obtain $\mu_{n,0}=0$.}
    \label{fig:histograms_extended_target}
\end{figure*}

\begin{figure*}
\captionsetup[subfigure]{labelformat=empty}%
    \centering%
    \subfloat[][]{
    \includegraphics[width=0.3\textwidth, trim=40 0 40 0]{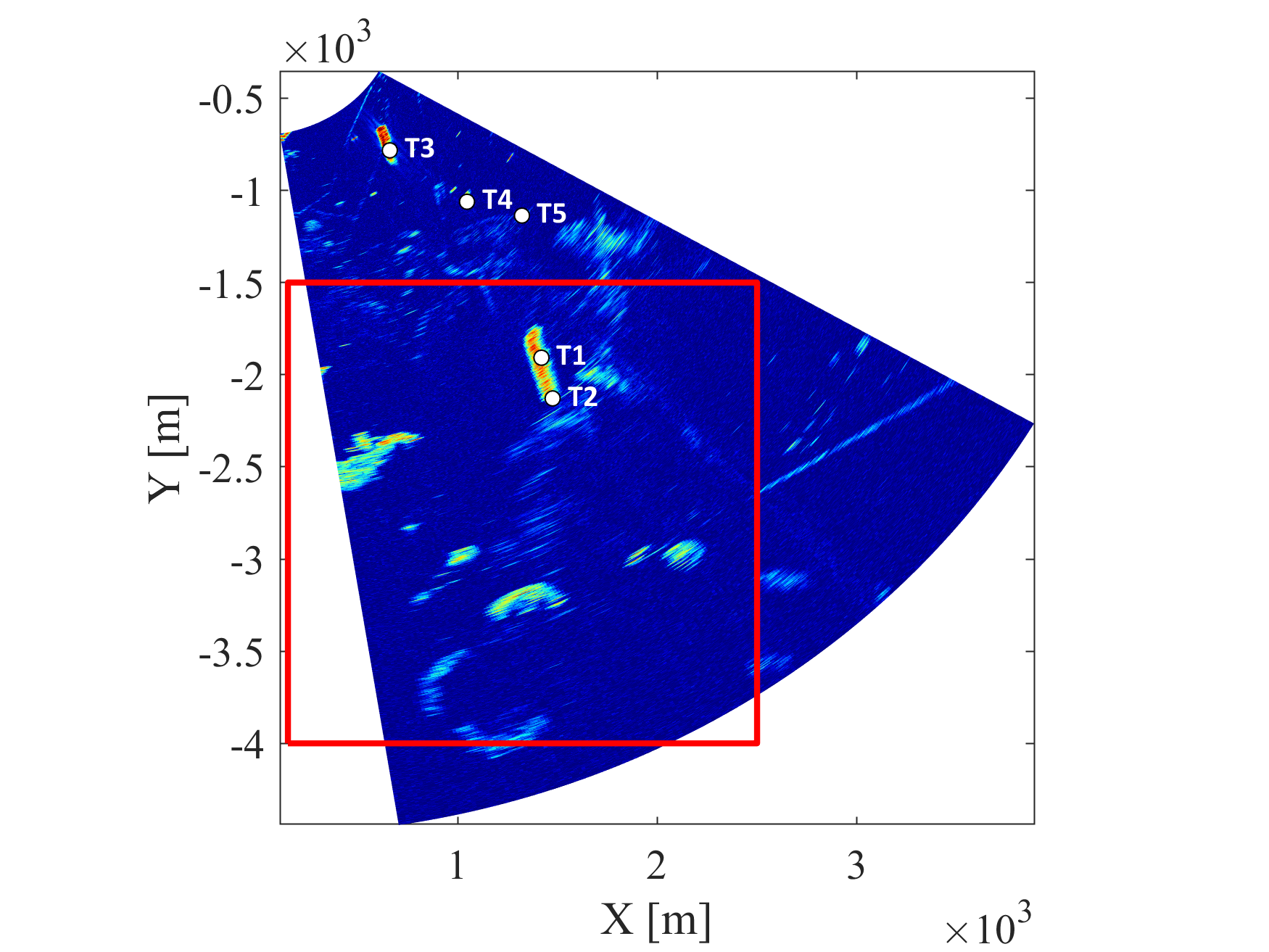}}%
    \hfil%
    \subfloat[][]{
    \includegraphics[width=0.3\textwidth, trim=40 0 40 0]{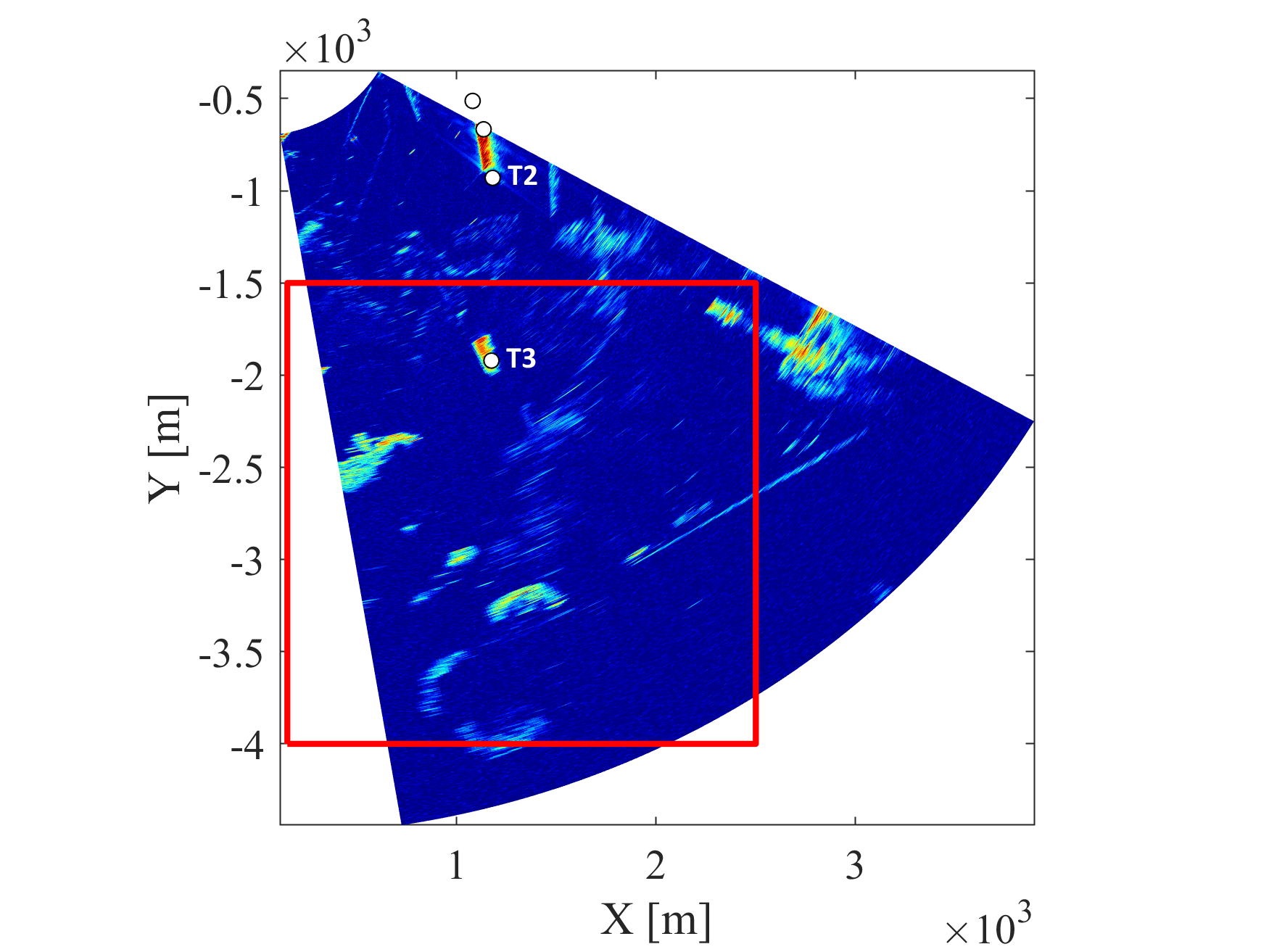}}
    \hfil%
    \subfloat[][]{
    \includegraphics[width=0.3\textwidth, trim=40 0 40 0]{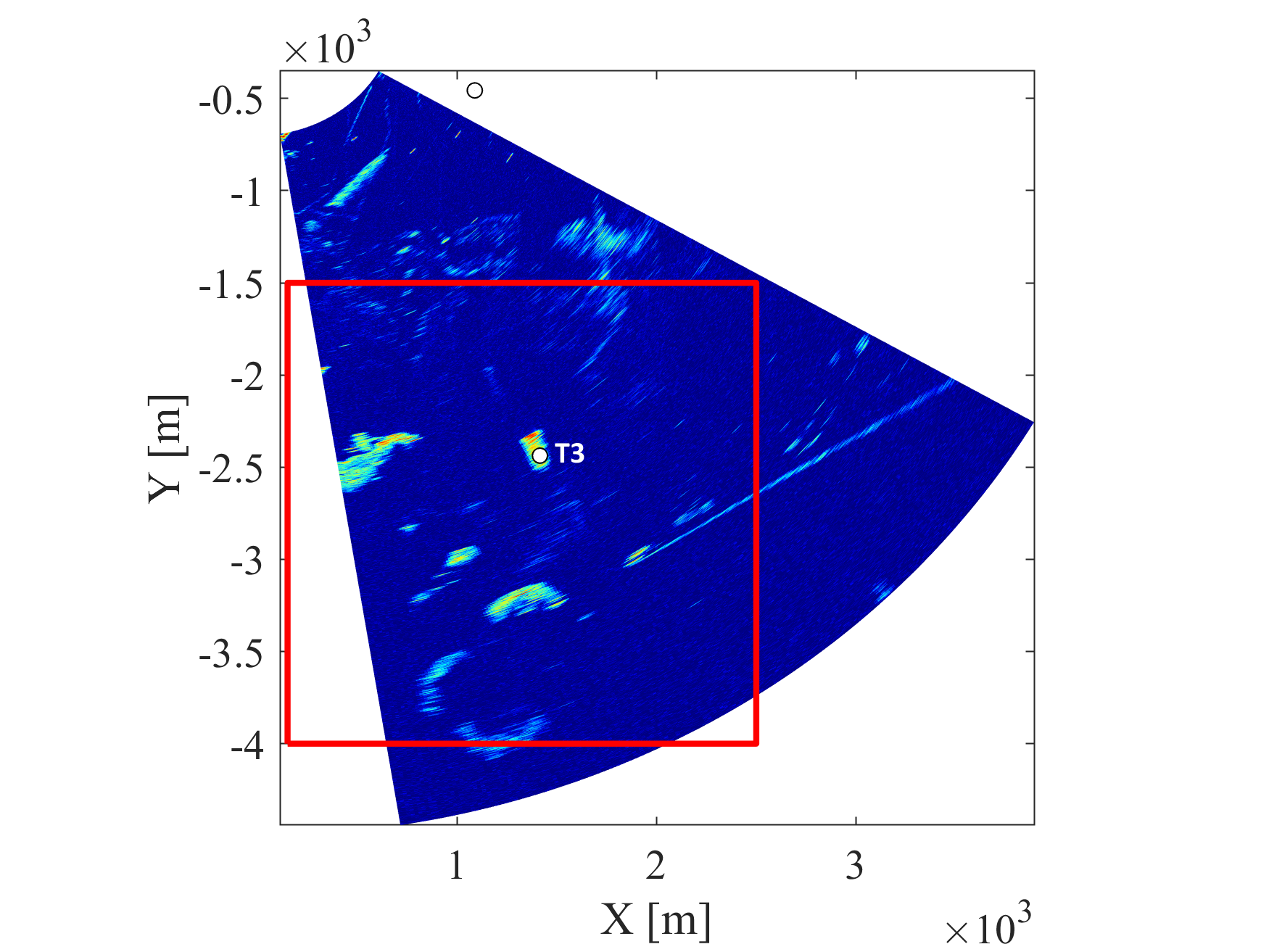}}
    \\[-2em]%
    \subfloat[][]{
    \includegraphics[width=0.3\textwidth, trim=40 0 40 0]{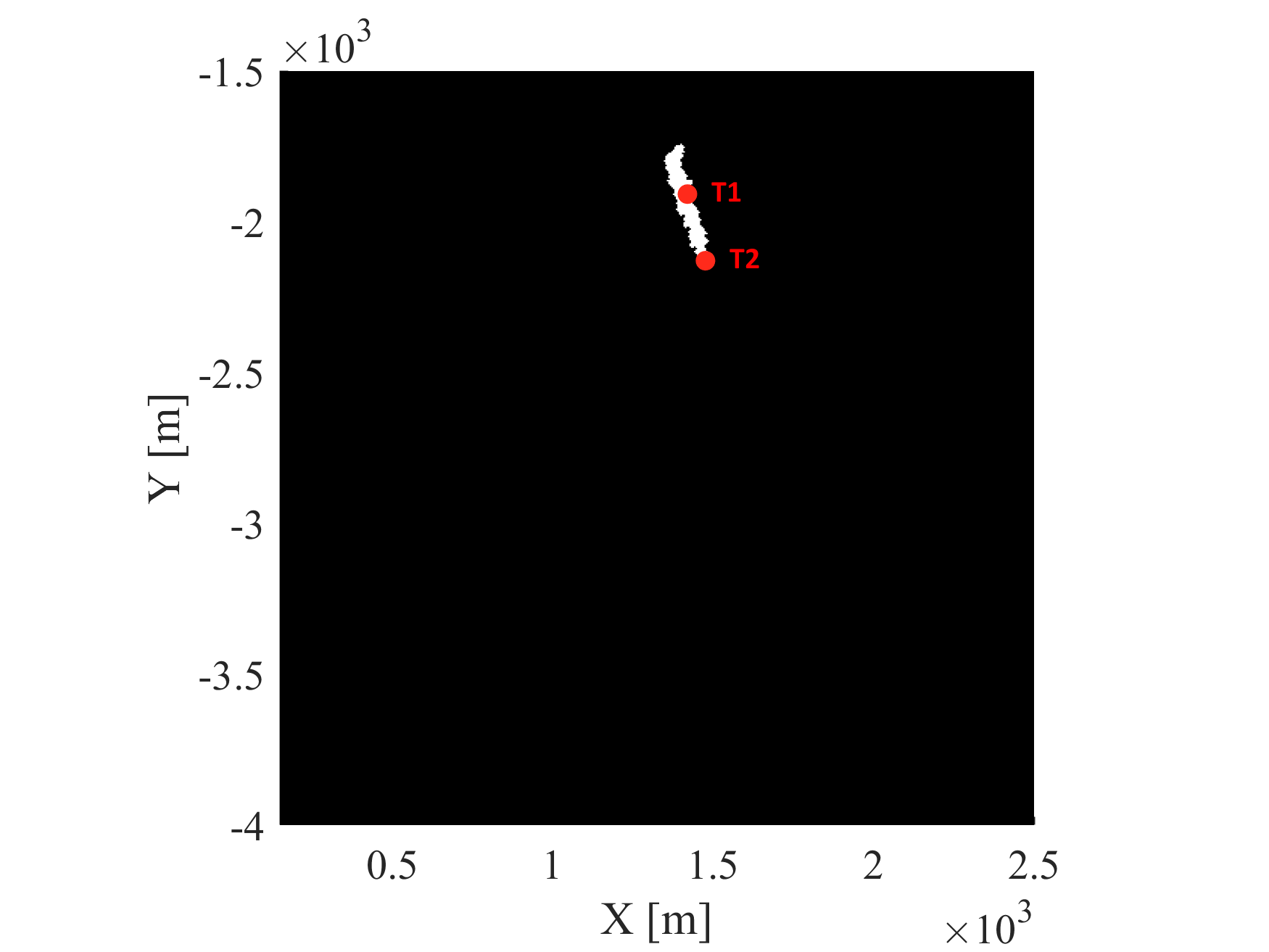}}
    \hfil%
    \subfloat[][]{
    \includegraphics[width=0.3\textwidth, trim=40 0 40 0]{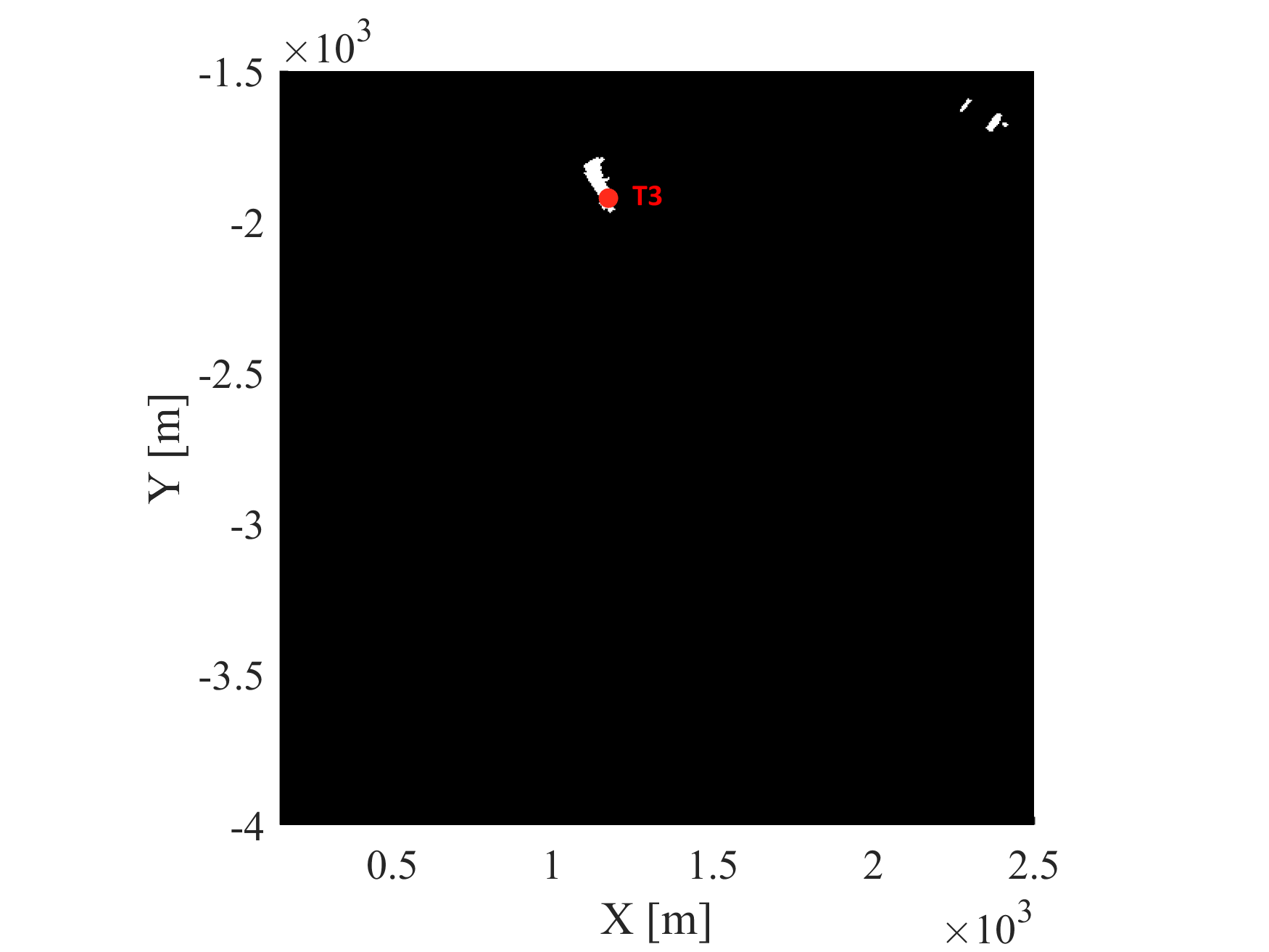}}
    \hfil%
    \subfloat[][]{
    \includegraphics[width=0.3\textwidth, trim=40 0 40 0]{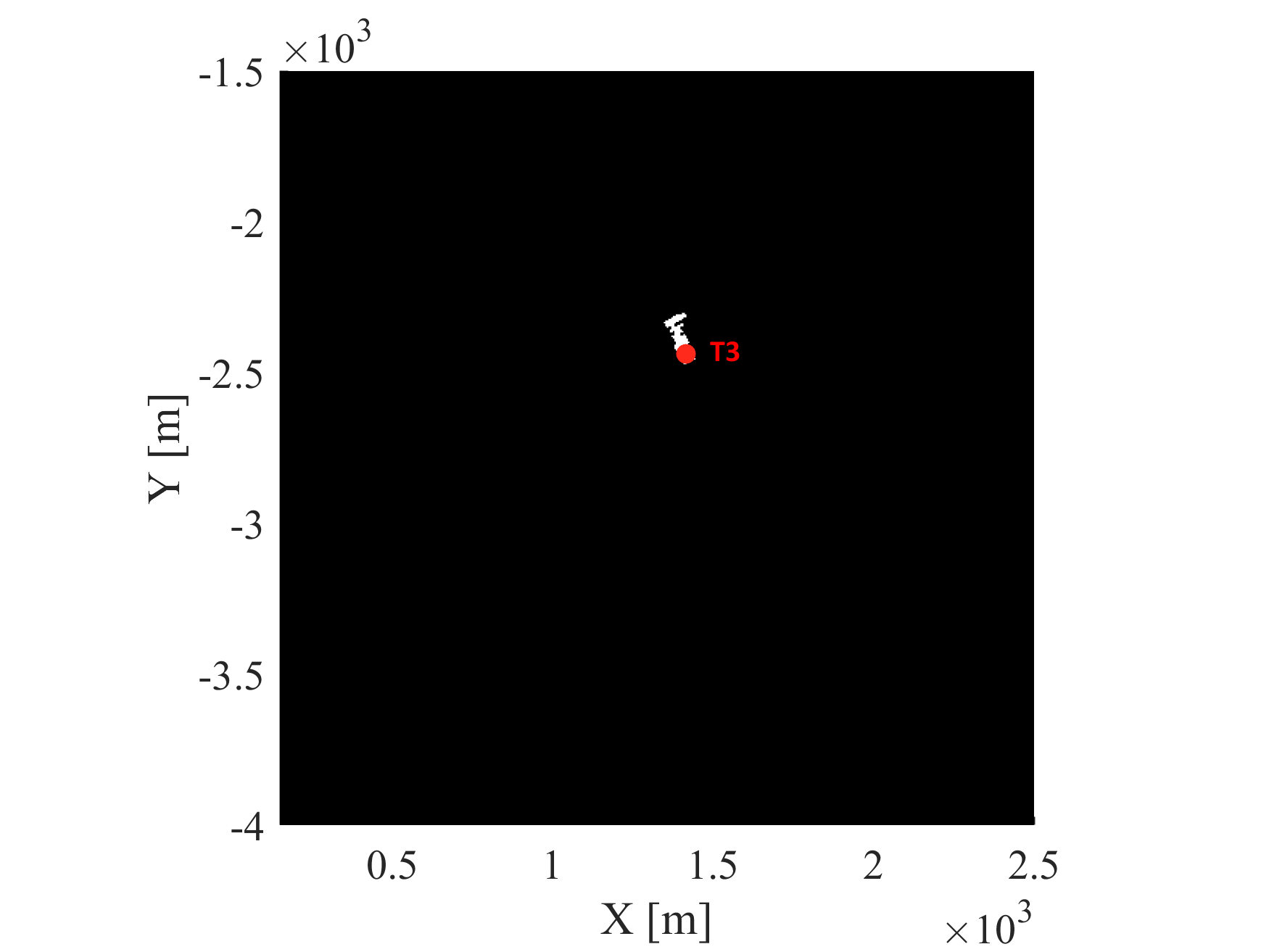}}
    \\[-2em]%
    \subfloat[][]{
    \includegraphics[width=0.3\textwidth, trim=40 0 40 0]{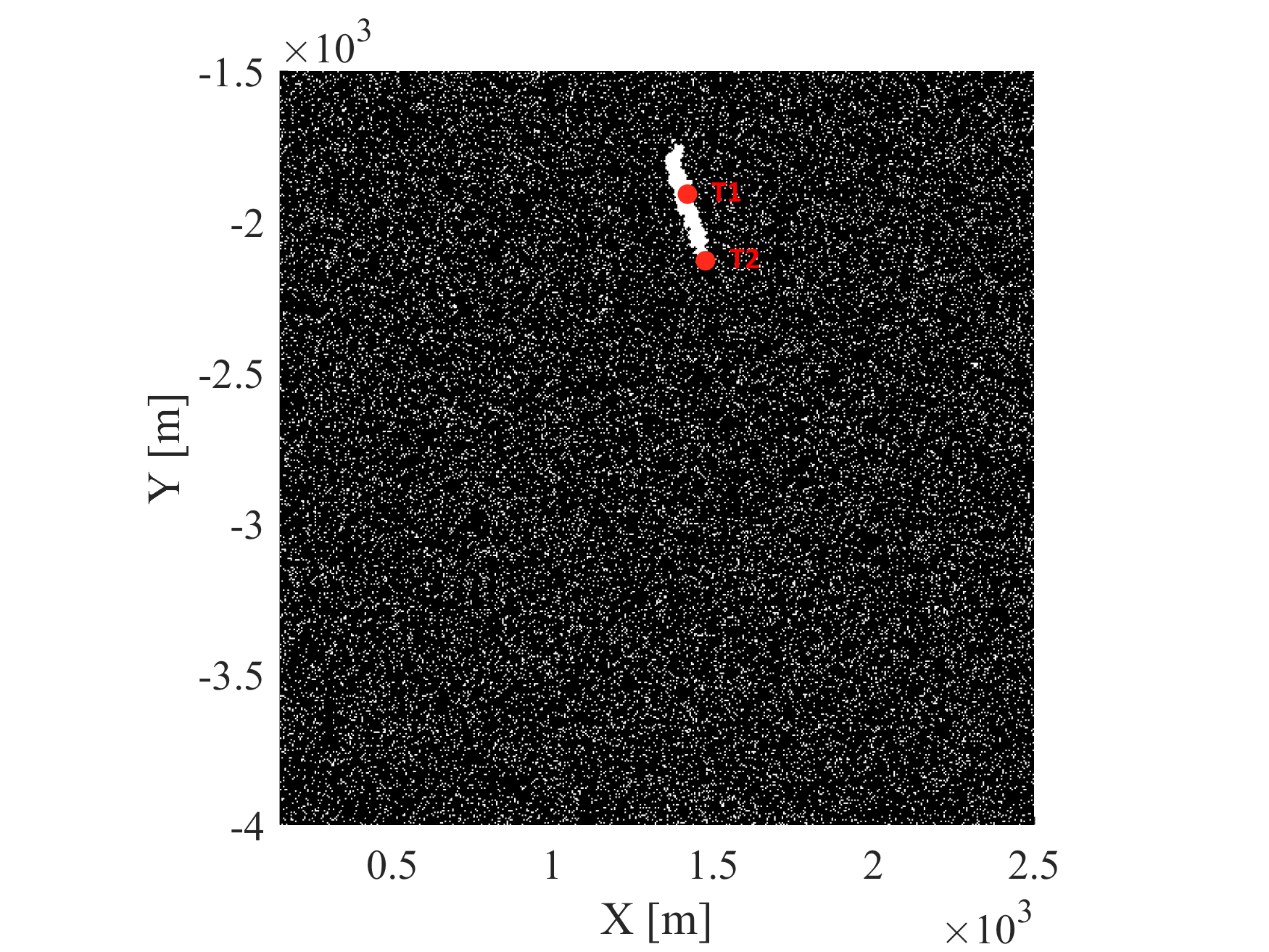}}
    \hfil%
    \subfloat[][]{
    \includegraphics[width=0.3\textwidth, trim=40 0 40 0]{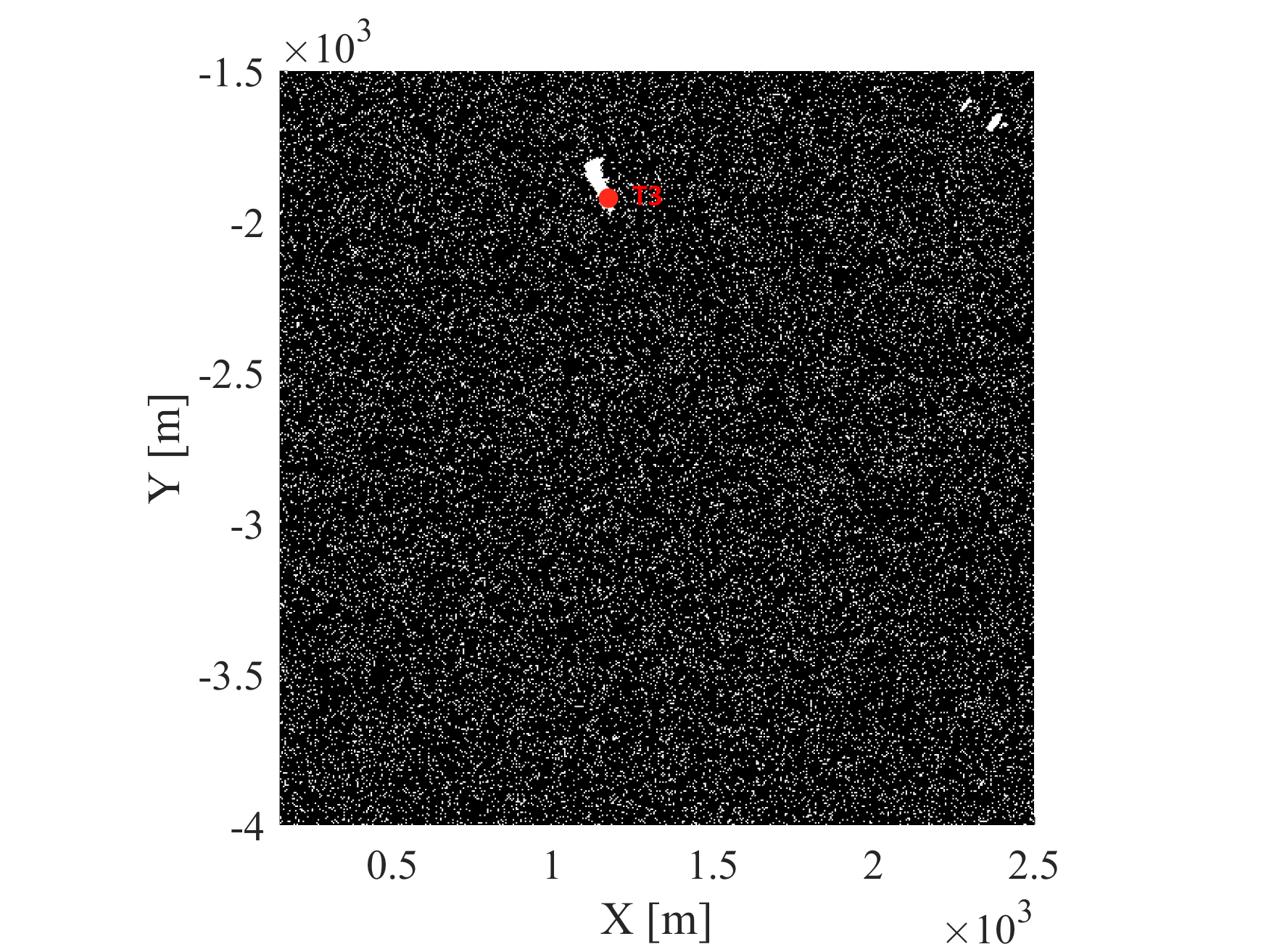}}
    \hfil%
    \subfloat[][]{
    \includegraphics[width=0.3\textwidth, trim=40 0 40 0]{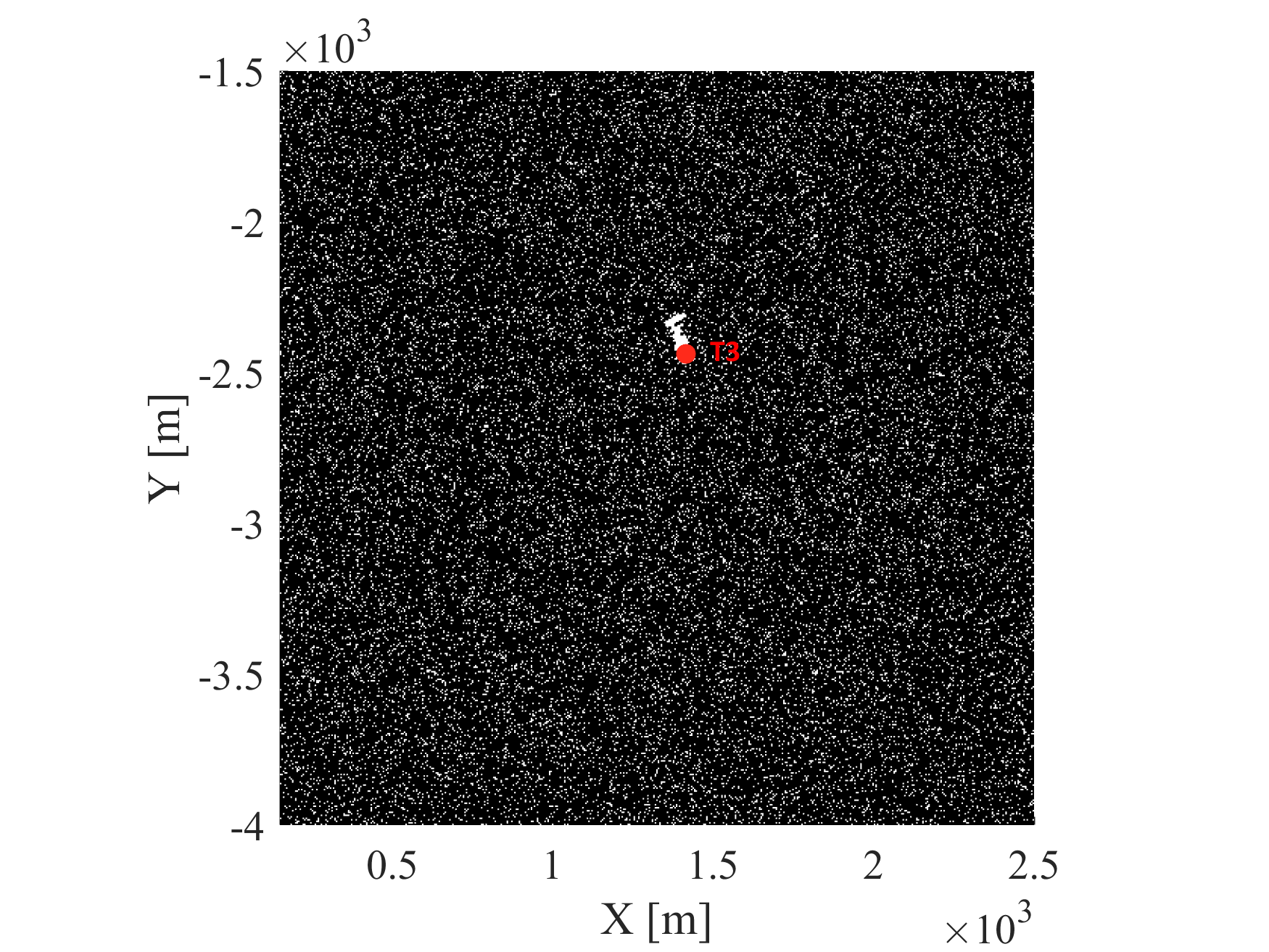}}
    \\[-2em]%
    \subfloat[][]{
    \includegraphics[width=0.3\textwidth, trim=130 240 140 230]{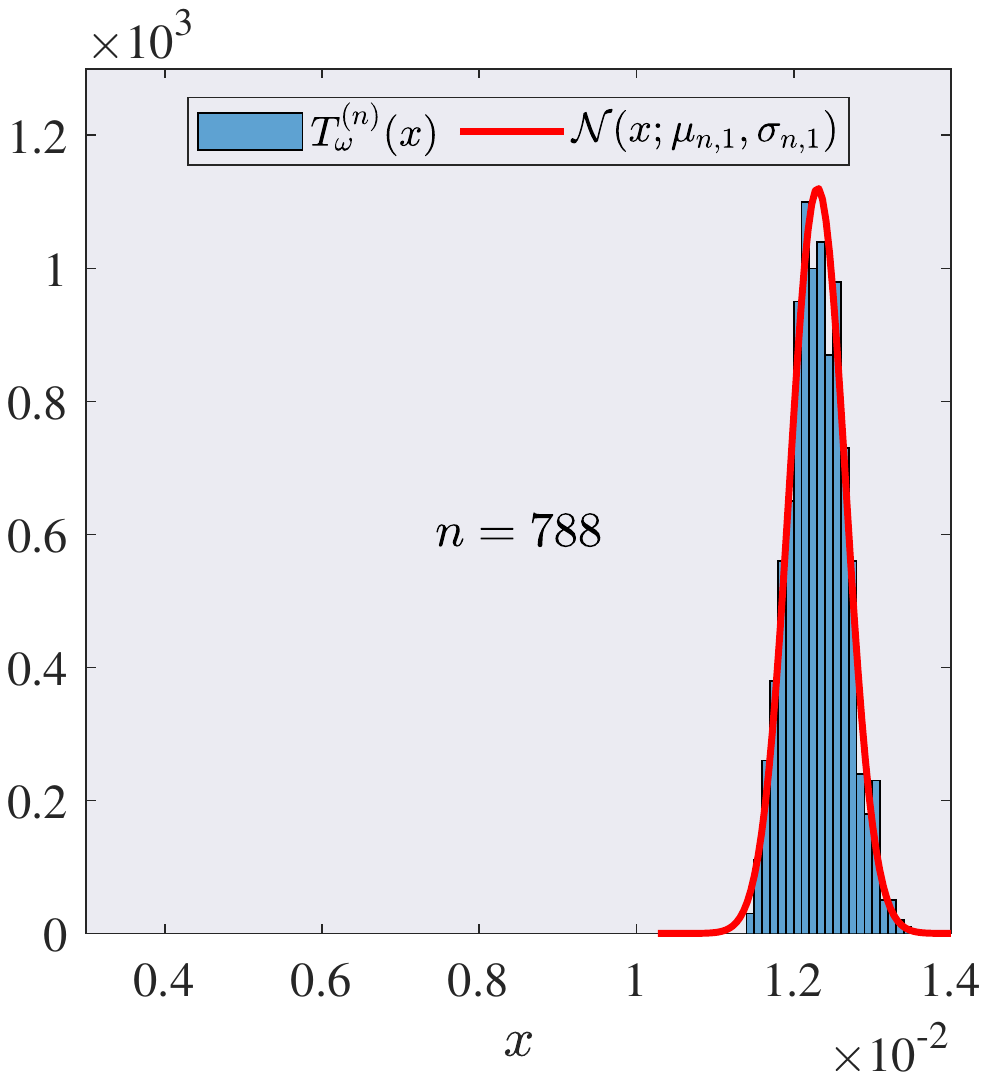}}
    \hfil%
    \subfloat[][]{
    \includegraphics[width=0.3\textwidth, trim=130 240 140 240]{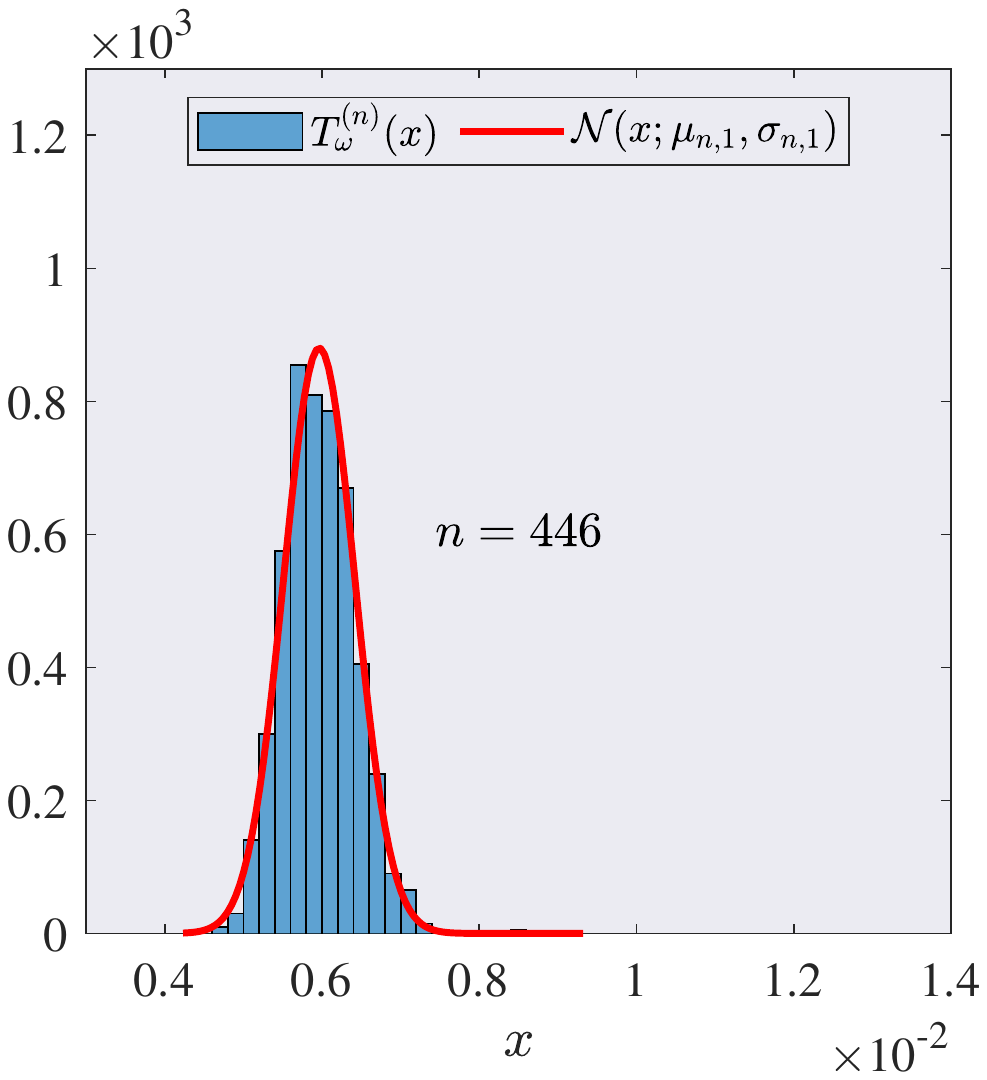}}
    \hfil%
    \subfloat[][]{
    \includegraphics[width=0.3\textwidth, trim=130 240 140 240]{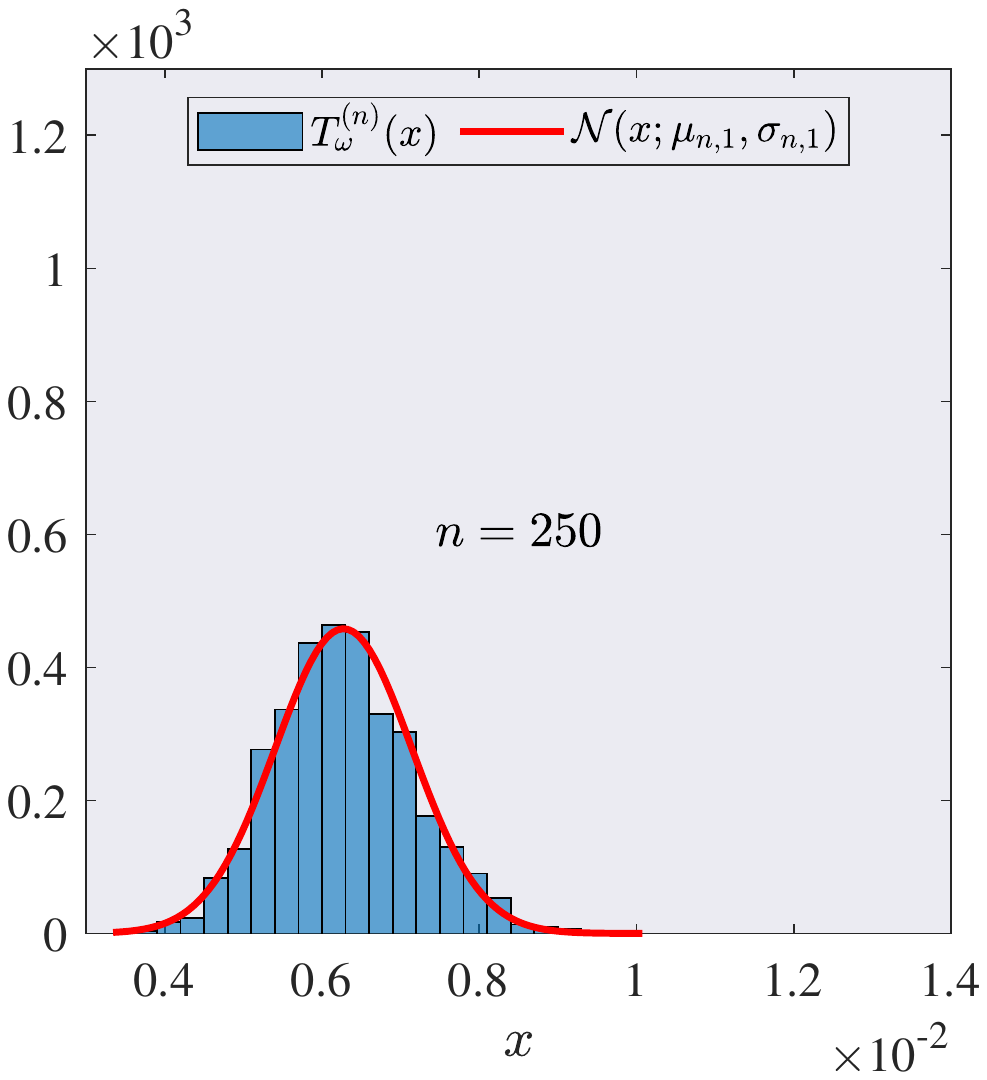}}
    \caption{Analysis of the normalized D3F statistics~\eqref{eq:D3F_dep_Tn} in the extended target scenario. Panels in the first row show three radar frames with AIS contacts overlaid. From these, a rectangular region (indicated with a red box) is extracted; then, persistent clutter is masked out and a detector applied. The output of this processing stage is illustrated in the second row, with the targets lighting up $n=788$, $n=446$ and $n=250$ respectively, from left to right. Since the false alarm level is extremely low, synthetic clutter is introduced (third row). Finally, the distribution of the normalized D3F statistic is computed from $1000$ independent realizations and showed in the panels on the last row; the superimposed red curves show the Gaussianity of the D3F statistic in all three cases.}
    \label{fig:ett_real}
\end{figure*}

\subsection{Extended target detection}
\label{sec:experiment_extended_target}
We consider now a more complex scenario to test what was presented in Sec.~\ref{sec:LDP_images}, where the data are representative of the output of a high-resolution sensor, e.g., a radar, a sonar, or an optical camera. The goal is to decide about the presence/absence of a target in the surveillance region. In this scenario, the D3F is the output of a DCNN, whose input is an image of $500$ by $500$ pixels; see Fig.~\ref{fig:d3fcnn_arch} for an illustration of the DCNN architecture. 

In this setup, the DCNN is trained starting from $5120$ examples of elliptical targets and as many examples of rectangular targets, whose position and dimensions are generated randomly; in the case of targets with elliptical shape, they are also rotated by a random angle (see Fig.~\ref{fig:ETT_Fig}). All the target shape parameters are selected independently, and the random generation allows for partial observations (target on the boundary only partially visible) to be represented in the training set. For each realization of the target shape parameters, $5$ independent realizations of the background noise are generated. In this way, the resulting training set is composed by $m_y=76800$ images, of which $25600$ generated from the null hypothesis (absence of target), and $51200$ from the alternative one (one half are targets with elliptical shape and the other half are targets with rectangular shape). All the images are generated in agreement with the model described in Sec.~\ref{sec:LDP_images}, and %
some notional examples of images are reported in Fig.~\ref{fig:ETT_Fig}. The training set is generated with %
pixel-wise detection and false alarm probabilities $p_1=0.9$ and $p_0=0.1$, respectively. The threshold is selected as in equation~\eqref{eq:gamma_image} to approximately achieve a desired false alarm level of $0.3$, $0.075$ and $0.005$. The related miss detection probability is reported in the upper panel of Fig.~\ref{fig:extended_target_clt_beta}, where we simulate a target with a circular shape, located in the center of the image, with $n$ being the number of resolution cells occupied by the target. The miss detection probability is estimated via Monte Carlo simulations with $10^5$ runs, and compared with the exact asymptotics via the direct estimation method and the Gaussian approximation.

To further shade light on the detector behaviour, we also perform a sensitivity analysis with respect to the pixel-wise detection and false alarm probabilities, $p_1$ and $p_0$, respectively. Specifically, we simulate the mismatched scenario where $p_1=0.8$ and $p_0=0.2$, reported in the upper panel of Fig.~\ref{fig:extended_target_clt_beta}. 

As observed in the previous subsections, the Gaussian approximation is quite good for small values of $n$, whereas (as it is expected) the empirical miss detection probabilities are closer to the exact asymptotic as $n$ increases. The curves, obtained via exact asymptotics, are displayed starting from values of $n$ where the approximation is meaningful, i.e., when the expected mean of the decision statistic is larger than the threshold (see the panel at the bottom of Fig.~\ref{fig:extended_target_clt_beta}) for the matched scenario. 

Clearly, the performance of the mismatched scenario is worse than the matched one given that the pixel-wise false alarm rate is higher ($0.2$ instead of $0.1$) and the pixel-wise detection rate is lower ($0.8$ instead of $0.9$). However, it is worthwhile highlighting that, even if the training and the actual input data exhibit a significant mismatch, the D3F is still able to perform target detection when the size of the target increases. 

For each value of the desired false alarm level $\alpha$, we estimate the actual false alarm level and plot it in the mid panel of Fig.~\ref{fig:extended_target_clt_beta}. We observe that the Gaussian approximation is less accurate as $\alpha$ decreases (between $10^{-4}$ and $10^{-3}$) because the tail of the decision statistic is not necessarily Gaussian, and would be more accurately ruled by the related LMGF and its Fenchel-Legendre transform. 
The Gaussian approximation, discussed in Sec.~\ref{sec:CLT_Dep}, is pretty accurate for both hypotheses as long as we focus on values of the decision statistic close to the mean, as we can observe in the histograms of the normalized statistic~\eqref{eq:D3F_dep_Tn}, reported in Fig.~\ref{fig:histograms_extended_target} for both the matched (left-side panels) and mismatched scenarios (right-side panels). A fixed offset to the decision statistic is applied so that the resulting expected mean of decision statistic distribution, under ${\cal H}_0$, is zero. As explained in Sec.~\ref{sec:CLT_Dep}, the decision statistic under ${\cal H}_0$ is independent on $n$ [see~\eqref{eq:gaussian_dep}]. Indeed, the standard deviation $\sigma_{0,n}$ of the normalized statistic~\eqref{eq:D3F_dep_Tn} converges to zero faster than $\frac{1}{\sqrt{n}}$ (see the bottom panel in Fig.~\ref{fig:extended_target_clt_beta}). However, the data are still well approximated with a Gaussian distribution, as illustrated in Fig.~\ref{fig:histograms_extended_target}. On the other hand, the convergence under ${\cal H}_1$ is empirically verified in agreement with the LDP. Indeed, $\mu_{1,n}$ converges to an asymptotic value $\mu_1$, whereas $\sqrt{n}\sigma_{1,n}$ converges to an asymptotic value $\sigma_1$, as reported in the panel at the bottom of Fig.~\ref{fig:extended_target_clt_beta}.

\subsection{Extended target detection in a real-world scenario using X-band radar data}
In this subsection we consider real-world data collected by a coherent high-resolution X-band maritime radar located in the Gulf of La Spezia, Italy. The radar is a low power, compact, lightweight, quickly deployable system, while still achieving high performance with relatively simple electronics. This is thanks to the use of pulse compression and linear frequency modulated continuous wave. The radar has an antenna mounted on a rotor with variable rotating speed and the possibility to lock and hold the position towards a specific direction with a $0.1^\circ$ accuracy. The range resolution is approximately $1$~m, while the angular resolution is $0.172^\circ$~\cite{Errasti-Alcala15}.
More details about the radar system, the signal processing chain, and the data collection are available in~\cite{Errasti-Alcala14,Errasti-Alcala15,vivone2016joint,Vivone16}. 

For our analysis, we consider three acquisitions from two large vessels and a tugboat, illustrated in the top row of Fig.~\ref{fig:ett_real}. The Automatic Identification System (AIS) is used as ground truth; this is an common approach in literature (see, e.g.,~\cite{vivone2016joint,Vivone2017OU,Braca_TGARS}). The vessels observed in the radar images are T1, a container ship (MMSI: 351361000) being pushed by T2, a tugboat (MMSI: 247222500), and a passenger ship (MMSI: 255803790). From the radar frame, a rectangular region is selected (indicated with a red box); then, persistent clutter is masked out. A pixel-wise maximum likelihood detector is applied in each resolution cell, see more details in~\cite{vivone2016joint}. The result of this processing step are binary images, reported in the second row of Fig.~\ref{fig:ett_real}. 
It can be seen that, for each radar frame, it is possible to detect correctly the extended targets, with some clutter still being present (in the upper right hand corner of the second row's central panel).

The false alarm rate is very low, creditably so from the surveillance perspective but not so useful to explore our performance prediction results. For this reason we %
introduce at this point some synthetic clutter, as %
in the model~(\ref{eq:dep_H1})-(\ref{eq:dep_H0}). 
Examples of binary images with synthetic clutter added are reported in the third row of Fig.~\ref{fig:ett_real}. 
These represent the input to compute
the normalized D3F statistic~\eqref{eq:D3F_dep_Tn} using the same DCNN, trained with synthetic data, described in the previous subsection. It is interesting to notice that the DCNN is able to detect the target correctly, even if it was not trained directly on radar data. 

Finally, from the generation of independent pixel-wise false alarms, with $p_0=0.1$, we have multiple independent realizations of the normalized D3F statistic for each radar acquisition. Then, we can construct histograms of the decision statistic (last row of Fig.~\ref{fig:ett_real}), which again appears evidently Gaussian, thus confirming the convergence already discussed in Sec.~\ref{sec:CLT_Dep} with real-world data.

\begin{figure*}[!t]
\normalsize
\begin{align}
    L^{(n)} &= \frac{1}{n} \log w_{\theta_*} \prod_{i=1}^n \frac{f (x_i | \theta_*)}{f (x_i | \theta_0)} + \frac{1}{n} \log \left(1 + \frac{ \displaystyle \sum_{\theta \in \Theta, \theta \neq \theta_*} w_\theta \prod_{i=1}^n f (x_i | \theta)}{w_{\theta_*}\prod_{i=1}^n f (x_i | \theta_*)} \right)  \nonumber \\
    &= \underbrace{\frac{1}{n}\log w_{\theta_*}}_{\rightarrow 0} + \underbrace{\frac{1}{n}\log \prod_{i=1}^n \frac{f (x_i | \theta_*)}{f (x_i | \theta_0)}}_{ L^{(n)}_{\theta_*} } + \frac{1}{n}  \log \left(1 + \underbrace{\displaystyle{ \sum_{\theta \in \Theta, \theta \neq \theta_*}} \frac{w_\theta}{w_{\theta_*}} \, \exp\left[{-n \left(\frac{1}{n} \sum_{i=1}^n \log \frac{f (x_i | \theta_*)}{f (x_i | \theta)}\right)}\right]}_{R_n} \right), \label{eq:L_n_derivation}
\end{align}
\hrulefill
\vspace*{4pt}
\end{figure*}

\section{Conclusion}
\label{sec:conclusion}

In this paper, we have proposed a novel method to analyze the performance of a generic Machine Learning (ML) classification technique. The ML classifier is based on a suitable decision statistic, referred to as the Data-Driven Decision Function (D3F), which is learned from training data, and its performance is defined in terms of error probabilities and their convergence rates. 

We have provided the mathematical conditions, based on the theory of large deviations, for the D3F to exhibit an error probability that vanishes exponentially with a parameter $n$, which represents the amount of information accumulated about the classification problem. Three different classification problems have been investigated: %
$i)$ independent and identically distributed data with simple hypotheses, $ii)$ conditionally independent data with composite hypotheses, and $iii)$ (weakly) dependent data with an unknown dependency structure. %
In the first two settings, the parameter $n$ is the number of observations available for testing; in the third setting, it represents the size of an extended target in a radar image. 

The mathematical conditions can be verified numerically exploiting the available set of data or by generating synthetic data. %
Thanks to the theory of large deviations, it has been shown that such conditions depend on the Fenchel-Legendre transform of the cumulant-generating function of the D3F. Moreover, we have derived approximations for %
two asymptotic regimes of the D3F statistic. In the first one, referred as to as the \emph{small deviations} regime (related to the central limit theorem), it is possible to establish the convergence of the normalized D3F statistic to a Gaussian distribution for a large enough value of $n$. This property can be used to set a desired, asymptotic false alarm probability, which turns out to be accurate also for %
operationally-relevant values of $n$. In the second regime, referred to as the \emph{large deviations} regime, we have established the conditions for the error probabilities to vanish to zero exponentially. Then, exploiting the exact asymptotics, which is a refinement of the large deviations theory, we have also provided an accurate approximation of the error probability curve as function of $n$. 
All the theoretical findings are corroborated and supported by extensive numerical simulations, as well as by an analysis of real-world data acquired by an X-band marine surveillance radar system. 

We believe that the proposed method is especially relevant in the context of model-based ML, where the statistical structure of problem is partially known or can be easily simulated, and in the context of reinforcement learning, where the forecast of performance is crucial to maximize the system reward.

\appendices
\section{Limit of the LLR's scaled LMGF with conditionally independent observations}
\label{sec:appendix_lmgf}

Let us compute the asymptotic scaled LMGF of $L^{(n)}$~\eqref{eq:conditional_Ln} under $\mathcal{H}_1$ when the true parameter is $\theta_*$ and the data are generated according to $f(\cdot|\theta_*)$.
We can write the equality~\eqref{eq:L_n_derivation} and note that the first term converges to zero assuming a non-trivial prior $w_{\theta_*}>0$. The terms at the exponent in $R_n$ converge to the KL divergence between $\theta_*$ and $\theta$ when $\theta_*$ is true. Specifically, given that the data are IID, the following holds
\begin{equation}
\frac{1}{n} \sum_{i=1}^n \log \frac{f (x_i | \theta_*)}{f_1 (x_i | \theta)} \xrightarrow[]{a.s.} {\cal D}(\theta_* || \theta) = \mathbb{E}\left[ \log \frac{f (x_i | \theta_*)}{f (x_i | \theta)} \right] > 0, 
\end{equation}
where the convergence is a.s. assuming mild regularity conditions, 
and that the hypotheses identified by the different values of $\theta$ are distinguishable, namely $ {\cal D}(\theta_* || \theta) > 0,\, \forall \, \theta_*, \theta \in \Theta$ with $\theta_* \neq \theta$. Under mild regularity conditions of the log-likelihood ratio the convergence holds also in mean square. At this point it is easy to recognize in~(\ref{eq:L_n_derivation}) that $R_n$ vanishes exponentially fast with $n$. The asymptotic LMGF of the first term $L^{(n)}_{\theta_*}$ is given by: 
\begin{equation}
     \lim_{n \rightarrow \infty} \frac{1}{n} \log \mathbb{E}\left[ \exp(n\,t\,L^{(n)}_{\theta_*}) \right] = \varphi_{\theta_*}(t),
\end{equation}
with $\varphi_{\theta_*}(t) = \log \mathbb{E}\left[ \exp\left( t\,\log \frac{f (x_1 | \theta_*)}{f (x_1 | \theta_0)} \right) \right]$. The asymptotic LMGF of $L^{(n)}$ is given by:
\begin{equation}
\lim_{n \rightarrow \infty} \frac{1}{n} \log \mathbb{E}\left[ \exp\left(n\, t\,L^{(n)}_{\theta_*}\right) (1+R_n)^t \right].
\label{eq:asy_LMGF}
\end{equation}

We shall now multiply and divide by the expected value $\mathbb{E}\left[ \exp\left(n\, t\,L^{(n)}_{\theta_*}\right)\right] = \exp \left(n\, \varphi_{\theta_*}(t) \right)$, obtaining
\begin{IEEEeqnarray}{rCl}
\IEEEeqnarraymulticol{3}{l}{
\frac{1}{n} \log \mathbb{E}\left[ \exp\left(n\, t\,L^{(n)}_{\theta_*}\right) (1+R_n)^t \right]}\nonumber\\* \quad
&=&\varphi_{\theta_*}(t) + \frac{1}{n} \log \frac{\mathbb{E}\left[ \exp\left(n\, t\,L^{(n)}_{\theta_*}\right) (1+R_n)^t \right]}{\exp \left(n\, \varphi_{\theta_*}(t) \right)}.
\label{eq:LMGF2}    
\end{IEEEeqnarray}

Since $R_n>0$, the second term in~\eqref{eq:LMGF2} is larger than $\frac{1}{n} \log \frac{\mathbb{E}\left[ \exp\left(n\, t\,L^{(n)}_{\theta_*}\right) \right]}{\exp \left(n\, \varphi_{\theta_*}(t) \right)} = 0$ (the denominator and the numerator are equal). Then, we have \begin{equation}
\frac{1}{n} \log \mathbb{E}\left[ \exp\left(n\, t\,L^{(n)}_{\theta_*}\right) (1+R_n)^t \right] \geq  \varphi_{\theta_*}(t).
\label{eq:lbound}
\end{equation}
We can now apply Hölder's inequality to the argument of the expectation in~(\ref{eq:asy_LMGF}) and obtain
\begin{IEEEeqnarray}{rCl}
\IEEEeqnarraymulticol{3}{l}{%
    \frac{1}{n} \log \mathbb{E}\left[ \exp\left(n\, t\,L^{(n)}_{\theta_*}\right) (1+R_n)^t \right] 
    }\nonumber\\* \quad
    &\leq& \frac{1}{n\,p} \log \mathbb{E}\left[ \exp\left(n\,p\, t\,L^{(n)}_{\theta_*}\right) \right] %
    + \frac{1}{n\,q} \log \mathbb{E}\left[ (1+R_n)^{q\,t} \right] \nonumber \\*
    &=& \frac{1}{p} \varphi_{\theta_*}(p\,t) + \frac{1}{n\,q} \log \mathbb{E}\left[ (1+R_n)^{q\,t} \right],
\end{IEEEeqnarray}
where $p^{-1} + q^{-1} = 1$. The second term in the equation above vanishes if $R_n$ (which is strictly positive) grows as $n^r$ with $r<1$. Now, given that $R_n$ vanishes exponentially in probability thanks to the continuous mapping, the term $(1+R_n)^{q\,t} \rightarrow 0$ in probability. The convergence in mean holds if the function $(1+x)^{q\,t}$ is bounded, which is the case assuming a limited interval $t_{min}<t<t_{max}$, or if the convergence of $R_n$ is a dominated convergence (see details in~\cite{newey1994large}). Consequently, we have 
\begin{eqnarray}
  \lim_{n \rightarrow \infty}  \frac{1}{n} \log \mathbb{E}\left[ \exp\left(n\, t\,L^{(n)}_{\theta_*}\right) (1+R_n)^t \right] &\leq& \frac{1}{p} \varphi_{\theta_*}(p\,t), \nonumber \\ 
 \lim_{n \rightarrow \infty}  \frac{1}{n} \log \mathbb{E}\left[ \exp\left(n\, t\,L^{(n)}_{\theta_*}\right) (1+R_n)^t \right] &\geq & \varphi_{\theta_*}(t),
   \label{eq:ubound}
\end{eqnarray}
where in the second inequality we have exploited the continuity of $\varphi_{\theta_*}(t)$; the first inequality holds also in the limit of $p \rightarrow 1$. Considering both~(\ref{eq:lbound}) and~(\ref{eq:ubound}), we can conclude that the asymptotic scaled LMGF under ${\cal H}_1$ is $\varphi_{\theta_*}(t)$.

Let us focus on ${\cal H}_0$, remembering that the data are generated according to $f(\cdot|\theta_0)$. We can rewrite the log-likelihood ratio~(\ref{eq:conditional_Ln}) as follows: 
\begin{IEEEeqnarray}{rCl}
    L^{(n)} &=&\frac{1}{n} \log w_{\theta_m} + \underbrace{\frac{1}{n} \log \prod_{i=1}^n \frac{f (x_i | \theta_m)}{f (x_i | \theta_0)}}_{L_{\theta_m}^{(n)}} \nonumber \\
    && +\> \frac{1}{n}  \log \left(1 + \underbrace{{ \sum_{\theta \in \Theta,\theta \neq \theta_m}} w_\theta \, \exp\left({-n D_n(\theta)}\right)}_{R_n} \right), \label{eq:L_n_derivation_H0}  \IEEEeqnarraynumspace%
\end{IEEEeqnarray}
where
\begin{equation*}
    \theta_m = \arg \min_{\theta \in \Theta}\,{\cal D}(\theta_0 || \theta)
\end{equation*}
and
\begin{equation*}
    D_n(\theta) = \frac{1}{n} \sum_{i=1}^n \log \frac{f (x_i | \theta_0)}{f (x_i | \theta)} - \frac{1}{n} \sum_{i=1}^n \log \frac{f (x_i | \theta_0)}{f (x_i | \theta_m)}.
\end{equation*}

The term $D_n(\theta)$ converges a.s. to the difference between the divergences ${\cal D}(\theta_0 || \theta) - {\cal D}(\theta_0 || \theta_m)$, which is strictly positive %
if there is a single point of minimum $\theta_m$ (in the case of multiple minimum points the procedure is analogous). The structure of the log-likelihood ratio~(\ref{eq:L_n_derivation_H0}) is formally the same of~(\ref{eq:L_n_derivation}), and the asymptotic scaled LMGF is given by $\varphi_{\theta_0}(t) = \log \mathbb{E}\left[ \exp\left( t\,\log \frac{f (x_1 | \theta_m)}{f (x_1 | \theta_0)} \right) \right]$, where the expectation is taken under ${\cal H}_0$ and the true parameter is $\theta_0$.

\section{Weak law of large numbers when the characterization set is equal to the training set}
\label{sec:appendixB}

It is possible to exploit the training set to compute the quantities of interest, such as the LMGF, as indicated in Sec.~\ref{sec:estimation_rate}. This is the case when the characterization set and the training set are the same or overlapped. The goal is to show that the sample means, such as~\eqref{eq:LMGF_est} and~\eqref{eq:scaled_LMGF_est}, converge to their expected means even if $\tau_j = t_{\bm{\omega}_{\cal Y}}(y_j)$, $\forall j=1,2,\dots,m_y$, where $\bm{\omega}_{\cal Y}$ is a stochastic function of the training data ${\cal Y} = (y_1,y_2,\dots,y_{m_y})$, provided by the learning mechanism.\footnote{Note that we skip the dependency from the hypothesis to simply the notation.} Therefore, we should consider a joint space probability between ${\cal Y}$ and $\bm{\omega}$, where $\bm{\omega}$ is referring to the aforementioned $\bm{\omega}_{\cal Y}$. Let us also assume that $\bm{\omega}$ takes values in a finite and discrete space $\Omega$.

Let us consider the following event for any given $\epsilon>0$
\begin{equation}
    {\cal A}_\epsilon(\bm{\omega},{\cal Y}) = \left\{ \left| \frac{1}{m_y} \sum_{j=1}^{m_y} g ( t_{\bm{\omega}}(y_j) ) - \bar{g}_{\bm{\omega}} \right| > \epsilon \right\},
\end{equation}
where $\bar{g}_{\bm{\omega}} = \mathbb{E} \left[g ( t_{\bm{\omega}}(y_1) ) | \bm{\omega} \right]$, where $y$ is distributed accordingly to the marginal distribution of an element in ${\cal Y}$, which is independent of $\bm{\omega}$. 
Let us now evaluate the probability of the event of interest
\begin{align}
\mathbb{P} \left[ {\cal A}_\epsilon (\bm{\omega},{\cal Y}) \right] &=
\sum_{\bar{\bm{\omega}}\in\Omega} \mathbb{P} \left[ {\cal A}_\epsilon(\bm{\omega},{\cal Y}) \cap \left\{\bm{\omega} = \bar{\bm{\omega}} \right\} \right]  \nonumber \\
& =  \sum_{\bar{\bm{\omega}}\in\Omega} \mathbb{P} \left[ {\cal A}_\epsilon(\bar{\bm{\omega}},{\cal Y}) \right] \mathbb{P} \left[ \left\{\bm{\omega} = \bar{\bm{\omega}} \right\} \left|  {\cal A}_\epsilon (\bar{\bm{\omega}},{\cal Y}) \right. \right]  \nonumber \\
& \leq \sum_{\bar{\bm{\omega}}\in\Omega} \mathbb{P} \left[ {\cal A}_{\epsilon}(\bar{\bm{\omega}},{\cal Y}) \right]. \label{eq:bound_appendix}
\end{align}

Note that by construction when we evaluate $\mathbb{P}\left[ {\cal A}_{\epsilon}(\bar{\bm{\omega}},{\cal Y}) \right]$, $\bm{\omega}$ is fixed to the value $\bar{\bm{\omega}}$, and the evaluation of this probability is carried on according to the marginal of ${\cal Y}$. Now, being the observations $y_i$ IID, by invoking the law of large numbers we have that for $m_y \rightarrow \infty$ for any $\bar{\bm{\omega}}$ 
\begin{equation}
\mathbb{P} \left[ {\cal A}_{\epsilon}(\bar{\bm{\omega}},{\cal Y}) \right] \rightarrow 0.
\label{eq:LLN}
\end{equation}
From~\eqref{eq:bound_appendix}--\eqref{eq:LLN}, it follows directly that $\mathbb{P} \left[ {\cal A}_\epsilon (\bm{\omega},{\cal Y}) \right] \rightarrow 0$.

\bibliographystyle{IEEEtran}
\bibliography{IEEEabrv,mybib_upd_clean}

\end{document}